\documentclass[runningheads]{llncs}

\usepackage{eccv}

\usepackage{eccvabbrv}

\usepackage{graphicx}
\usepackage{multirow}
\usepackage{booktabs}
\usepackage{adjustbox}
\usepackage[table]{xcolor}
\usepackage{caption}
\usepackage{array}
\usepackage{longtable}

\usepackage[accsupp]{axessibility}  

\usepackage{hyperref}

\usepackage{orcidlink}
\usepackage[title]{appendix} 
\usepackage{titletoc}        
\usepackage{chngcntr}      
\usepackage[most]{tcolorbox} 
\newtcolorbox{PromptBox}[1]{
    enhanced,
    boxrule=0.8pt,
    colframe=black,
    colback=gray!10,
    coltext=black,
    colbacktitle=black,
    coltitle=white,
    fonttitle=\bfseries\sffamily,
    before upper={\fontsize{8pt}{13pt}\selectfont\raggedright},
    left=12pt, right=12pt, 
    top=10pt, bottom=10pt,
    title=#1,
    sharp corners
}
\makeatletter
\newcommand{\samethanks}{\textsuperscript{\@fnsymbol{1}}}
\makeatother

\begin{document}

\title{MV-STRIDE: Enabling MLLMs to Master Multi-View Spatial Reasoning via Hierarchical Capability Modeling} 


\author{
Jin Xu\inst{1}\orcidlink{0000-0002-1104-5875} \and
Xiaojian Huang\inst{1} \and
Zhuodong Luo\inst{1}\and
Zhihong Zhang\inst{1}\and
Xin Liu\inst{2}\and
Jiansheng Wei\inst{2}\and
Xinzhi Wang\inst{2}\and
Jie Zhao\inst{3}\thanks{Corresponding author.}\and
Xuejin Chen\inst{1}\samethanks
}

\authorrunning{MV-STRIDE,~Xu et al.}

\institute{MoE Key Laboratory of Brain-inspired Intelligent Perception and Cognition, University of Science and Technology of China, China \\
\email{xjchen99@ustc.edu.cn}\and
Huawei Noah's Ark Lab \and
Hefei University of Technology, China\\
\email{jzhaoch@hfut.edu.cn}
}

\maketitle

\begin{abstract}
  Despite the rapid progress of Multimodal Large Language Models (MLLMs) in 2D vision-language tasks, robust multi-view spatial reasoning remains a fundamental bottleneck due to the lack of structured 3D cognitive pathways in existing datasets. To address this, we introduce MV-STRIDE, a \textbf{M}ulti-\textbf{V}iew hierarchical \textbf{S}pa\textbf{T}ial \textbf{R}easoning dataset with \textbf{I}nterdependent and \textbf{D}ecomposed capabiliti\textbf{E}s. Moving beyond flat data structures, MV-STRIDE explicitly models the dependency relationships between foundational perception, scene understanding, and complex contextual reasoning, providing a coherent learning pathway aligned with human spatial cognition. We develop a systematic QA generation pipeline leveraging diverse 3D scene sources that enforces cross-view dependency constraints to prevent single-view solvability, generating multi-level spatial reasoning tasks supported by cognitively grounded chain-of-thought supervision for complex inference. Extensive evaluations demonstrate that our multi-stage training framework based on our hierarchical dataset achieves state-of-the-art performance across multiple spatial reasoning benchmarks, notably the multi-view oriented MMSI-Bench. Our approach enables MLLMs to maintain robust, 3D-consistent spatial reasoning across diverse viewpoints. The code and dataset are available at \url{https://co1dspring.github.io/MV-STRIDE/}.
  \keywords{Multimodal Large Language Model \and Dataset Construction \and Multi-view Spatial Reasoning}
\end{abstract}

\section{Introduction}
\begin{figure}[tb]
  \centering
  \includegraphics[width=\linewidth]{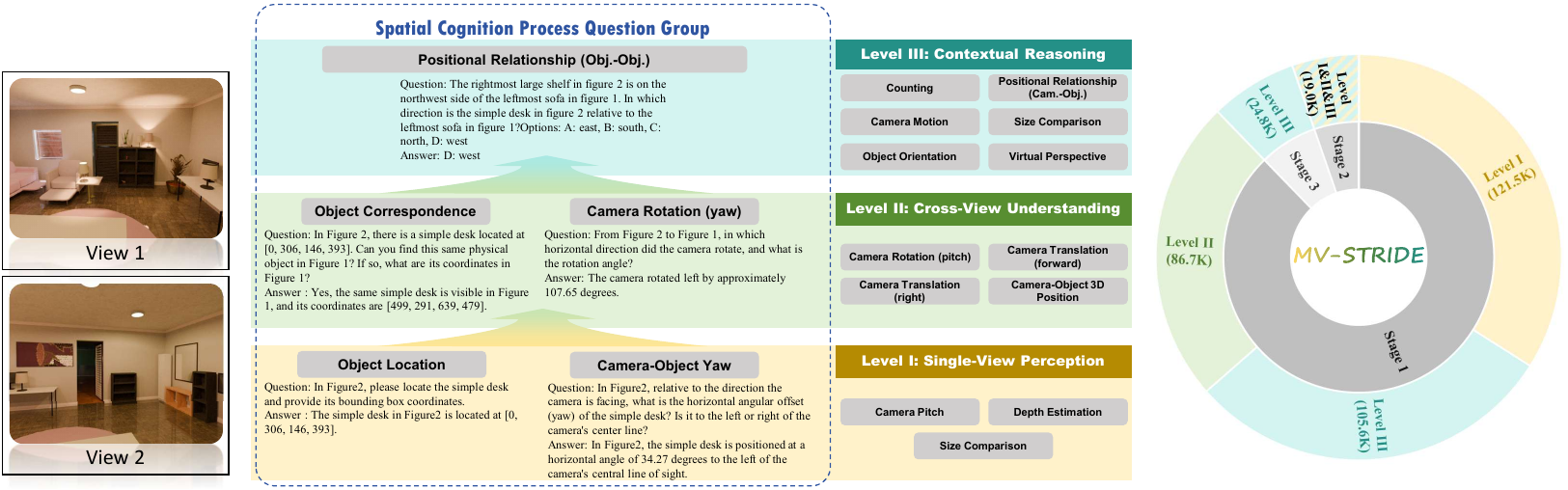}
  \caption{Overview of our MV-STRIDE. It encompasses a hierarchy of spatial tasks across three levels and 18 categories, sourced from over 600 synthetic and real indoor environments. This cognition-aligned structure supports the entire training lifecycle from SFT to RL, enabling MLLMs to develop human-like, multi-view consistent spatial reasoning capabilities.}
  \label{fig:teaser}
\end{figure}

In recent years, Multimodal Large Language Models (MLLMs)\cite{openai_gpt5,openai_gpt4o,guo2025seed1,hong2025glm,lu2025ovis2} have excelled in general vision–language tasks, especially fundamental spatial perception, which are essential for downstream applications such as embodied intelligence and autonomous driving. Despite these advances, robust spatial reasoning under multi-view visual conditions remains major unresolved challenges. Extensive evaluations of multi-view spatial reasoning benchmarks\cite{fu2024blink,yang2025mmsi,li2025viewspatial,ma20243dsrbench,xue2025reasoning} consistently reveal a significant performance gap between humans and MLLMs. Humans naturally integrate shared objects or structures across views into a coherent 3D cognitive map enabling flexibly viewpoint transformations and reasoning about arbitrary spatial relationship. In contrast, current MLLMs struggle to reliably establish such correspondences, maintain geometric consistency across views, or perform stable viewpoint-dependent reasoning. Error analyses of existing benchmarks\cite{yang2025mmsi,li2025viewspatial,xue2025reasoning} indicate that failures predominantly arise from deficiencies in scene-level reconstruction, cross-view correspondence, and viewpoint transformation, rather than from isolated perceptual errors.

A key cause of multi-view bottleneck lies in the design limitations of existing spatial reasoning datasets. Most spatial reasoning datasets\cite{chen2024spatialvlm,cheng2024spatialrgpt,cai2024spatialbot,wang2025spacevllm,lee2025perspective,liu2025spatialcot,wang2025spatial457,liu2025ssr} focus on single-view 2D planes, failing to support the formation of consistent 3D spatial cognition. Although recent efforts have introduced multi-view data\cite{ray2024sat,yang2025visual,zhang2025flatland,xu2025multi,zhan2025actial,wang2025towards,li2025spatialladder}, many datasets emphasize low-level quantitative perception (\eg depth estimation), which strengthens isolated perceptual skills but does not support the development of structured 3D reasoning. Furthermore, multi-view spatial reasoning is frequently treated as a flat or loosely defined capability, with tasks designed independently and without explicit dependency between perception, scene understanding, and reasoning. As a result, models lack a clear learning pathway from basic spatial perception to high-level multi-view reasoning, hindering their ability to generalize to complex, multi-step spatial inference.

To bridge this gap, we introduce MV-STRIDE, a \textbf{M}ulti-\textbf{V}iew hierarchical \textbf{S}pa\textbf{T}ial \textbf{R}easoning dataset with \textbf{I}nterdependent and \textbf{D}ecomposed capabiliti\textbf{E}s. Aligned with human cognition, MV-STRIDE organizes tasks into three levels: single-view perception, cross-view scene understanding, and complex contextual spatial reasoning. Unlike simple difficulty-based hierarchies, our framework explicitly models the dependency relationships among different spatial capabilities across levels. This design provides a coherent pathway for MLLMs to acquire stable 3D spatial awareness and reasoning behavior aligned with human cognition.



Following this design, we develop a pipeline that leverages diverse 3D-grounded assets to generate MV-STRIDE. Our pipeline employs a top-down generation strategy that simultaneously populates all three levels of the hierarchy. We first construct complex Level III tasks—specifically enforced by cross-view dependency constraints to preclude single-view shortcuts—and then decompose them into prerequisite Level I and II sub-questions. These tasks are organized into spatial cognition problem groups, where the ordered answers to sub-questions naturally form the constituent steps of Chain-of-Thought (CoT) annotations. This structured data facilitates a capability-aligned training strategy, spanning from supervised fine-tuning to reinforcement learning, ensuring model reasoning remains strictly aligned with the human spatial cognition process.
Models trained with our dataset and methodology achieve state-of-the-art performance across multiple spatial reasoning benchmarks, particularly in multi-view settings, thereby validating the superior effectiveness of our proposed approach.

Our main contributions are summarized as follows:

\begin{itemize}
    \item We introduce a hierarchical, capability-dependent organization of multi-view spatial reasoning tasks, which explicitly models the dependency structure from foundational spatial perception to high-level multi-view contextual reasoning.
    
    \item We design a cognition-driven QA generation pipeline and construct a multi-level multi-view spatial reasoning dataset MV-STRIDE with accurate 3D supervision, where tasks across different levels are explicitly connected through capability dependencies and organized into coherent spatial cognition problem groups.
    
    \item We empirically validate the effectiveness of the proposed dataset by multi-stage training, demonstrating consistent improvements on multiple spatial reasoning benchmarks.
\end{itemize}

\section{Related Work}
\subsection{Multimodal Spatial Reasoning Dataset.}
Recently, a plethora of large-scale spatial reasoning datasets has emerged to bolster the 3D awareness of MLLMs. Early efforts relied on pseudo-labeling pipelines to automatically generate extensive training annotations. SpatialVLM \cite{chen2024spatialvlm} pioneered automated 3D annotation for internet images, followed by refinements in region-level scene graphs \cite{cheng2024spatialrgpt} and RGB-D perception \cite{cai2025spatialbot}. To mitigate the inherent inaccuracies of pseudo-labels, recent research has shifted toward precise 3D ground-truth. Datasets such as SPAR-7M \cite{zhang2025flatland}, RefSpatial \cite{zhou2025roborefer}, SAT\cite{ray2024sat}, VST\cite{yang2025visual} and CA-1M \cite{daxberger2025mm} leverage real-world reconstructions (e.g., ScanNet++\cite{yeshwanth2023scannet++}) or simulated environments \cite{deitke2022} as data sources to provide accurate supervision across various spatial reasoning tasks. While these works have scaled single-view reasoning, they lack the structured, multi-layered pathways necessary to systematically foster complex 3D geometric consistency.

\subsection{Multimodal Multi-view Spatial Reasoning Dataset.}
Despite the growth in data scale, robust multi-view spatial reasoning remains a bottleneck as many existing datasets either treat multi-frame inputs as isolated contexts \cite{daxberger2025mm} or provide limited multi-view task categories \cite{ray2024sat, yang2025visual, zhang2025flatland}. Dedicated multi-view datasets like MultiSPA \cite{xu2025multi} offer large-scale perception labels (e.g., depth, motion) but are restricted to low-level quantitative estimation. Other efforts focus on isolated capabilities, such as viewpoint recognition \cite{zhan2025actial}, point correspondence \cite{wang2025towards}, or grounding-based reasoning \cite{li2025spatialladder}. Unlike these narrowly focused or perception-heavy datasets, MV-STRIDE introduces a comprehensive, three-level hierarchy that explicitly models the dependency between foundational perception and high-level contextual reasoning, ensuring robust 3D-consistent spatial inference.

\section{Method}
\begin{figure}[tb]
  \centering
  \includegraphics[width=\linewidth]{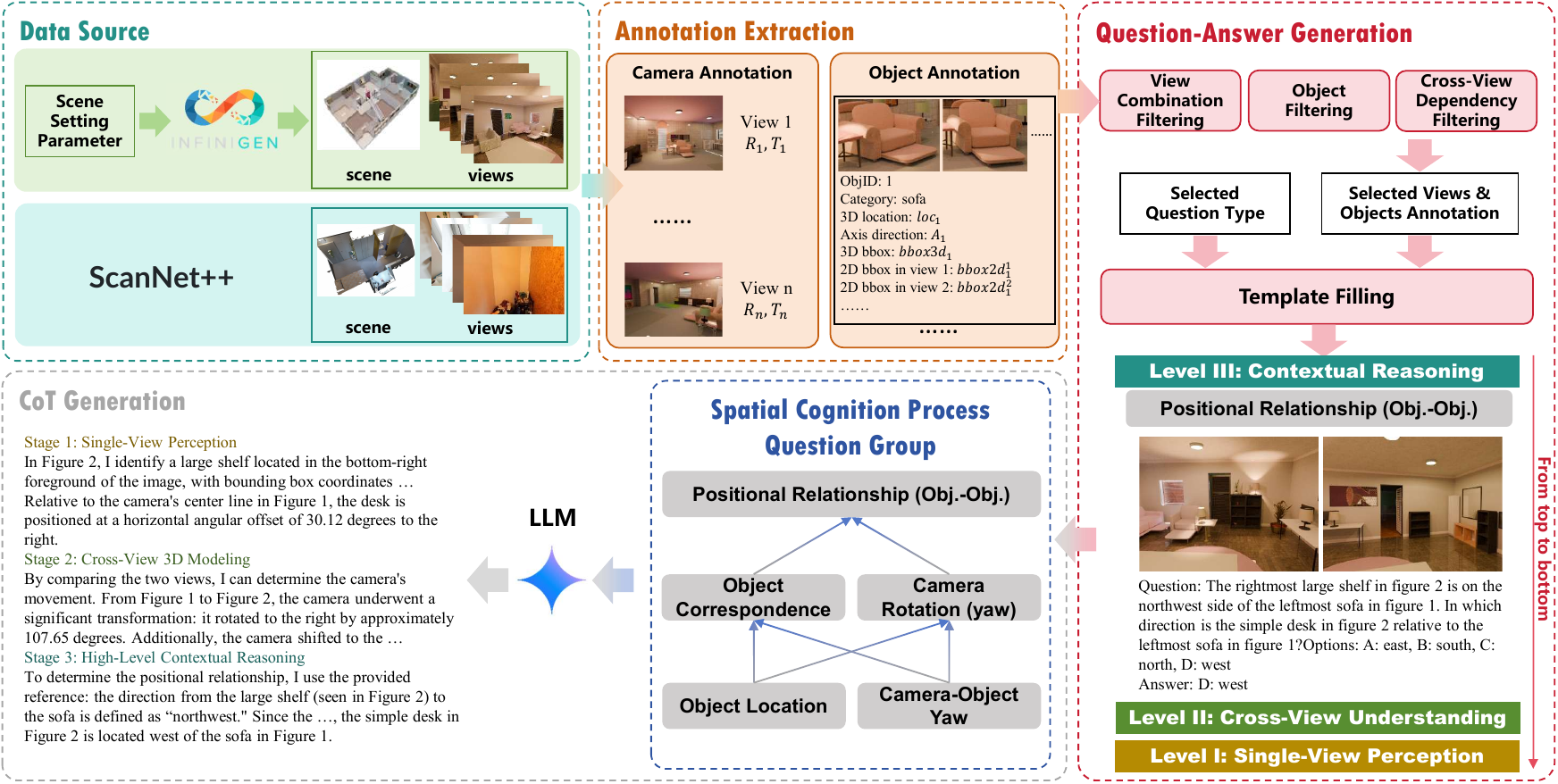}
  \caption{Overview of MV-STRIDE Construction Pipeline. We convert 3D scene assets into hierarchical multi-view spatial reasoning tasks. Scene annotations are first extracted from synthetic and real-world scene datasets. A filtering stage selects valid cameras and objects while enforcing cross-view dependency constraints. Based on the filtered metadata, hierarchical QA pairs are generated across three spatial cognition levels. Finally, cognition-process-based question groups are used to synthesize chain-of-thought reasoning annotations for high-level tasks.}
  \label{fig:pipeline}
\end{figure}

To address the bottleneck of systematic multi-view spatial reasoning capability in existing MLLMs, we propose a hierarchical, cognition-driven dataset construction methodology tailored for 3D-consistent multi-view spatial reasoning. Our method is grounded in a key observation: robust spatial reasoning does not emerge from isolated tasks, but from a progressive acquisition of spatial perception, scene-level understanding, and high-level contextual reasoning abilities. Consequently, we organize spatial reasoning tasks following a "Perception, Modeling, Reasoning" cognitive hierarchy and construct a layered question-answering (QA) dataset that explicitly models capability dependencies across levels. 

In the following, Section 3.1 introduces the hierarchical definition of spatial reasoning tasks and their capability dependencies. Section 3.2 describes the data generation pipeline used to construct the multi-level QA dataset. Section 3.3 presents the multi-stage training strategy built upon the proposed dataset.

\subsection{Hierarchical Definition of Spatial Reasoning Capabilities}

We decompose multi-view spatial reasoning into three tightly coupled levels, each corresponding to a distinct stage in human spatial cognition and addressing a specific limitation of current MLLMs. These levels jointly form a unified spatial reasoning training system. The hierarchical architecture and task composition of MV-STRIDE are illustrated in \cref{fig:teaser}.
\subsubsection{Level I: Single-View Spatial Perception.}
Level I focuses on the establishment of robust, egocentric spatial perception within single perspective. As the bedrock of the MV-STRIDE hierarchy, this level transitions the model's focus to quantitative geometric awareness. Specifically, the task types include recognition of the spatial attributes of objects and cameras, and relative position judgment.
By ensuring precise perception at the single-view level, we prevent hallucinated geometry from propagating into subsequent 3D scene understanding stages. From a cognitive perspective, it mimics the human ability to instantly perceive the layout of a visible environment, providing the necessary spatial clues to build a global scene representation. 
\subsubsection{Level II: 3D-Consistent Scene Understanding.}
Level II focuses on scene-level spatial understanding by establishing geometric correspondences and viewpoint transformations across two perspectives. The primary objective is to bridge isolated perspectives into a coherent 3D cognitive map. Tasks at this level require MLLM to resolve inter-view relationships, including cross-view object correspondence and relative camera motion estimation (covering translation in horizontal planar, yaw, and pitch). By mastering these capabilities, the model is encouraged to transition from independent image processing to a global modeling of the scene structure, effectively overcoming spatial inconsistency and providing the necessary structural foundation for subsequent complex reasoning.
\subsubsection{Level III: Multi-View High-Level Contextual Reasoning.}
Level III focuses on high-order, context-dependent spatial reasoning, representing the most comprehensive task suite for multi-view spatial reasoning to date. Building upon the structural representations established in Level II, this level requires the model to perform sophisticated inference across a diverse set of tasks, including multi-view object counting, measurement estimation, orientation changing and complex relative positioning across multiple reference frames. To achieve high-level proficiency, these tasks are designed to necessitate virtual viewpoint transformations and allocentric reference frame reasoning based on full 6-DoF camera states, pushing the model’s complexity well beyond simple ego-centric observation.

Crucially, to prevent the model from relying on single-view shortcuts, we enforce a strict cross-view dependency constraint: every question is designed to be unsolvable from any single perspective. For example, the model may be asked to determine the spatial direction of an object visible only in View B relative to the camera's position in View A. This design encourages the model to perform accurate perception and comprehensive cross-view information integration before reasoning, guiding it to develop genuine 3D-aware spatial reasoning ability instead of read 2D information. 
\subsubsection{Spatial Cognition Process Question Group.}
Based on the hierarchical definition of spatial reasoning tasks, we further introduce a cognition-process-based grouping strategy to explicitly model the reasoning trajectory required for complex tasks. In this approach, we treat the high-level multi-view contextual reasoning problems in Level III as the apex. For each complex problem, we perform a top-down decomposition of the underlying spatial cognitive requirements, reverse-engineering the necessary perceptual and scene-understanding steps. These intermediate capabilities are then matched and instantiated as specific sub-questions within Level I and Level II. Consequently, for a given scene and viewpoint set, questions across all three levels are organized into a coherent spatial cognition process question group.

By explicitly encoding these capability dependencies and logical transitions, this design ensures that high-level reasoning is strictly grounded in verifiable perceptual and scene-level understanding, thereby guiding models to follow a cognitively plausible reasoning trajectory that transitions from local perception to global integration. Furthermore, this structured organization allows for the construction of principled, long-chain reasoning supervision, providing an efficient pathway to acquire stable 3D spatial awareness and robust multi-view reasoning behavior.
\subsection{Data Construction Pipeline}
We develop a systematic, cognition-driven pipeline to transform raw 3D scene data into structured, hierarchical spatial reasoning tasks, as illustrated in \cref{fig:pipeline}. By grounding natural-language reasoning in verifiable 3D metadata, this framework bridges the gap between low-level visual perception and high-level contextual inference.
\subsubsection{Scene Assets and Preprocessing.}
Our pipeline leverages two distinct types of data sources to ensure both realism and structural diversity: procedural 3D scene generators and real-world 3D reconstruction datasets.

\textbf{Procedural Synthesis.} We utilize the Infinigen\cite{infinigen2023infinite} framework to procedurally generate a vast array of novel indoor environments. This allows us to create diverse room layouts, varied object distributions, and randomized viewpoint trajectories. For each generated scene, we render multi-view images while simultaneously capturing exhaustive ground-truth metadata, including 3D object bounding boxes, precise camera intrinsics/extrinsics, and view-dependent 2D instance segmentations.

\textbf{Real-World Reconstruction.} To ground our dataset in real-world complexity, we incorporate ScanNet++\cite{yeshwanth2023scannet++}, which provides high-fidelity 3D reconstructions of actual indoor spaces. From this source, we directly extract the requisite geometric and semantic information, ensuring that our model encounters realistic textures, lighting, and spatial configurations.

By integrating these dual sources, we establish a robust foundation of accurate 3D object and camera annotations. This geometric grounding is essential for maintaining strict multi-view consistency and generating high-precision QA pairs across our hierarchical framework.
\subsubsection{Viewpoint and Object Filtering.}
To ensure the dataset effectively facilitates 3D-aware learning, we implement a rigorous multi-stage filtering process:

\textbf{Camera Pair Selection.} We filter camera combinations based on their spatial proximity and orientation. Pairs with excessively large or negligible spatial separation are excluded to maintain a reasonable level of difficulty. Crucially, we require each camera pair to share a minimum number of common objects, providing sufficient spatial anchors to ensure that cross-view relationships are solvable for observers.

\textbf{Object Visibility and Ambiguity Control.} We select target objects that are clearly visible and can be uniquely identified through unambiguous text descriptions. Objects that are heavily occluded or too small to provide reliable visual cues are discarded.

\textbf{Cross-View Dependency Constraint.} To uphold the single-view unsolvability of Level III tasks, we apply deterministic rules during selection. Specifically, we only retain questions that require integrating information across the entire view set, such as referencing an object visible only in View A from the camera coordinate system of View B, or evaluating spatial relationships between multiple objects that never co-occur in a single frame. This eliminates shortcuts based on local cues, compelling the model to synthesize a globally consistent scene representation to arrive at the correct answer.

To empirically validate this constraint, we randomly sample 1K Level III questions and evaluate Gemini-3-Flash under two input settings. When provided with only a random single view, the model obtains 31.5\% accuracy, while using the full multi-view inputs improves the accuracy to 41.2\%. This gap indicates that Level III questions require cross-view information integration and cannot be reliably solved from a single-view shortcut.
\subsubsection{Hierarchical QA Generation.}
Following the cognition-process-based grouping strategy, the generation proceeds in a top-down manner. We first generate complex Level III questions based on the filtered viewpoints and selected objects. Once a Level III task is defined, the pipeline automatically identifies the requisite perceptual and scene understanding capabilities and generates the corresponding sub-questions of Level I and Level II.
To bridge the gap between geometric data and natural language, our pipeline first transforms the raw 3D metadata of filtered objects and viewpoints into structured spatial semantic information, which is then used to populate our predefined QA templates. To ensure high response quality and linguistic diversity, we utilize human-designed templates enriched by Large Language Models to achieve rich variability. While Level I and Level II tasks are formulated as short-answer questions to foster perceptual precision, Level III tasks are structured as multiple-choice questions to facilitate the learning of complex, multi-step spatial inference. This hierarchical design ensures that all tasks, ranging from low-level perception to high-level reasoning, are derived from a unified 3D scene representation and the same set of multi-view visual inputs, maintaining strict cross-level geometric consistency.

\subsubsection{Construction of Long-Chain Reasoning Annotations.}
Leveraging the cognition-process-based question groups, we construct high-quality CoT annotations for Level III tasks. For each high-level problem at the group's apex, we treat its associated Level I and Level II sub-questions and their ground-truth answers as necessary prerequisites. These components are sequentially organized and fed into an LLM\cite{google_gemini_3}, which serves as a linguistic synthesizer to refine them into a coherent natural-language reasoning chain.

This strategy ensures that each CoT follows the three essential stages of human spatial cognition: local perception, cross-view integration, and high-level contextual inference. By grounding the final answer in verifiable intermediate steps, our approach provides transparent and interpretable supervision for long-chain reasoning. This enables the scalable generation of complex spatial reasoning chains without manual annotation, that remains strictly aligned with the underlying cognition logic, as further detailed in the Appendix.

\subsubsection{Quality Verification.}
We further conduct sampled quality verification by human and LLM for the generated QA pairs and CoT annotations, yielding an overall accuracy of 88.0\% for QA pairs and 97.5\% for CoT annotations. These results indicate that MV-STRIDE maintains reliable automatical annotation quality at scale, with only minor noise. More details of the verification protocol and error analysis are provided in the Appendix.

\subsection{Multi-Stage Training Strategy.}
To fully exploit the hierarchical and grouped nature of our dataset, we design a progressive three-stage training strategy that systematically develops the model’s spatial intelligence from fundamental perception to complex reasoning.

\textbf{Stage 1: Foundational Spatial SFT.} In the first stage, we perform Supervised Fine-Tuning (SFT) on the base model using 60\% of the scene data, encompassing tasks from Level I to Level III. This stage is designed to equip the model with foundational capabilities in spatial perception, cross-view scene understanding, and multi-view contextual reasoning.

\textbf{Stage 2: Cold-Start with CoT.} While the first stage establishes core capabilities, it does not inherently enable the model to generate interpretable reasoning chains. Therefore, we utilize 20\% of the scenes from ScanNet++ to construct CoT annotations for cold-start training. This stage focuses on teaching the model the structured reasoning templates required for complex multi-view tasks, fostering an initial reasoning awareness.

\textbf{Stage 3: Reinforcement Learning (RL).} In the final stage, we employ the Group Relative Policy Optimization (GRPO)\cite{shao2024deepseekmath} algorithm to further refine the model. We first apply the Stage I trained model to perform multiple inference passes on the remaining data and select a subset of samples with moderate difficulty based on empirical correctness statistics. Using these samples, the model then performs reinforcement learning to further explore diverse reasoning paths, building upon the reasoning templates acquired during the cold-start stage. This reinforcement learning process encourages the model to find an optimal balance between interpretability and accuracy, ultimately achieving robust 3D-aware spatial reasoning.

\section{Experiments}
\subsection{Experimental Setup}
\subsubsection{Implementation Details.}
We use Qwen3-VL-8B-Instruct\cite{bai2025qwen3} as the base model for all experiments. Training is conducted in three stages: large-scale SFT (Stage 1), CoT cold-start training (Stage 2), and RL (Stage 3).

In Stage 1, the model is trained with a learning rate of $1 \times 10^{-5}$ on a diverse mixture of datasets
including the combined set of Levels I–III from 60\% of the scenes, comprising 313.8K QA pairs, additional open-source spatial reasoning and GQA datasets\cite{zhang2025flatland,zhou2025roborefer,ray2024sat,ma2025spatialreasoner,jung2025right,acharya2019tallyqa,deitke2025molmo,li2024llava,fan2025vlm}. 
In Stage 2, we perform cold-start training on 19.0K CoT annotated Level III samples with a learning rate of 1e-6.
In Stage 3, we employ GRPO algorithm for reinforcement learning on 24.8K high-level reasoning samples, using a learning rate of 1e-6, with 8 rollouts per prompt.

For Stage 1 and Stage 2, the global batch size is set to 256, with a warmup ratio of 0.1. In Stage 3, a warmup ratio of 0.01 is used. All remaining training details, including dataset composition ratios and optimization hyperparameters, are provided in the Appendix.
\subsubsection{Benchmarks. }To evaluate the effectiveness of our proposed method in spatial reasoning, particularly under multi-view conditions, we conduct extensive experiments across several representative benchmarks. These include CV-Bench\cite{tong2024cambrian} for foundational single-view spatial perception, MMSI-Bench\cite{yang2025mmsi} and ViewSpatial-Bench\cite{li2025viewspatial} for multi-viewpoint spatial localization and reasoning, and 3DSRBench\cite{ma20243dsrbench} for comprehensive 3D-aware spatial inference. This diverse selection of benchmarks allows for a holistic assessment of our model's performance improvements across different levels of spatial complexity.

\subsubsection{Baselines.}
We evaluate our method against a diverse range of MLLMs, including proprietary MLLMs (\eg, GPT-5\cite{singh2025openai}), representative open-source models across various parameter scales (\eg, InternVL3-38B\cite{zhu2025internvl3}), and specialized spatial reasoning MLLMs(\eg, SPAR\cite{zhang2025flatland}). For baselines without publicly available scores on specific benchmarks, we evaluate their performance under the same experimental conditions as our own model to ensure a fair and rigorous comparison.
\subsection{Main Results}
\subsubsection{Quantitative Results on Spatial Reasoning Benchmarks.}
\begin{table}[ht]
  \centering
  \caption{\textbf{Performance of different models across four spatial reasoning benchmarks.} The best overall results in each column are \textbf{bolded}, while the second-best overall results are \textit{italicized}. The best results among open-source models are \underline{underlined}.}
  \label{tab:main_results}
  \begin{adjustbox}{width=\textwidth}
  \setlength{\tabcolsep}{2pt} 
  \begin{small}
  \begin{tabular}{l ccc cccc ccccc ccc}
    \toprule
    \multirow{3}{*}{\textbf{Model}} & \multicolumn{12}{c}{\textbf{MMSI-Bench}} & \multirow{3}{*}{\textbf{CV}} & \multirow{3}{*}{\textbf{VS}} & \multirow{3}{*}{\textbf{3DSR}} \\
    \cmidrule(lr){2-13}
    & \multicolumn{2}{c}{Attribute} & \multicolumn{2}{c}{Motion (M)} & \multicolumn{6}{c}{Positional Relationship (PR)} & \multirow{2}{*}{MSR} & \multirow{2}{*}{\textbf{Avg.}} & & & \\
    \cmidrule(lr){2-3} \cmidrule(lr){4-5} \cmidrule(lr){6-11}
    & Appr. & Meas. & C & O & C-C & C-O & C-R & O-O & O-R & R-R & & & Avg. & Avg. & Avg. \\
    \midrule
    \rowcolor[gray]{.95} \multicolumn{16}{l}{\textit{Proprietary MLLMs}} \\
    GPT-5\cite{singh2025openai} & 36.40 & \textbf{60.90} & 32.40 & \textit{36.80} & \textit{43.00} & 48.80 & 51.80 & 35.10 & 42.40 & 32.10 & \textbf{42.40} & \textbf{41.90} & - & - & \textbf{66.70} \\
    GPT-4o\cite{hurst2024gpt} & \textit{37.90} & 37.50 & 37.80 & 31.60 & 37.60 & 20.90 & 33.70 & 25.50 & 32.90 & 27.20 & 32.80 & 32.10 & 78.90 & 34.98 & 60.30 \\
    Gemini2.5-Pro\cite{comanici2025gemini} & 33.30 & \textbf{60.90} & 27.00 & 28.90 & \textbf{45.20} & 38.40 & 45.80 & 35.10 & 41.20 & 32.10 & 33.30 & 37.60 & - & - & 64.30 \\
    Seed1.5-VL\cite{guo2025seed1} & 21.20 & 35.90 & 37.80 & 27.60 & 27.60 & 20.90 & 28.90 & 30.90 & 40.00 & 27.20 & 30.80 & 30.70 & - & - & 64.00 \\
    \midrule
    \rowcolor[gray]{.95} \multicolumn{16}{l}{\textit{Open-source MLLMs}} \\
    InternVL3-78B\cite{zhu2025internvl3} & 21.20 & 40.60 & 33.80 & 31.60 & 33.30 & 18.60 & 24.10 & 22.30 & 36.50 & 32.10 & 27.80 & 28.90 & - & - & 61.30 \\
    InternVL2.5-78B\cite{chen2024expanding} & 22.70 & 35.90 & 20.30 & 34.20 & 35.50 & 30.20 & 42.20 & 20.20 & 40.00 & \textit{38.30} & 29.30 & 31.50 & - & - & 58.00 \\
    Qwen2-VL-72B-Instruct\cite{wang2024qwen2} & 27.30 & 42.20 & 33.80 & 25.00 & 21.50 & 24.40 & 27.70 & 31.90 & 41.20 & \textit{38.30} & 33.30 & 31.50 & - & - & 57.50 \\
    LLaVA-OneVision-72B\cite{li2024llava} & 18.20 & 32.80 & 16.20 & 28.90 & 40.90 & 30.20 & 30.10 & 25.50 & 35.30 & 32.10 & 16.70 & 26.90 & - & - & 57.90 \\
    InternVL3-38B\cite{zhu2025internvl3} & 30.30 & 42.20 & 20.30 & 32.90 & 25.80 & 23.30 & 28.90 & 24.50 & 37.60 & 29.60 & 28.30 & 29.00 & - & - & 59.10 \\
    Ovis2-34B\cite{ovis2} & 31.80 & 39.10 & 20.30 & 32.90 & 31.20 & 26.70 & 21.70 & 27.70 & \textit{44.70} & 32.10 & 31.30 & 30.80 & - & - & 58.60 \\
    Qwen2.5-VL-32B-Instruct\cite{bai2025qwen25vltechnicalreport} & 36.40 & 39.10 & 27.00 & 34.20 & 26.90 & 26.70 & 33.70 & 26.60 & 30.60 & 29.60 & 21.70 & 28.90 & - & - & 56.90 \\
    InternVL3-14B\cite{zhu2025internvl3} & 24.20 & 34.40 & 27.00 & 27.60 & 15.10 & 27.90 & 22.90 & 29.80 & 35.30 & 24.70 & 26.80 & 26.70 & - & 40.28 & 56.00 \\
    InternVL3-8B\cite{zhu2025internvl3} & 22.70 & 32.80 & 17.60 & 31.60 & 21.50 & 30.20 & 37.30 & 27.70 & 43.50 & 30.90 & 30.80 & 29.90 & 85.50 & - & 55.50 \\
    InternVL3.5-8B\cite{wang2025internvl3} & 19.69 & 35.93 & 17.56 & 26.31 & 21.50 & 18.60 & 28.91 & 24.46 & 29.41 & 19.75 & 28.28 & 24.9 & 82.86 & 43.24 & - \\
    MiMo-VL-7B-RL\cite{coreteam2025mimovltechnicalreport} & 24.20 & 25.00 & 28.40 & 23.70 & 25.50 & 22.10 & 27.70 & 29.80 & \underline{\textbf{45.90}} & 30.90 & 32.80 & 30.30 & 82.30 & - & 60.10 \\
    MiMo-VL-7B-SFT\cite{coreteam2025mimovltechnicalreport} & 19.70 & 37.50 & 17.57 & 23.68 & 27.96 & 25.58 & 21.69 & 34.04 & 31.76 & 35.80 & \underline{\textit{34.34}} & 29.00 & 82.33 & 40.37 & 59.31 \\
    Qwen2.5-VL-7B-Instruct\cite{bai2025qwen25vltechnicalreport} & 21.20 & 34.40 & 23.00 & 31.60 & 24.70 & 29.10 & 28.90 & 21.30 & 35.30 & 23.50 & 25.30 & 26.80 & 73.00 & 36.85 & 53.20 \\
    \midrule
    \rowcolor[gray]{.95} \multicolumn{16}{l}{\textit{Spatial-reasoning MLLMs}} \\
    SPAR\cite{zhang2025flatland} & - & - & - & - & - & - & - & - & - & - & - & - & 79.91 & - & 57.48 \\
    VST-7B-RL\cite{yang2025visual} & 26.60 & \underline{\textit{48.50}} & 33.80 & 31.60 & 35.50 & 38.30 & 30.90 & \underline{\textbf{54.60}} & 35.30 & \underline{\textbf{47.00}} & 22.70 & 35.30 & \textit{86.50} & - & 60.10 \\
    \midrule
    \rowcolor[gray]{.95} \multicolumn{16}{l}{\textit{Ours}} \\
    Base Model(Qwen3-VL-8B-Instruct\cite{bai2025qwen3}) & 19.70 & 39.06 & 24.32 & 26.32 & 19.35 & 27.91 & 37.35 & 29.79 & 38.82 & 28.40 & 29.80 & 29.20 & 84.31 & 40.51 & 59.98 \\
    \textbf{MV-STRIDE-SFT(Full)} & \underline{\textbf{43.94}} & 43.75 & \underline{\textbf{45.95}} & 31.58 & 34.41 & \underline{\textbf{66.28}} & 56.63 & \textit{41.49} & 34.12 & 30.86 & 22.73 & \underline{\textit{38.90}} & \underline{\textbf{86.73}} & 48.35 & 64.51 \\
    \textbf{MV-STRIDE-SFT(Stage 1)} & 27.27 & 29.69 & \textit{40.54} & 34.21 & \underline{\textit{43.00}} & \textit{63.95} & \underline{\textbf{66.27}} & 36.17 & 34.12 & 28.40 & 21.72 & 37.20 & 86.39 & \underline{\textbf{50.28}} & \underline{\textit{65.52}} \\
    \textbf{MV-STRIDE-ColdStart(Stage 2)} & 30.30 & 31.25 & \textit{40.54} & 35.53 & 40.86 & 56.98 & \textit{61.45} & 39.36 & 29.41 & 28.40 & 23.74 & 36.70 & 85.94 & 47.30 & 61.08 \\
    \textbf{MV-STRIDE-RL(Stage 3)} & 31.82 & 32.81 & \textit{40.54} & \underline{\textbf{40.79}} & 39.78 & 62.79 & 59.04 & 38.30 & 34.12 & 27.16 & 22.73 & 37.50 & 85.94 & \textit{49.30} & 59.43 \\
    \bottomrule
  \end{tabular}
  \end{small}
  \end{adjustbox}
\end{table}
In this section, we evaluate the performance of our models across different training phases: MV-STRIDE-SFT(Full), a comprehensive baseline trained by pooling all data from all three stages for SFT; MV-STRIDE-SFT(Stage 1), the model obtained after the SFT stage; MV-STRIDE-ColdStart(Stage 2), the model obtained after the cold start stage; and MV-STRIDE-RL(Stage 3), the final version further refined through RL, comparing them against various MLLMs. 

As illustrated in \cref{tab:main_results}, our models demonstrate superior performance across multiple spatial reasoning benchmarks. Specifically, MV-STRIDE-SFT(Full) achieves state-of-the-art (SOTA) results among open-source MLLMs with an average accuracy of \textbf{38.9\%} on the MMSI-Bench, yielding a substantial \textbf{9.7\%} improvement over the Base Model. This performance surpasses several models with significantly larger parameter scales and remains highly competitive with leading proprietary MLLMs, effectively narrowing the gap between open-source models and closed-source counterparts in complex spatial reasoning.

A more granular analysis of MMSI-Bench sub-categories reveals that our models excel in dimensions highly correlated with our proposed dataset, such as Motion and Positional Relationship. Notably, it attains SOTA performance across all tested models in several challenging sub-tasks, including M-C and PR-C-O\etal. These results provide compelling evidence for the effectiveness of our data in addressing complex multi-view spatial reasoning challenges.

Furthermore, our model exhibits remarkable generalization capabilities. MV-STRIDE-SFT(Stage 1) achieves the overall best performance on ViewSpatial-Bench and the top result among open-source models on 3DSRBench, both of which emphasize 3D and multi-view spatial reasoning. On the single-view CV-Bench, MV-STRIDE-SFT(Full) reaches the best performance among all compared models. Collectively, these empirical findings strongly validate that our proposed dataset significantly augments the model's spatial reasoning capacity, particularly in synthesizing multi-view information for comprehensive 3D spatial understanding.

We further validate the effectiveness of MV-STRIDE on MLLMs with different parameter scales and architectures, and observe consistent improvements across benchmarks.The detailed results are provided in Appendix.
\subsubsection{Qualitative Analysis of 3D-Consistent Reasoning.}
\begin{figure}[tb]
  \centering
  \includegraphics[width=\linewidth]{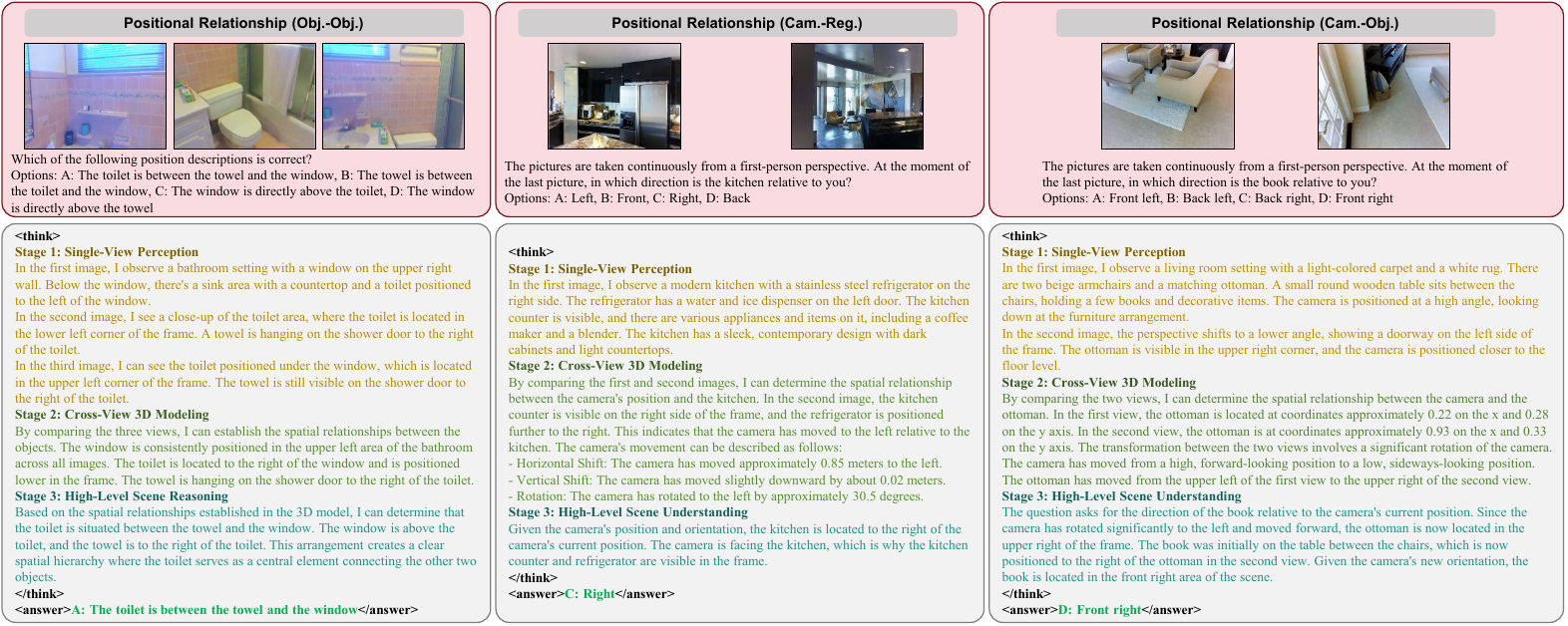}
  \caption{CoT produced by MV-STRIDE-RL on several multi-view spatial reasoning examples from MMSI-Bench.}
  \label{fig:results}
\end{figure}

While the quantitative benchmarks of MV-STRIDE-RL may not always surpass our other variants, it exhibits a significant advantage in generating highly interpretable and structured spatial reasoning chains. As illustrated in \cref{fig:results}, through cold-start fine-tuning and RL, the model has autonomously evolved a multi-stage systematic paradigm for addressing multi-view challenges.

This hierarchical process mirrors the human spatial cognition path: it begins with Single-View Perception, identifying localized assets like kitchen appliances; progresses to Cross-View 3D Modeling, establishing spatial anchors between frames and estimating camera pose (\eg, quantifying horizontal shifts and rotations); and culminates in High-Level Scene Reasoning to derive the final answer.

The samples in \cref{fig:results} demonstrate this qualitative leap: the model doesn't merely predict a label but constructs a consistent internal 3D representation, explicitly calculating the camera's translation (\eg, $0.85$m shift) and rotation (\eg, $30.5^\circ$). This level of mathematical grounding and logical coherence underscores the capacity of reinforcement learning to cultivate internal spatial consistency—providing a "white-box" reasoning process that exceeds the reach of standard supervised learning.
\subsection{Ablation Studies}
\subsubsection{Effectiveness of MV-STRIDE.}
\begin{figure}[tb]
  \centering
  \begin{subfigure}{0.48\linewidth}
    \centering
    \includegraphics[width=\linewidth]{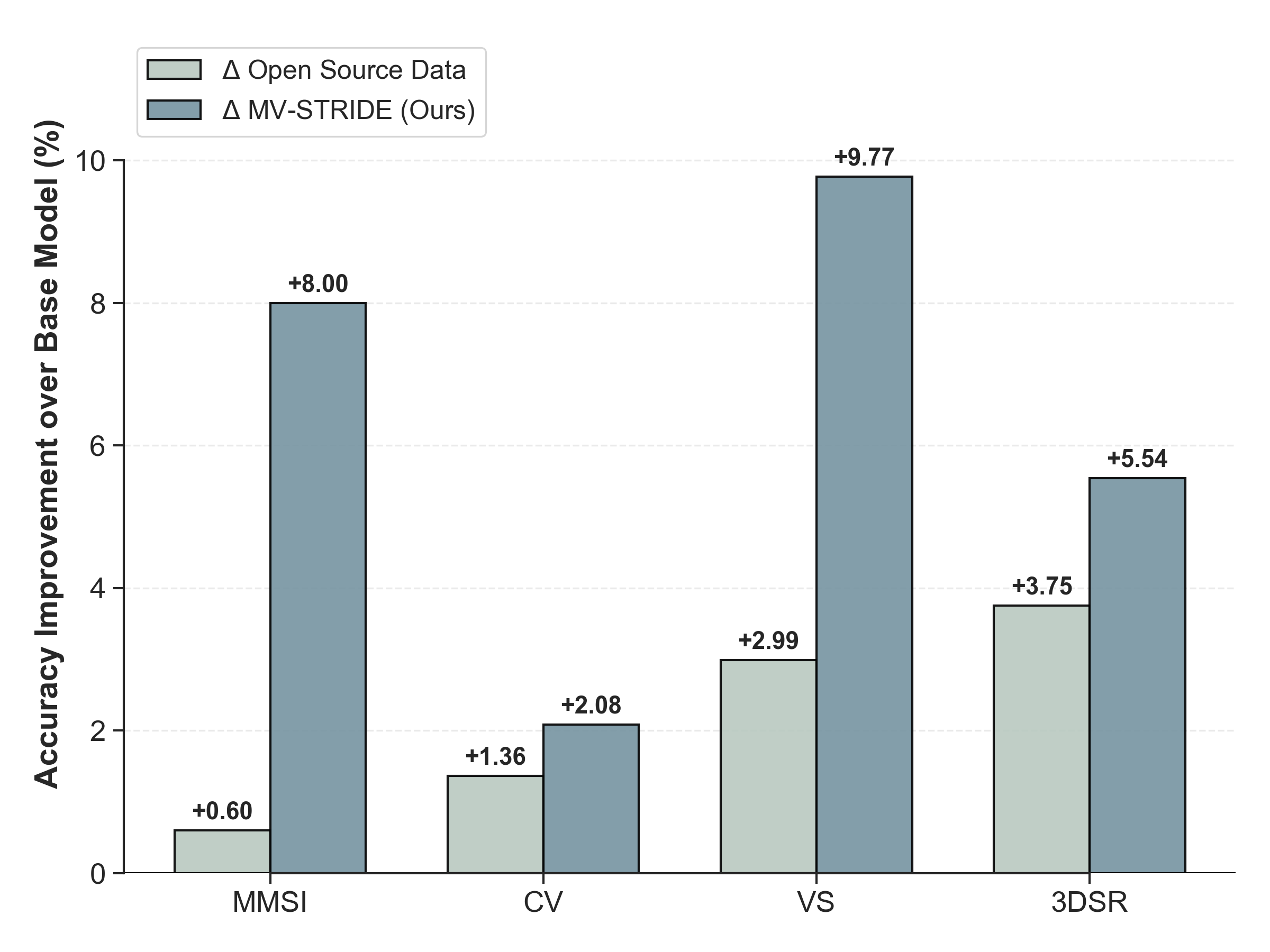}
    \caption{Impact of data composition in Stage 1.}
    \label{fig:ablation-a}
  \end{subfigure}
  \hfill 
  \begin{subfigure}{0.48\linewidth}
    \centering
    \includegraphics[width=\linewidth]{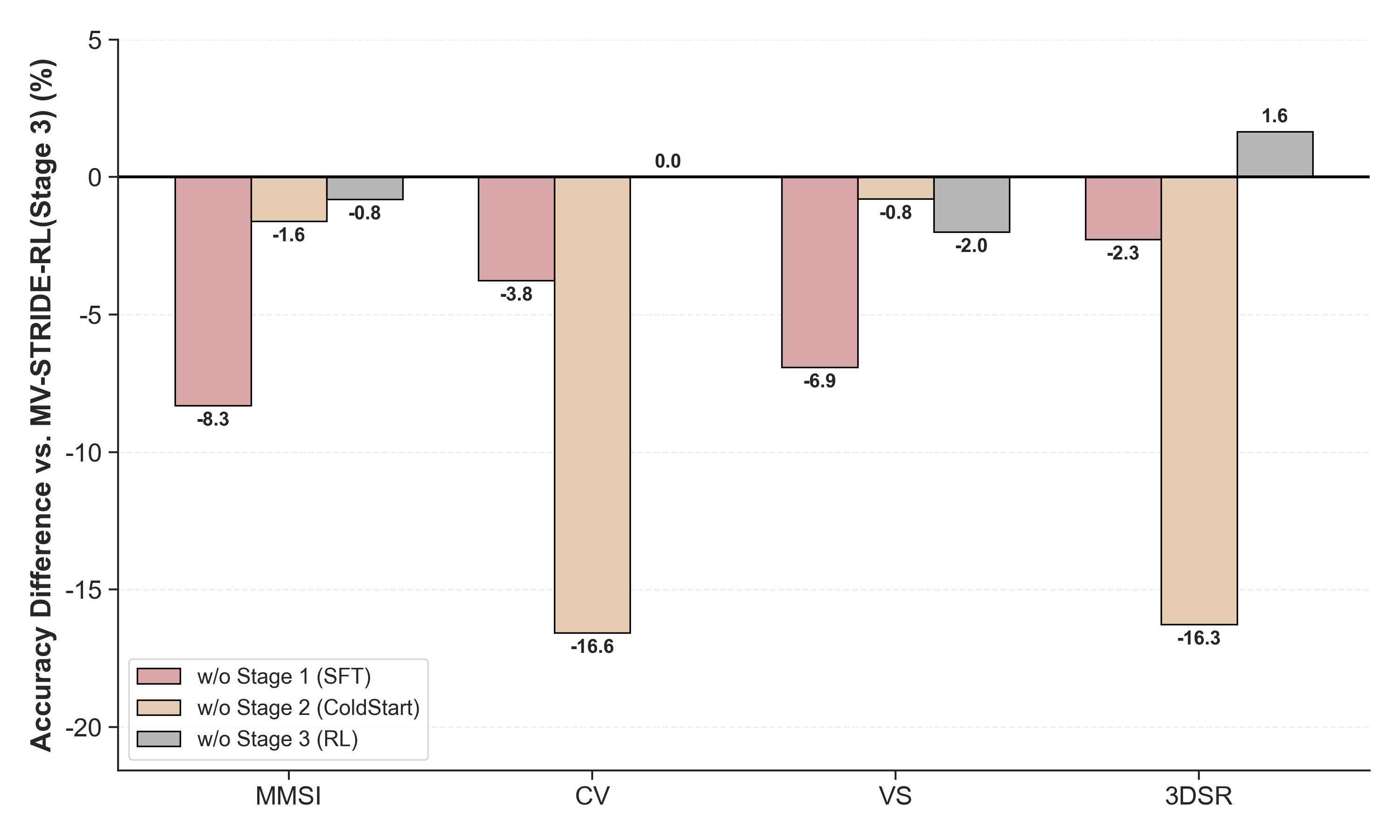}
    \caption{Ablation on multi-stage training strategies.}
    \label{fig:ablation-b}
  \end{subfigure}
  \caption{Ablation study results. Results are reported in percentage points (pp.) relative to corresponding baselines.}
  \label{fig:ablation}
\end{figure}
To investigate the specific contribution of our proposed dataset to the model's spatial reasoning capabilities, we conduct a series of ablation experiments on the training data composition in Stage 1. As established in our training pipeline, we introduce various open-source datasets including spatial reasoning, caption and GQA data as auxiliary data to fortify the model's fundamental perception. However, to demonstrate that MV-STRIDE is the primary driver for the leap in spatial reasoning—especially in multi-view scenarios—we compare the performance of models trained with different data combinations.

As summarized in \cref{fig:ablation-a}, incorporating only open-source auxiliary data yields marginal improvements over the base model across most benchmarks, with almost negligible gains on the multi-view-focused MMSI-Bench. In stark contrast, the inclusion of MV-STRIDE leads to a substantial performance surge. Specifically, accuracy on MMSI-Bench and ViewSpatial experiences a significant leap, far outstripping the improvements brought by general open-source data. Furthermore, consistent and marked enhancements are observed across the other two benchmarks. These empirical results provide robust evidence that our proposed dataset is the essential catalyst for empowering the model to synthesize complex multi-view information and achieve superior 3D spatial understanding.
\subsubsection{Effect of Progressive Multi-Stage Training.}
To verify the necessity of each phase in our proposed pipeline, we evaluate models at different training stages and perform a ablation study on every stage. The results are summarized in \cref{fig:ablation-b} and \cref{tab:main_results}.

\textbf{Necessity of Individual Stages.} 
We conduct an ablation study by systematically removing stages, with \cref{fig:ablation-b} illustrating the performance delta relative to MV-STRIDE-RL (Stage 3). Without Stage 1's foundational spatial knowledge, the model lacks a viable exploration space for reinforcement learning, leading to convergence failure on high-order tasks. Omitting the Stage 2 cold-start prevents the internalization of structured reasoning; during inference, these models suffer significant degradation on CV-Bench and 3DSRBench, proving that RL requires supervised warm-up to develop coherent logic. Crucially, without Stage 3, the cold-start model merely imitates SFT templates through rote memorization. In contrast, the RL phase empowers the model to autonomously verify spatial logic and perspective consistency, facilitating a transition from passive pattern matching to active, generalized problem-solving (see Appendix for qualitative comparisons).

\textbf{Progressive Integration and the Reasoning Tax.} Comparing the stages of progressive training in \cref{tab:main_results}, we observe a slight decline in raw accuracy on certain benchmarks after transitioning from Stage 1 to Stage 2 and Stage 3. 
During the transition from Stage 1 to Stage 2, the model changes from the short-answer format used in foundational SFT to CoT-style supervision. This shift alters the output distribution and may reduce direct-answer accuracy on some benchmarks, which we refer to as a "reasoning tax". From Stage 2 to Stage 3, the effect becomes benchmark- and task-dependent: multi-view-oriented benchmarks such as MMSI-Bench and ViewSpatial-Bench benefit more from RL training, whereas single-view or more general 3D reasoning benchmarks may remain stable or slightly decrease due to the focus of Stage 3 on multi-view high-level reasoning. Despite these fluctuations, the multi-stage training process enables the model to internalize a more systematic reasoning pattern, improving interpretability and supporting cross-view spatial reasoning.

\textbf{Multi-Stage Strategy vs. All-in-one SFT.} We compare our multi-stage strategy with MV-STRIDE-SFT (Full), which pools all multi-stage data for a single-pass SFT in \cref{tab:main_results}. While MV-STRIDE-SFT (Full) achieves higher accuracy on certain benchmarks such as MMSI-Bench, this is largely due to its stronger pattern-matching capability from seeing all CoT data as direct supervision. However, our multi-stage approach remains competitive and even superior in specific 3D-heavy contexts. More importantly, our approach ensures that the model's reasoning is actively derived through RL optimization rather than just passively imitated. This leads to a model that is more robust to novel, out-of-distribution spatial scenarios where simple rote memorization of SFT data fails.
\subsubsection{Impact of Synthetic vs. Real-world Data Sources.}
\begin{table}[ht]
  \centering
  \caption{\textbf{Ablation study on data sources for Level III spatial reasoning.} Performance is evaluated on MMSI-Bench (Acc.). "\# Samples" indicates the volume of Level 3 data used for SFT.}
  \label{tab:ablation_datasource}
  \begin{tabular}{lccc}
    \toprule
    \textbf{Data Source} & \textbf{\# Infinigen} & \textbf{\# ScanNet++} & \textbf{MMSI-Bench} \\
    \midrule
    Base Model & - & - & 29.20 \\
    Synthetic Only & 99.3k & - & 36.70 \\
    Real-world Only & - & 99.5k & 37.50 \\
    Hybrid & 49.7k & 49.8k & 38.00 \\
    Unified & 99.3k & 99.5k & \textbf{39.40} \\
    \bottomrule
  \end{tabular}
\end{table}
This section investigates the complementarity of synthetic and real-world data sources for training spatial reasoning models. As illustrated in Table~\cref{tab:ablation_datasource}, we conduct supervised fine-tuning using equal quantities of Level 3 spatial reasoning data from different sources to evaluate their respective impacts.

The experimental results reveal that although synthetic data sources like Infinigen\cite{infinigen2023infinite} offer exceptional generation flexibility and theoretical reproducibility, their diversity remains limited by the underlying procedural generation algorithms. Additionally, the fidelity gap compared to real-world scenes causes the performance gains from synthetic data to be slightly lower than those from real-world sources such as ScanNet++\cite{yeshwanth2023scannet++}. Nevertheless, the hybrid use of both synthetic and real-world data yields the optimal performance. This synergy stems from the fact that synthetic data provides engine-generated ground truth with absolute precision, effectively helping the model establish rigorous spatial geometric concepts. Concurrently, real-world data accounts for the complexity of visual distributions, ensuring that the acquired spatial reasoning capabilities generalize effectively to real-world inputs. These findings demonstrate the necessity of combining controllable simulation with high-fidelity real-world data for the development of large-scale spatial reasoning datasets.

\section{Conclusion}
This paper presents MV-STRIDE, a hierarchical dataset and training framework designed to bridge the gap in multi-view spatial reasoning for MLLMs. By explicitly modeling capability dependencies—from foundational perception to complex 3D reasoning—we provide a cognitively grounded learning pathway that significantly enhances spatial inference compared to conventional "flat" datasets. Our evaluations demonstrate that while supervised fine-tuning establishes a robust performance baseline, the incorporation of reinforcement learning is essential for cultivating internal spatial consistency. Notably, we find that RL-refined models autonomously develop structured, interpretable reasoning chains, moving beyond rote memorization toward genuine 3D spatial logic. Furthermore, our results underscore that the synergy between precise synthetic grounding and real-world visual complexity is critical for balancing geometric accuracy with robustness. Ultimately, MV-STRIDE offers a systematic roadmap for advancing MLLMs from isolated 2D perception toward holistic, multi-view consistent spatial cognition.



%
%
\bibliographystyle{splncs04}
\bibliography{main}


\clearpage
\appendix

\renewcommand{\thesection}{\Alph{section}}
\renewcommand{\thefigure}{\thesection.\arabic{figure}}
\renewcommand{\thetable}{\thesection.\arabic{table}}
\renewcommand{\theequation}{\thesection.\arabic{equation}}
\setcounter{section}{0}
\setcounter{figure}{0}
\setcounter{table}{0}
\setcounter{equation}{0}

\begin{center}
    {\LARGE \textbf{Supplementary Material for MV-STRIDE}} \\
\end{center}

\section*{Table of Contents (Supplementary Material)}
\noindent
\textbf{A \quad Implementation and Training Details} \dotfill \pageref{sec:impl} \\
\textbf{B \quad Extended Experimental Results} \dotfill \pageref{sec:results} \\
\textbf{C \quad Dataset Construction and Specifications} \dotfill \pageref{sec:data_construct} \\
\textbf{D \quad Dataset Gallery} \dotfill \pageref{sec:gallery} \\
\textbf{E \quad Hierarchical CoT and RL Data Construction Strategy} \dotfill \pageref{sec:cot_rl} \\

\noindent\rule{\textwidth}{0.4pt}

\section{Implementation and Training Details}
\label{sec:impl}
\subsection{Hyperparameters}
The detailed hyperparameter settings are as shown in \cref{tab:hyperparams_stage_rows}.
\begin{table}[ht]
    \centering
    \caption{Hyperparameters of the three training stages.}
    \label{tab:hyperparams_stage_rows}
    \renewcommand{\arraystretch}{1.4} 
    \begin{adjustbox}{width=\textwidth}
    \begin{tabular}{l|cccccccc} 
    \toprule
    \textbf{Stage} & \textbf{Optimizer} & \textbf{LM LR} & \textbf{ViT LR} & \textbf{Packing} & \textbf{Global Batch} & \textbf{Warmup} & \textbf{Scheduler} & \textbf{Duration} \\
    \midrule
    \textbf{Stage 1: SFT} & AdamW & $1 \times 10^{-5}$ & \textit{Frozen} & 8192 & 256 & 0.1 & Cosine & 1 Epoch\\
    \textbf{Stage 2: Cold-start} & AdamW & $1 \times 10^{-6}$ & \textit{Frozen} & 8192 & 256 & 0.05 & Cosine & 1 Epoch \\
    \textbf{Stage 3: RL} & AdamW & $1 \times 10^{-6}$ & \textit{Frozen} & 8192 & 256 & 0.01 & Cosine & 1 Epoch  \\
    \bottomrule
    \end{tabular}
    \end{adjustbox}
\end{table}
\noindent \textbf{Stage 3: Reinforcement Learning Details.} 
For the final stage of MV-STRIDE, we employ GRPO algorithm to refine the model's long-chain reasoning ability for multi-view spatial reasoning task. For each spatial query, we set the group size to $G=8$ samples. The sampling temperature is set to 1.0 to ensure sufficient exploration of the spatial logic space. To maintain stability and prevent the policy from drifting too far from the cold-start model, we set the coefficient of the KL penalty term to 0.001. The policy is optimized using a multi-faceted reward function:
\begin{itemize}
    \item \textbf{Accuracy Reward:} A binary reward is assigned based on the terminal prediction. Specifically, the system extracts the predicted option from within the \texttt{<answer></answer>} tags and compares it directly with the ground-truth (GT) label. The reward is set to 1.0 if the predicted option matches the GT, and 0.0 otherwise.
    \item \textbf{Format Reward:} A dense reward designed to enforce structural consistency. The model is rewarded for correctly encapsulating its entire reasoning process within \texttt{<think></think>} tags and isolating the final answer within \texttt{<answer></answer>} tags. This ensures the output adheres to the predefined CoT format, facilitating automated parsing and logic extraction.
    \item \textbf{Penalty Terms:} To prevent degenerated outputs, we implement a repetition penalty based on $n$-gram redundancy and a soft length constraint\cite{yu2025dapo}. Specifically, we set $n=6$ for $n$-gram, $soft\_max\_length=4096$ and $soft\_cache\_length=1024$ to penalize excessively verbose or repetitive reasoning chains without hard-truncating valid logic.
\end{itemize}
\subsection{Inference Setting} 
Regarding the inference strategy, we employ different prompting techniques tailored to each training stage and model type. Specifically, for most other baseline models and our Stage 1 SFT model, we use standard direct-answering prompts. In contrast, for our models involving the cold-start or RL stage, as well as other reasoning-capable models (e.g., MiMo-VL-7B-RL), we consistently utilize CoT prompting. This ensures that the models' capacity for long-chain spatial reasoning and cross-view consistency is fully elicited and evaluated. The detailed CoT prompt is specified as follows:
\begin{PromptBox}{Inference Prompt for Reasoning Triggering}
Output your step-by-step thinking process in <think> </think> tags and the final choice (e.g., A: option) in <answer> </answer> tags.
\end{PromptBox}

\section{Extended Experimental Results}
\subsection{Detailed Results on Other Benchmarks} 
In addition to the per-category results on the MMSI-Bench presented in \cref{tab:main_results}, we provide detailed per-category performance metrics for the other three benchmarks in \cref{tab:cv_detailed,tab:vs_detailed,tab:3dsr_detailed}.
\label{sec:results}

\begin{table}[ht]
  \centering
  \caption{\textbf{Detailed performance of different models on CV-Bench.}}
  \label{tab:cv_detailed}
  \begin{adjustbox}{width=\textwidth}
  \small
  \begin{tabular}{l cccc c}
    \toprule
    \multirow{2}{*}{\textbf{Model}} & \multicolumn{2}{c}{\textbf{2D Metrics}} & \multicolumn{2}{c}{\textbf{3D Metrics}} & \multirow{2}{*}{\textbf{Avg.}} \\
    \cmidrule(lr){2-3} \cmidrule(lr){4-5}
    & Counting & Spatial Relation & Depth & Distance & \\
    \midrule
    \rowcolor[gray]{.95} \multicolumn{6}{l}{\textit{Proprietary MLLMs}} \\
    GPT-4o\cite{hurst2024gpt} & 65.90 & 85.70 & 87.80 & 78.20 & 78.90 \\
    \rowcolor[gray]{.95} \multicolumn{6}{l}{\textit{Open-source MLLMs}} \\
    InternVL3-8B\cite{zhu2025internvl3} & - & - & 86.10 & 86.00 & 85.50 \\
    InternVL3.5-8B\cite{wang2025internvl3} & 73.35 & 86.62 & 86.83 & 84.67 & 82.86 \\
    MiMo-VL-7B-RL\cite{coreteam2025mimovltechnicalreport} & - & - & - & - & 82.33 \\
    MiMo-VL-7B-SFT\cite{coreteam2025mimovltechnicalreport} & 65.23 & 90.77 & 86.33 & 87.00 & 82.33 \\
    Qwen2.5-VL-7B-Instruct\cite{bai2025qwen25vltechnicalreport} & 67.44 & 82.15 & 60.17 & 69.00 & 73.00 \\
    \rowcolor[gray]{.95} \multicolumn{6}{l}{\textit{Spatial-reasoning MLLMs}} \\
    SPAR\cite{zhang2025flatland} & - & - & - & - & 79.91 \\
    VST-7B-RL\cite{yang2025visual} & - & - & - & - & 86.5 \\
    \rowcolor[gray]{.95} \multicolumn{6}{l}{\textit{Ours}} \\
    Base Model(Qwen3-VL-8B-Instruct\cite{bai2025qwen3}) & 68.53 & 90.92 & 93.33 & 88.83 & 84.31 \\
    \textbf{MV-STRIDE-SFT(Full)} & 73.86 & 96.31 & 93.17 & 86.63 & 86.73 \\
    \textbf{MV-STRIDE-SFT(Stage 1)} & 71.57 & 95.85 & 94.00 & 88.00 & 86.39 \\
    \textbf{MV-STRIDE-ColdStart(Stage 2)} & 71.57 & 95.23 & 93.33 & 87.33 & 85.94 \\
    \textbf{MV-STRIDE-RL(Stage 3)} & 71.32 & 95.28 & 93.17 & 87.67 & 85.94 \\
    \bottomrule
  \end{tabular}
  \end{adjustbox}
\end{table}
\begin{table}[ht]
  \centering
  \caption{\textbf{Detailed performance of different models on the ViewSpatial-Bench.}}
  \label{tab:vs_detailed}
  \begin{adjustbox}{width=\textwidth}
  \small
  \begin{tabular}{l ccccc c}
    \toprule
    \multirow{2}{*}{\textbf{Model}} & \multicolumn{2}{c}{Camera Perspective} & \multicolumn{3}{c}{Person Perspective} & \multirow{2}{*}{\textbf{Avg.}} \\
    \cmidrule(lr){2-3} \cmidrule(lr){4-6}
    & Obj. View Ori. & Rel. Dir. & Obj. View Ori. & Rel. Dir. & Scene Sim. Rel. & \\
    \midrule
    \rowcolor[gray]{.95} \multicolumn{7}{l}{\textit{Proprietary MLLMs}} \\
    GPT-4o\cite{hurst2024gpt} & 19.58 & 41.46 & 42.97 & 40.86 & 26.79 & 34.98 \\
    \rowcolor[gray]{.95} \multicolumn{7}{l}{\textit{Open-source MLLMs}} \\
    InternVL2.5-8B\cite{chen2024expanding} & 41.27 & 49.41 & 46.79 & 42.04 & 32.85 & 43.24 \\
    MiMo-VL-7B-SFT\cite{coreteam2025mimovltechnicalreport} & 31.22 & 50.54 & 48.19 & 42.16 & 23.89 & 40.37 \\
    Qwen2.5-VL-7B-Instruct\cite{bai2025qwen25vltechnicalreport} & 29.32 & 46.64 & 37.05 & 35.04 & 28.78 & 36.85 \\
    \rowcolor[gray]{.95} \multicolumn{7}{l}{\textit{Ours}} \\
    Base Model(Qwen3-VL-8B-Instruct\cite{bai2025qwen3}) & 27.61 & 52.79 & 44.98 & 38.60 & 29.86 & 40.51 \\
    \textbf{MV-STRIDE-SFT(Full)} & 33.13 & 62.55 & 48.09 & 39.43 & 46.33 & 48.35 \\
    \textbf{MV-STRIDE-SFT(Stage 1)} & 33.03 & 64.97 & 53.01 & 44.30 & 44.34 & 50.28 \\
    \textbf{MV-STRIDE-ColdStart(Stage 2)} & 30.22 & 58.77 & 48.09 & 42.40 & 47.33 & 47.30 \\
    \textbf{MV-STRIDE-RL(Stage 3)} & 31.43 & 61.93 & 50.10 & 44.06 & 48.42 & 49.30 \\
    \bottomrule
  \end{tabular}
  \end{adjustbox}
\end{table}
\begin{table}[ht]
  \centering
  \caption{\textbf{Detailed performance of different models on 3DSRBench.}}
  \label{tab:3dsr_detailed}
  \begin{adjustbox}{width=\textwidth}
  \small
  \begin{tabular}{l cccc c}
    \toprule
    \textbf{Model} & Height & Location & Orientation & Multi-Object & \textbf{Avg.} \\
    \midrule
    \rowcolor[gray]{.95} \multicolumn{6}{l}{\textit{Proprietary MLLMs}} \\
    GPT-5\cite{openai_gpt5} & 71.80 & 78.91 & 62.69 & 62.59 & 66.70 \\
    GPT-4o\cite{hurst2024gpt} & 65.90 & 75.72 & 51.02 & 56.70 & 60.30 \\
    Gemini2.5-Pro\cite{comanici2025gemini} & 72.20 & 79.49 & 53.82 & 61.45 & 64.30 \\
    Seed1.5-VL\cite{guo2025seed1} & 62.20 & 81.48 & 53.31 & 60.81 & 64.00 \\
    \midrule
    \rowcolor[gray]{.95} \multicolumn{6}{l}{\textit{Open-source MLLMs}} \\
    InternVL3-78B\cite{zhu2025internvl3} & 63.30 & 78.15 & 51.15 & 57.15 & 61.30 \\
    InternVL2.5-78B\cite{chen2024expanding} & 58.80 & 75.13 & 47.14 & 55.74 & 58.00 \\
    Qwen2-VL-72B-Instruct\cite{wang2024qwen2} & 57.20 & 73.43 & 51.11 & 53.48 & 57.50 \\
    LLaVA-OneVision-72B\cite{li2024llava} & 63.10 & 72.57 & 46.23 & 55.38 & 57.90 \\
    InternVL3-38B\cite{zhu2025internvl3} & 60.60 & 73.74 & 49.52 & 57.27 & 59.10 \\
    Ovis2-34B\cite{ovis2} & 61.80 & 75.35 & 48.20 & 55.51 & 58.60 \\
    Qwen2.5-VL-32B-Instruct\cite{bai2025qwen25vltechnicalreport} & 56.10 & 72.02 & 47.88 & 55.51 & 56.90 \\
    InternVL3-14B\cite{zhu2025internvl3} & 56.90 & 73.33 & 46.38 & 53.50 & 56.00 \\
    InternVL3-8B\cite{zhu2025internvl3} & 55.50 & 71.94 & 47.04 & 52.84 & 55.50 \\
    MiMo-VL-7B-RL\cite{coreteam2025mimovltechnicalreport} & 65.10 & 78.52 & 47.92 & 56.10 & 60.10 \\
    MiMo-VL-7B-SFT\cite{coreteam2025mimovltechnicalreport} & 63.41 & 76.24 & 49.98 & 55.08 & 59.31 \\
    Qwen2.5-VL-7B-Instruct\cite{bai2025qwen25vltechnicalreport} & 52.20 & 67.53 & 46.49 & 51.05 & 53.20 \\
    \midrule
    \rowcolor[gray]{.95} \multicolumn{6}{l}{\textit{Spatial-reasoning MLLMs}} \\
    SPAR\cite{zhang2025flatland} & - & - & - & - & 57.48 \\
    VST-7B-RL\cite{yang2025visual} & - & - & - & - & 60.10 \\
    \midrule
    \rowcolor[gray]{.95} \multicolumn{6}{l}{\textit{Ours}} \\
    Base Model(Qwen3-VL-8B-Instruct\cite{bai2025qwen3}) & 56.23 & 72.21 & 49.71 & 55.52 & 59.98 \\
    \textbf{MV-STRIDE-SFT(Full)} & 55.22 & 77.06 & 57.43 & 60.05 & 64.51 \\
    \textbf{MV-STRIDE-SFT(Stage 1)} & 59.13 & 78.53 & 55.69 & 61.09 & 65.52 \\
    \textbf{MV-STRIDE-ColdStart(Stage 2)} & 49.86 & 70.69 & 54.82 & 59.81 & 61.08 \\
    \textbf{MV-STRIDE-RL(Stage 3)} & 42.61 & 68.87 & 54.54 & 59.75 & 59.43 \\
    \bottomrule
  \end{tabular}
  \end{adjustbox}
\end{table}

\subsection{Ablation of Multiple Stages} 
We present the detailed performance metrics for the model trained at each stage, as well as the ablation studies for each stage, in \cref{tab:ablation_stages} to complement the results shown in \cref{fig:ablation-b}.
\begin{table}[ht]
  \centering
  \caption{Ablation study on the multi-stage training strategy.}
  \label{tab:ablation_stages}
  \begin{tabular}{l cccc}
    \toprule
    \textbf{Configuration} & \textbf{MMSI} & \textbf{CV} & \textbf{VS} & \textbf{3DSR} \\
    \midrule
    Base Model & 29.20 & 84.31 & 40.51 & 59.98 \\
    \midrule
    \rowcolor[gray]{.95} \multicolumn{5}{l}{\textit{Progressive Training Stage}} \\
    MV-STRIDE-SFT (Stage 1) & 37.20 & 86.39 & \textbf{50.28} & \textbf{65.52} \\
    MV-STRIDE-ColdStart (Stage 2) & 36.70 & 85.94 & 47.30 & 61.08 \\
    MV-STRIDE-RL (Stage 3) & 37.50 & 85.94 & 49.30 & 59.43 \\
    \midrule
    \rowcolor[gray]{.95} \multicolumn{5}{l}{\textit{Leave-one-out Verification}} \\
    w/o Stage 1 & 29.20 & 82.18 & 42.38 & 57.17 \\
    w/o Stage 2 & 35.90 & 69.37 & 48.51 & 43.16 \\
    w/o Stage 3 & 36.70 & 85.94 & 47.30 & 61.08 \\
    \midrule
    \rowcolor[gray]{.95} \multicolumn{5}{l}{\textit{Data All-in-one Baseline}} \\
    MV-STRIDE-SFT (Full) & \textbf{38.90} & \textbf{86.73} & 48.35 & 64.51 \\
    \bottomrule
  \end{tabular}
\end{table}
\subsection{Reasoning Chain Comparisons} 
\begin{figure}[tb]
  \centering
  \includegraphics[width=\linewidth]{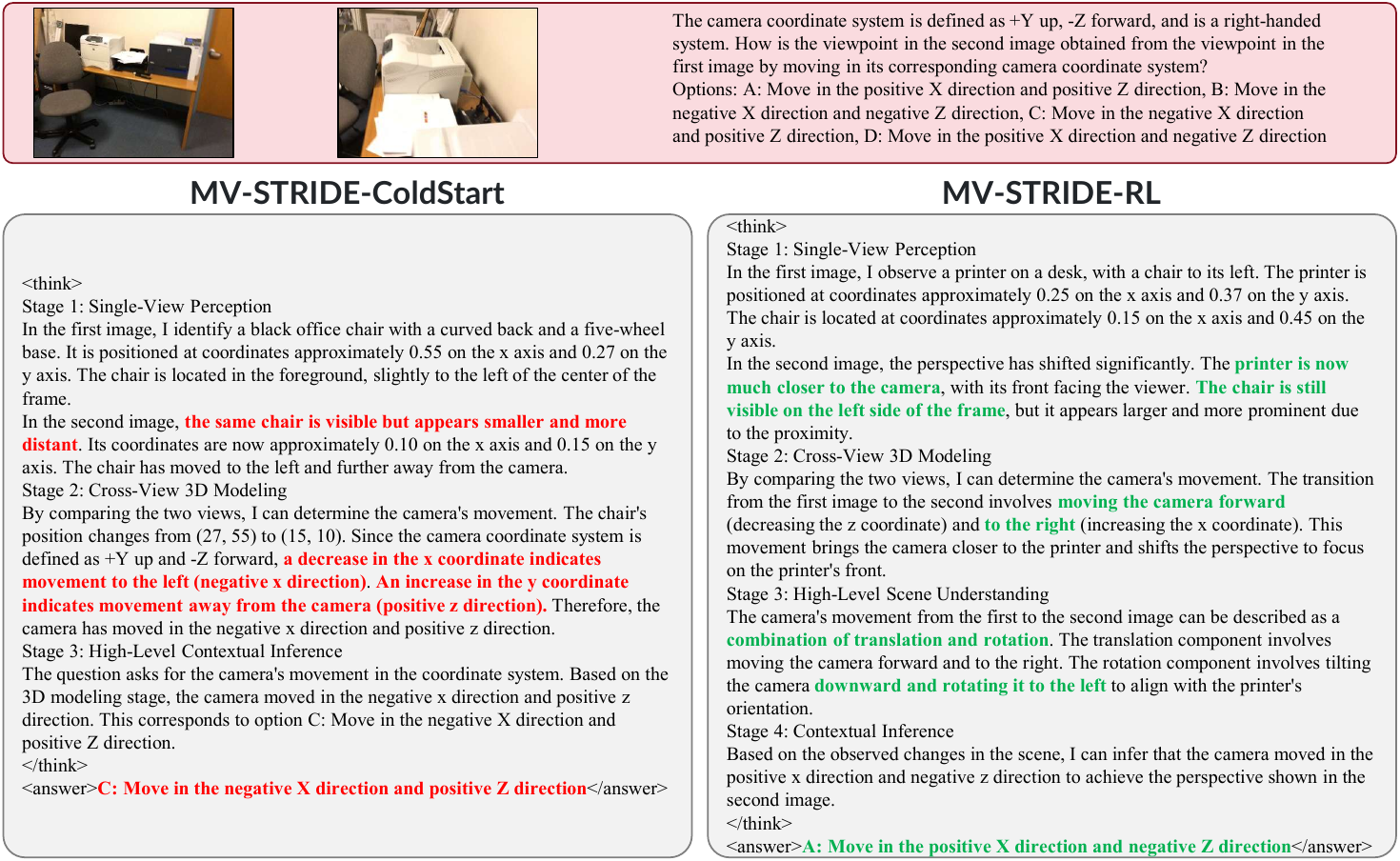}
  \caption{Comparison of CoT reasoning between MV-STRIDE-ColdStart and MV-STRIDE-RL.}
  \label{fig:cot_comparison}
\end{figure}
In this section, we compare the CoT outputs generated by MV-STRIDE-ColdStart (Stage 2) and MV-STRIDE-RL (Stage 3) in \cref{fig:cot_comparison}. Our analysis reveals that MV-STRIDE-ColdStart tends to imitate the output patterns present in the training datasets; consequently, it frequently exhibits perception and logical errors in intermediate reasoning steps, sometimes even reaching incorrect conclusions from correct premises. In contrast, the CoT generated by MV-STRIDE-RL is notably more flexible, demonstrating a coherent, step-by-step reasoning process that consistently leads to accurate answers.

\subsection{Generalization across scenario types} 
Since MV-STRIDE is constructed from indoor 3D scenes, we further examine whether the learned spatial reasoning ability can generalize beyond the training-domain scenario type. Therefore, our goal is to evaluate whether training on MV-STRIDE improves performance not only on indoor scenario, but also on dynamic, egocentric, and outdoor cases in benchmark. To this end, we use Gemini-3-Flash to classify the images inputs of MMSI-Bench into the above four scenario types. We then report the accuracy of the base model and MV-STRIDE-SFT(Full) in each category. As shown in \cref{tab:category_acc}, MV-STRIDE consistently improves spatial reasoning ability across all scenario types, demonstrating the effectiveness and generalization ability of our data construction pipeline. 
\begin{table}[t] 
\centering 
\caption{Generalization analysis of MV-STRIDE on different scenario types of MMSI-Bench.} 
\label{tab:category_acc} 
\small \begin{tabular}{lccccc} 
\toprule Model & Dynamic & Egocentric & Indoor & Outdoor & All \\ 
\midrule Qwen3-VL-7B-Instruct & 26.98 & 26.73 & 29.31 & 27.08 & 28.80 \\ 
MV-STRIDE-SFT(Full) & \textbf{31.75} & \textbf{29.70} & \textbf{41.24} & \textbf{29.17} & \textbf{38.90} \\ 
\bottomrule 
\end{tabular} 
\end{table}

\subsection{Validation across Architectures and Scales of Base Model}
\label{app:base_model_validation}
To examine whether the effectiveness of MV-STRIDE generalizes beyond a single base model, we further fine-tune multiple MLLMs with different parameter scales and architectures using our dataset. The experment setting is the same as MV-STRIDE-SFT(Full). As shown in \cref{tab:scale_validation}, MV-STRIDE consistently improves performance across all evaluated models and benchmarks. These results demonstrate that the gains brought by MV-STRIDE are not tied to a specific backbone, but reflect the general effectiveness of our hierarchical multi-view spatial reasoning data.
\begin{table}[t]
\centering
\caption{Validation across different architectures and parameter scales of base model.}
\label{tab:scale_validation}
\small
\begin{tabular}{lcccc}
\toprule
Model & MMSI & CV & VS & 3DSR \\
\midrule
Qwen3-VL-2B & 28.50 & 76.80 & 36.13 & 56.04 \\
+ MV-STRIDE & \textbf{32.50} & \textbf{83.81} & \textbf{45.64} & \textbf{61.70} \\
\midrule
Qwen3-VL-4B & 27.20 & 83.43 & 38.32 & 59.05 \\
+ MV-STRIDE & \textbf{36.20} & \textbf{86.35} & \textbf{49.26} & \textbf{64.36} \\
\midrule
MiMo-VL-7B & 29.00 & 82.33 & 40.37 & 59.31 \\
+ MV-STRIDE & \textbf{38.90} & \textbf{85.03} & \textbf{50.42} & \textbf{61.78} \\
\bottomrule
\end{tabular}
\end{table}

\subsection{Comparison with Recent Spatial Reasoning Methods}
To make comparison, we collected their reported results and evaluated SpatialLadder on MMSI-Bench and 3DSRBench. The results are shown in \cref{tab:recent_methods}. While 3DThinker scores higher on some benchmarks due to \textbf{model-level modifications} (3D latent representations, geometric alignment), MV-STRIDE focuses on \textbf{data-centric hierarchical multi-view QA} without changing the architecture. The two approaches are complementary. Reproducing 3DThinker results would need prohibitively long training, but we believe with our generated dataset, the performance can be further improved.
SpatialLadder emphasizes dataset design and progressive training, but on multi-view spatial reasoning tasks, our MV-STRIDE performs much better, highlighting the effectiveness of high-quality hierarchical multi-view QA. 
\begin{table}[t] 
\centering 
\caption{Comparison with recent spatial reasoning methods.} 
\label{tab:recent_methods} 
\small 
\begin{tabular}{lcccc} 
\toprule Method & MMSI & CV & VS & 3DSR \\ 
\midrule 
3DThinker & \textbf{43.30} & 81.10 & \textbf{68.60} & - \\ 
GeoThinker & 30.90 & \underline{85.10} & 45.90 & \underline{51.90} \\ SpatialLadder & 26.10 & 73.70 & 44.20 & 49.16 \\ 
MV-STRIDE-SFT(Full) & \underline{38.90} & \textbf{86.73} & \underline{48.35} & \textbf{64.51} \\ 
\bottomrule 
\end{tabular} 
\end{table}

\subsection{Ablation on Hierarchical Data Levels}
We conduct this controlled ablation during Stage 1 using only our generated QA data, where each level is uniformly subsampled to 50K QA pairs.
As shown in \cref{tab:level_ablation}, removing lower-level data decreases performance across most benchmarks, demonstrating that each level contributes to model learning and that sufficient training data from all levels is beneficial for optimal multi-view spatial reasoning.
\begin{table}[t] 
\centering 
\caption{Ablation on hierarchical data levels. } 
\label{tab:level_ablation} 
\small 
\begin{tabular}{lcccc} 
\toprule Training data & MMSI & CV & VS & 3DSR \\ 
\midrule 
L1+L2+L3 & \textbf{38.50} & \textbf{83.78} & \textbf{51.17} & \textbf{53.75} \\ 
w/o L1 & \underline{37.30} & \underline{82.79} & \underline{48.77} & 52.82 \\ 
w/o L1,L2 & 36.80 & 81.35 & 46.43 & \underline{53.03} \\ 
\bottomrule 
\end{tabular} 
\end{table}
\section{Dataset Construction and Specifications}
\label{sec:data_construct}
\subsection{Synthetic Data Generation via Infinigen}
\label{sec:synthetic_data}
We utilize Infinigen Indoors \cite{raistrick2024infinigen} as a procedural generator to synthesize high-fidelity 3D assets and indoor scenes. Following the standard pipeline of the project, we sequentially generate coarse- and fine-grained scene geometries. To ensure the quality of training samples, we employ a strategy of generating massive candidates followed by heuristic filtering to determine valid camera poses. 

During the scene generation phase, we maintain most of the default procedural parameters while introducing specific constraints tailored for our spatial QA requirements. Specifically, we enforce indoor lighting to remain active and restrict each scene to a single room (set to 1) to focus on dense object interactions. For camera synthesis, we adopt the Stereo Matching mode to generate 10 distinct viewpoints per scene. By disabling fixed random seeds to maximize diversity, we successfully generated 441 unique indoor environments, resulting in a total of 4,410 high-resolution images.

Following the synthesis and rendering of each scene, we extract both 2D and 3D ground-truth annotations required for QA generation by parsing the preserved metadata and the corresponding Blender (.blend) files. For 3D annotations, we retrieve comprehensive object-level attributes, including semantic categories, 3D centroid coordinates, and the orientations of the local coordinate axes relative to the world frame. Furthermore, we extract precise 3D bounding boxes and camera parameters, encompassing both intrinsic matrices and extrinsic poses (rotation and translation). For 2D annotations, we project the 3D object properties onto each image plane to obtain 2D bounding boxes and identify visible object instances per viewpoint. This rigorous data extraction pipeline ensures spatial consistency across multi-view observations, providing a robust foundation for our hierarchical reasoning tasks.
\subsection{Real-world Data Processing of ScanNet++}
\label{sec:scannet_processing}

To complement our synthetic data, we incorporate 124 real-world indoor scenes from the ScanNet++ \cite{yeshwanth2023scannet++} training set to construct our QA dataset. To ensure the highest precision of camera parameters, we exclusively utilize the \textit{iPhone} image subset. While the original capture density is high, not all viewpoints are provided with valid camera intrinsic and extrinsic metadata; therefore, we only retain frames with verified pose annotations. Given that these annotated viewpoints exhibit relatively moderate overlap, further temporal downsampling is not required.

ScanNet++ provides calibrated camera parameters and high-quality 3D meshes with fine-grained instance segmentation. We extract the 3D attributes of each object directly from the annotated meshes and obtain corresponding 2D labels (e.g., bounding boxes and visibility) by rendering the 3D instances onto the image planes using the provided camera poses. After extracting all raw metadata, we perform a rigorous coordinate unification process to align the ScanNet++ data format with our Infinigen pipeline. This unification covers the ordering of 3D and 2D bounding box coordinates, as well as the orientation conventions of the camera coordinate systems, ensuring a consistent spatial representation across both synthetic and real-world domains.
\subsection{Coordinate System and Bounding Boxes}
\label{sec:coord_systems}

To ensure consistency across heterogeneous data sources, we unify the coordinate system conventions for all metadata. We follow the \textbf{Blender-style coordinate definition}, where the camera's local axes are defined with $-Z$ pointing forward (viewing direction), $+Y$ pointing upward, and $+X$ pointing to the right. The camera extrinsics are represented by Camera-to-World ($C2W$) matrices.

In the generated QA pairs, we intentionally avoid ambiguous geometric jargon to maintain natural language clarity. Instead, we describe camera movements using intuitive, common-sense terms (e.g., "rotating the camera up, down, left, or right"). For tasks requiring precise spatial grounding, we explicitly define the reference frame within the prompt (e.g., ``The camera coordinate system is defined as a right-handed system with $+Z$ as up and $+X$ as forward''). Regarding 2D localization, object positions in the image plane are represented by 2D bounding boxes using the $[x_1, y_1, x_2, y_2]$ format, where $(x_1, y_1)$ and $(x_2, y_2)$ denote the top-left and bottom-right coordinates. To maintain architectural consistency with the base model (Qwen3-VL\cite{bai2025qwen3}), we normalize and rescale all coordinates to a range of $[0, 1000]$.

\subsection{Quality Verification}
We verified quality of randomly sampled QA and CoT of MV-STRIDE, with 200 samples each. QA samples were double-checked by two human annotators for visual validity, answer correctness, language clarity, cross-view dependency, and question reasonableness, with error counts of 8, 7, 2, 0, and 7 respectively, yielding an overall accuracy of 88\%. CoT annotations were checked by Gemini-3-Flash for factual consistency, reasoning faithfulness, final-answer consistency, and hallucination, with error counts of 4, 0, 1, and 0, corresponding to an overall accuracy of 97.5\%. 
The results show that CoT annotations are highly accurate and QA samples contain minor noise, but training on our large automatically generated dataset still significantly improves the model’s spatial reasoning.

\section{Dataset Gallery}
\label{sec:gallery}
\subsection{Hierarchical Dataset Definitions and Examples} 
As introduced in the main text, our dataset is defined at three hierarchical levels, each containing multiple task categories. In what follows, we present the definitions and illustrative examples of all spatial reasoning tasks in a hierarchical manner.
\subsubsection{Level I: Single-View Spatial Perception.}
Level I tasks focus on fundamental perception within a single static observation. In \cref{tab:depth_estimation,tab:camera_object_yaw,tab:camera_pitch,tab:object_location,tab:size_comparison}, we present the comprehensive definitions and illustrative examples for all categories within this level.
\begin{table}[!htbp]
    \centering
    \captionof{table}{Definition and illustrative examples for Depth Estimation in Level I.}
    \label{tab:depth_estimation}
    \small
    \renewcommand{\arraystretch}{1.5}
    \begin{tabular}{|l|p{0.75\textwidth}|}
    \hline
    \textbf{Category}   & Depth Estimation \\ \hline
    \textbf{Definition} & {Estimate the relative distance of a specified object from the camera.} \\ \hline
    \multicolumn{2}{|l|}{\cellcolor[HTML]{F5F5F5}\textbf{Example 1}} \\ \hline
    \textbf{Input Image} & 
        \begin{minipage}{0.72\textwidth}
            \centering
            \includegraphics[width=0.48\linewidth]{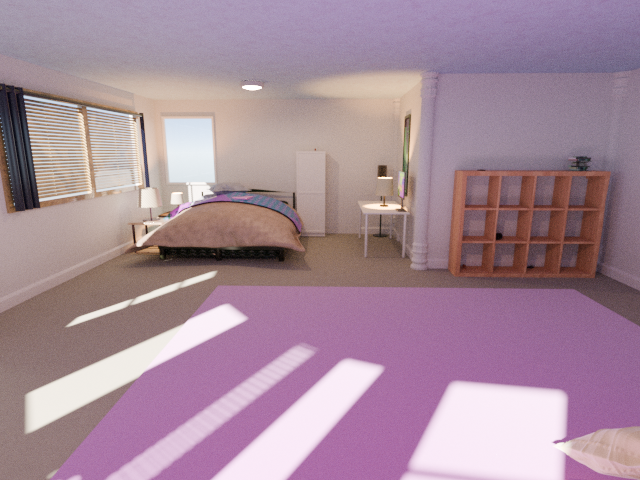}
        \end{minipage} \\ \hline
    \textbf{QA Content}  & 
        \begin{minipage}{0.72\textwidth}
            \textbf{Q:} In Figure, what is the straight-line distance from the camera to the 3D center of the cell shelf, measured in meters? \par
            \textbf{A:} In Figure, the cell shelf is located at a depth of approximately 6.37 meters from the camera.
        \end{minipage} \\ \hline
    \end{tabular}
\end{table}
\begin{table}[!htbp]
    \centering
    \captionof{table}{Definition and illustrative examples for Size Comparison in Level I.}
    \label{tab:size_comparison}
    \small
    \renewcommand{\arraystretch}{1.5}
    \begin{tabular}{|l|p{0.75\textwidth}|}
    \hline
    \textbf{Category}   & Size Comparison \\ \hline
    \textbf{Definition} & {Estimate the physical dimension of objects along a specific axis and calculate their relative ratio.} \\ \hline
    \multicolumn{2}{|l|}{\cellcolor[HTML]{F5F5F5}\textbf{Example 1}} \\ \hline
    \textbf{Input Image} & 
        \begin{minipage}{0.72\textwidth}
            \centering
            \includegraphics[width=0.48\linewidth]{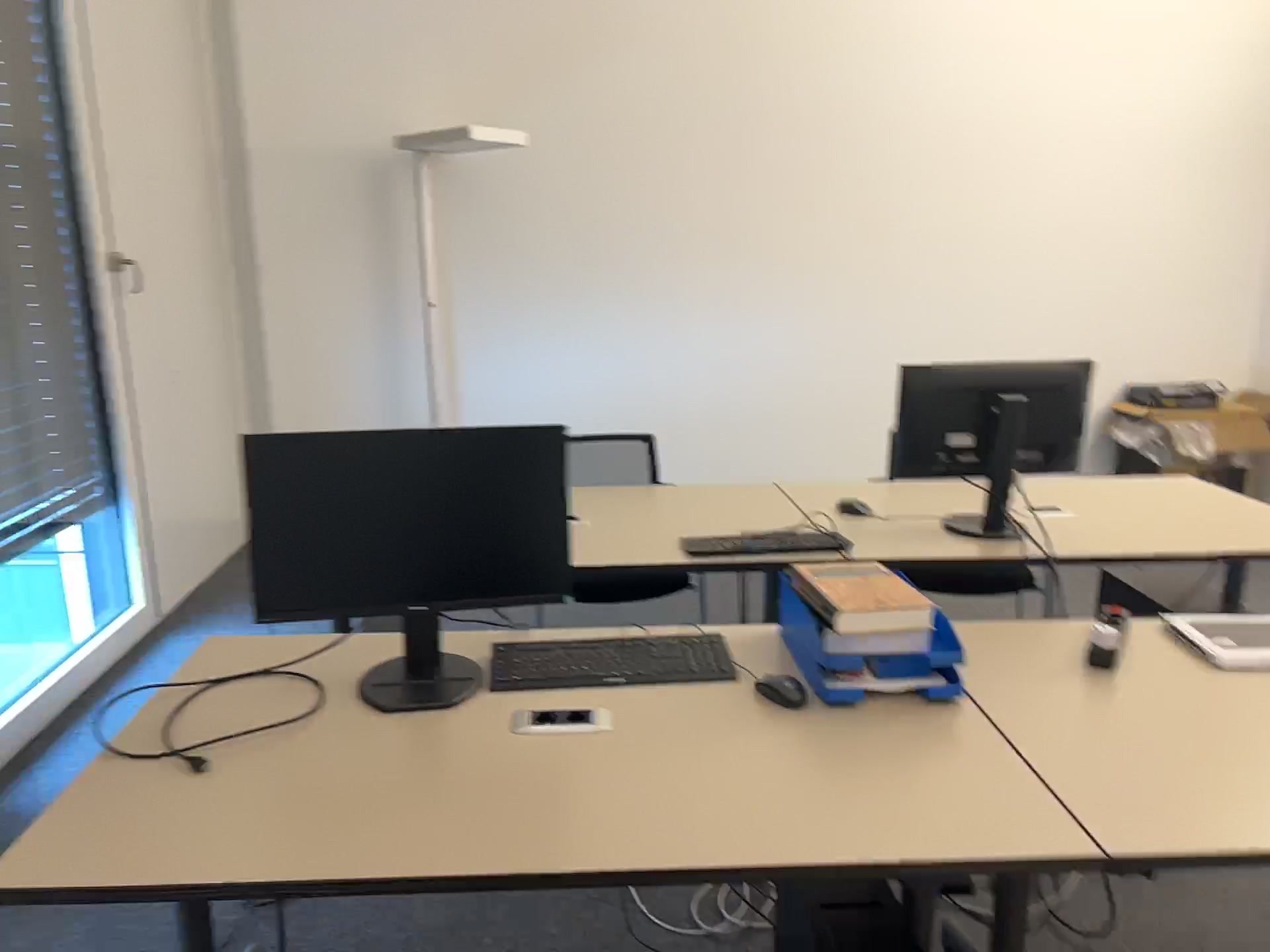}
        \end{minipage} \\ \hline
    \textbf{QA Content}  & 
        \begin{minipage}{0.72\textwidth}
            \textbf{Q:} In Figure, please estimate the physical heights of the rightmost monitor and the leftmost monitor. What is the ratio of the rightmost monitor's height to the leftmost monitor's? \par
            \textbf{A:} The rightmost monitor has an estimated height of 0.64 meters, and the leftmost monitor stands 0.6 meters tall. Thus, the rightmost monitor is 0.94 times the height of the leftmost monitor.
        \end{minipage} \\ \hline
    \end{tabular}
\end{table}
\begin{table}[!htbp]
    \centering
    \captionof{table}{Definition and illustrative examples for Object Location in Level I.}
    \label{tab:object_location}
    \small
    \renewcommand{\arraystretch}{1.5}
    \begin{tabular}{|l|p{0.75\textwidth}|}
    \hline
    \textbf{Category}   & Object Location \\ \hline
    \textbf{Definition} & {Identify the 2D bounding box of a specific object based on its natural language description.} \\ \hline
    \multicolumn{2}{|l|}{\cellcolor[HTML]{F5F5F5}\textbf{Example 1}} \\ \hline
    \textbf{Input Image} & 
        \begin{minipage}{0.72\textwidth}
            \centering
            \includegraphics[width=0.48\linewidth]{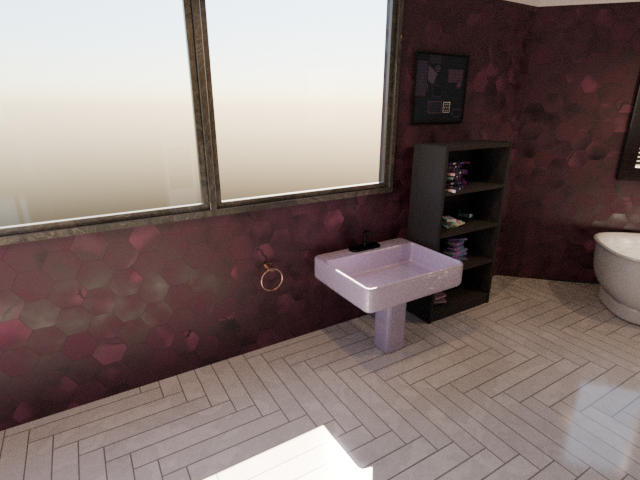}
        \end{minipage} \\ \hline
    \textbf{QA Content}  & 
        \begin{minipage}{0.72\textwidth}
            \textbf{Q:} In Figure, please locate the standing sink and provide its bounding box coordinates. \par
            \textbf{A:} The standing sink in Figure is located at [491, 479, 722, 735].
        \end{minipage} \\ \hline
    \end{tabular}
\end{table}
\begin{table}[!htbp]
    \centering
    \captionof{table}{Definition and illustrative examples for Camera-Object Yaw in Level I.}
    \label{tab:camera_object_yaw}
    \small
    \renewcommand{\arraystretch}{1.5}
    \begin{tabular}{|l|p{0.75\textwidth}|}
    \hline
    \textbf{Category}   & Camera-Object Yaw \\ \hline
    \textbf{Definition} & {Estimate the horizontal angular offset of a specific object relative to the camera's forward-facing axis, identifying whether it is positioned to the left or right.} \\ \hline
    \multicolumn{2}{|l|}{\cellcolor[HTML]{F5F5F5}\textbf{Example 1}} \\ \hline
    \textbf{Input Image} & 
        \begin{minipage}{0.72\textwidth}
            \centering
            \includegraphics[width=0.48\linewidth]{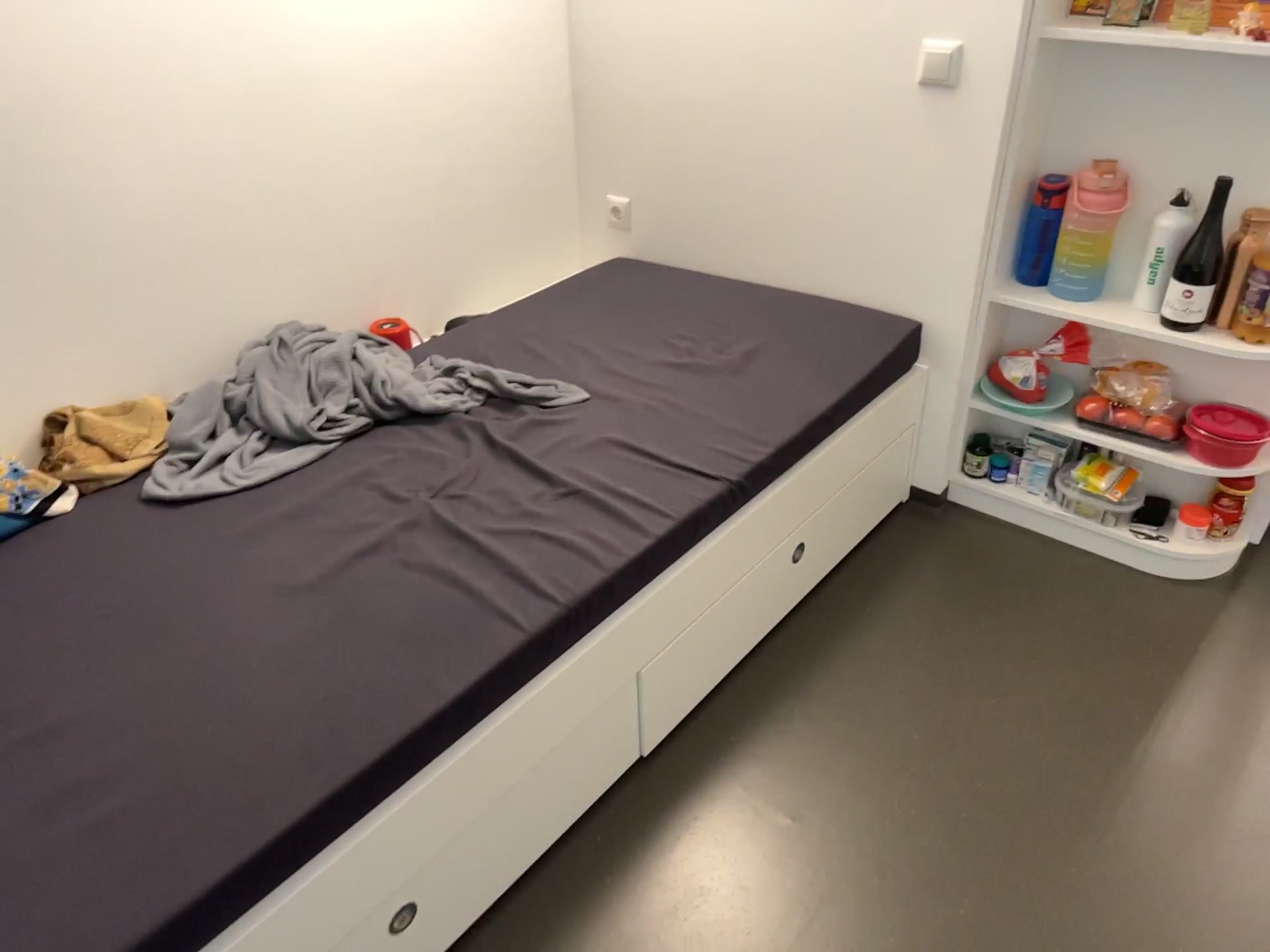}
        \end{minipage} \\ \hline
    \textbf{QA Content}  & 
        \begin{minipage}{0.72\textwidth}
            \textbf{Q:} In Figure, relative to the direction the camera is facing, what is the horizontal angular offset (yaw) of the shelf? Is it to the left or right of the camera's center line? \par
            \textbf{A:} In Figure, the shelf is positioned at a horizontal angle of 32.73 degrees to the right of the camera's central line of sight.
        \end{minipage} \\ \hline
    \end{tabular}
\end{table}
\begin{table}[!htbp]
    \centering
    \captionof{table}{Definition and illustrative examples for Camera Pitch in Level I.}
    \label{tab:camera_pitch}
    \small
    \renewcommand{\arraystretch}{1.5}
    \begin{tabular}{|l|p{0.75\textwidth}|}
    \hline
    \textbf{Category}   & Camera Pitch \\ \hline
    \textbf{Definition} & {Determine the vertical angle of the camera relative to the horizontal ground plane of the scene.} \\ \hline
    \multicolumn{2}{|l|}{\cellcolor[HTML]{F5F5F5}\textbf{Example 1}} \\ \hline
    \textbf{Input Image} & 
        \begin{minipage}{0.72\textwidth}
            \centering
            \includegraphics[width=0.48\linewidth]{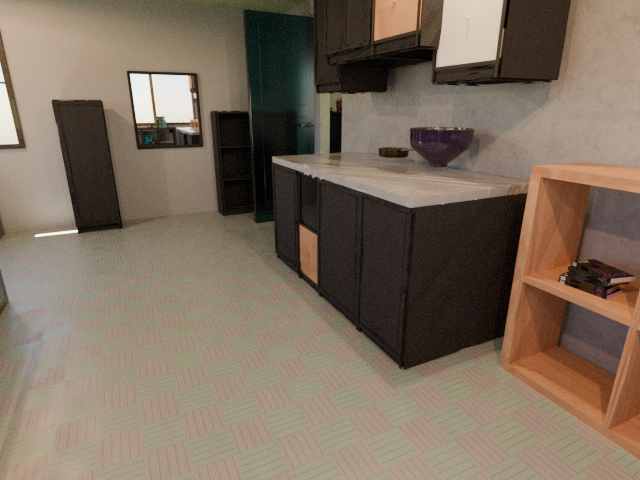}
        \end{minipage} \\ \hline
    \textbf{QA Content}  & 
        \begin{minipage}{0.72\textwidth}
            \textbf{Q:} Considering the perspective and the layout of the scene in Figure, what is the vertical tilt (pitch) of the camera? Is it pointing upwards, downwards, or staying level? \par
            \textbf{A:} Based on the visual evidence in Figure, the camera is tilted 18.16 degrees downwards relative to the horizontal plane.
        \end{minipage} \\ \hline
    \end{tabular}
\end{table}
\subsubsection{Level II: 3D-Consistent Scene Understanding.}
Level II shifts from static perception to 3D-consistent scene understanding, requiring the model to maintain spatial coherence across multiple viewpoints. In \cref{tab:object_correspondence,tab:camera_rotation_pitch,tab:camera_rotation_yaw,tab:camera_translation_forward,tab:camera_translation_right,tab:camera_object_3d_position}, we present the comprehensive definitions and illustrative examples for all categories within this level.
\begin{table}[!htbp]
    \centering
    \captionof{table}{Definition and illustrative examples for Object Correspondence in Level II.}
    \label{tab:object_correspondence}
    \small
    \renewcommand{\arraystretch}{1.5}
    \begin{tabular}{|l|p{0.75\textwidth}|}
    \hline
    \textbf{Category}   & Object Correspondence \\ \hline
    \textbf{Definition} & {Locate the corresponding 2D bounding box of a specific object in a secondary view, given its position in a primary input view.} \\ \hline
    \multicolumn{2}{|l|}{\cellcolor[HTML]{F5F5F5}\textbf{Example 1}} \\ \hline
    \textbf{Input Image} & 
        \begin{minipage}{0.72\textwidth}
            \centering
            \includegraphics[width=0.48\linewidth]{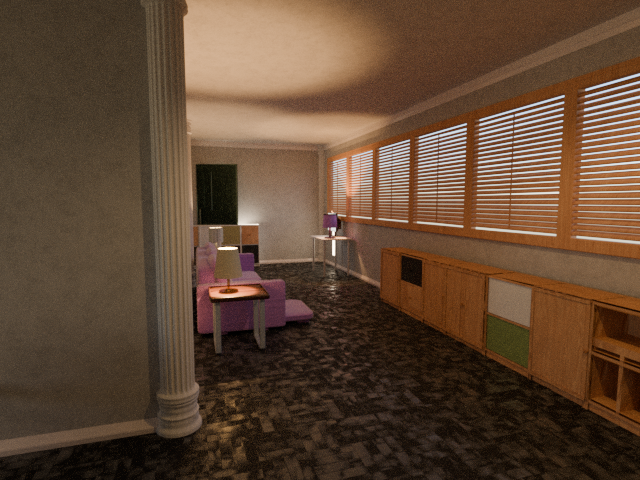} \hfill
            \includegraphics[width=0.48\linewidth]{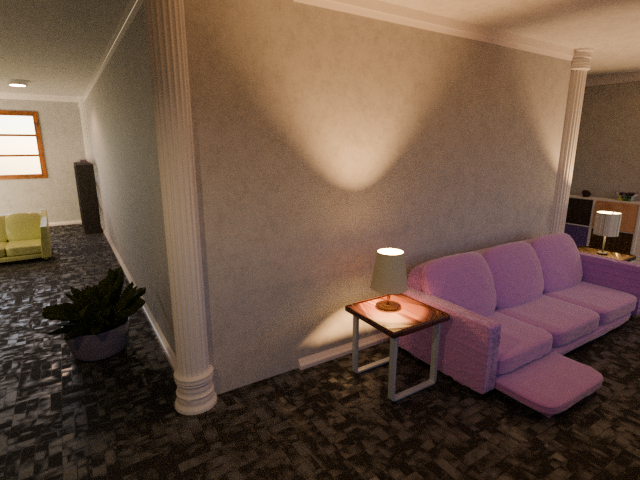}
        \end{minipage} \\ \hline
    \textbf{QA Content}  & 
        \begin{minipage}{0.72\textwidth}
            \textbf{Q:} In Figure 2, there is a leftmost desk lamp located at [578, 517, 636, 646]. Can you find this same physical object in Figure 1? If so, what are its coordinates in Figure 1? \par
            \textbf{A:} Yes, the same leftmost desk lamp is visible in Figure 1, and its coordinates are [334, 515, 378, 608].
        \end{minipage} \\ \hline
    \end{tabular}
\end{table}
\begin{table}[!htbp]
    \centering
    \captionof{table}{Definition and illustrative examples for Camera Rotation (yaw) in Level II.}
    \label{tab:camera_rotation_yaw}
    \small
    \renewcommand{\arraystretch}{1.5}
    \begin{tabular}{|l|p{0.75\textwidth}|}
    \hline
    \textbf{Category}   & Camera Rotation (yaw) \\ \hline
    \textbf{Definition} & {Estimate the relative horizontal angular change between two camera viewpoints based on the scene's geometric context.} \\ \hline
    \multicolumn{2}{|l|}{\cellcolor[HTML]{F5F5F5}\textbf{Example 1}} \\ \hline
    \textbf{Input Image} & 
        \begin{minipage}{0.72\textwidth}
            \centering
            \includegraphics[width=0.48\linewidth]{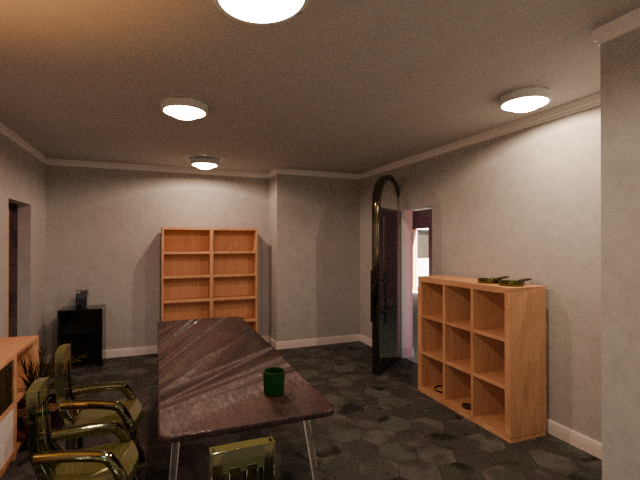} \hfill
            \includegraphics[width=0.48\linewidth]{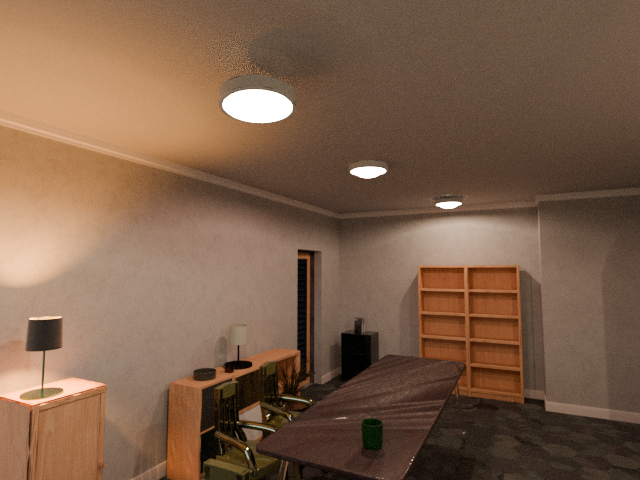}
        \end{minipage} \\ \hline
    \textbf{QA Content}  & 
        \begin{minipage}{0.72\textwidth}
            \textbf{Q:} From Figure 1 to Figure 2, in which horizontal direction did the camera rotate, and what is the rotation angle? \par
            \textbf{A:} The camera rotated left by approximately 48.68 degrees.
        \end{minipage} \\ \hline
    \end{tabular}
\end{table}
\begin{table}[!htbp]
    \centering
    \captionof{table}{Definition and illustrative examples for Camera Rotation (pitch) in Level II.}
    \label{tab:camera_rotation_pitch}
    \small
    \renewcommand{\arraystretch}{1.5}
    \begin{tabular}{|l|p{0.75\textwidth}|}
    \hline
    \textbf{Category}   & Camera Rotation (pitch) \\ \hline
    \textbf{Definition} & {Estimate the relative vertical angular change between two camera viewpoints based on the scene's geometric context.} \\ \hline
    \multicolumn{2}{|l|}{\cellcolor[HTML]{F5F5F5}\textbf{Example 1}} \\ \hline
    \textbf{Input Image} & 
        \begin{minipage}{0.72\textwidth}
            \centering
            \includegraphics[width=0.48\linewidth]{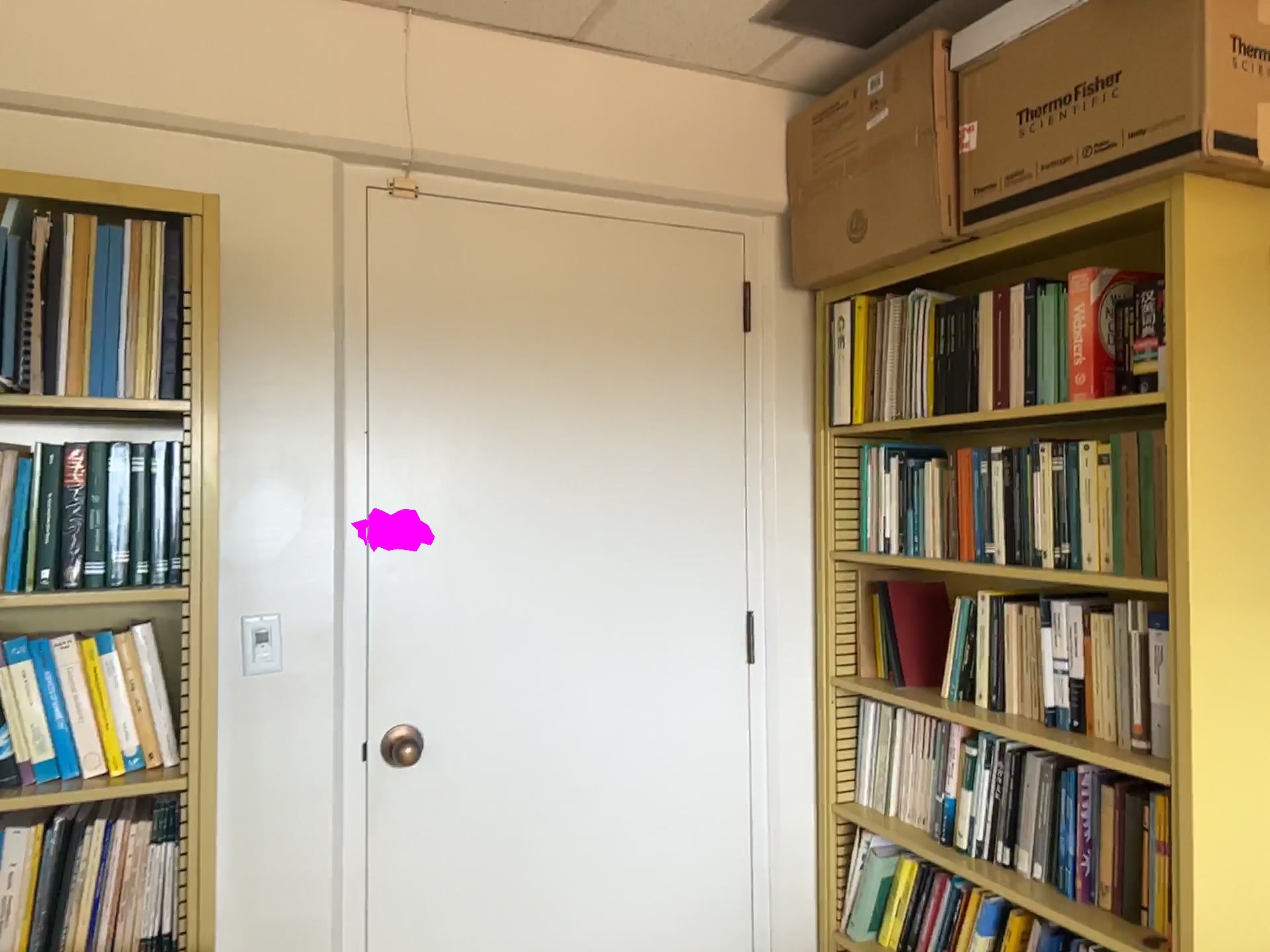} \hfill
            \includegraphics[width=0.48\linewidth]{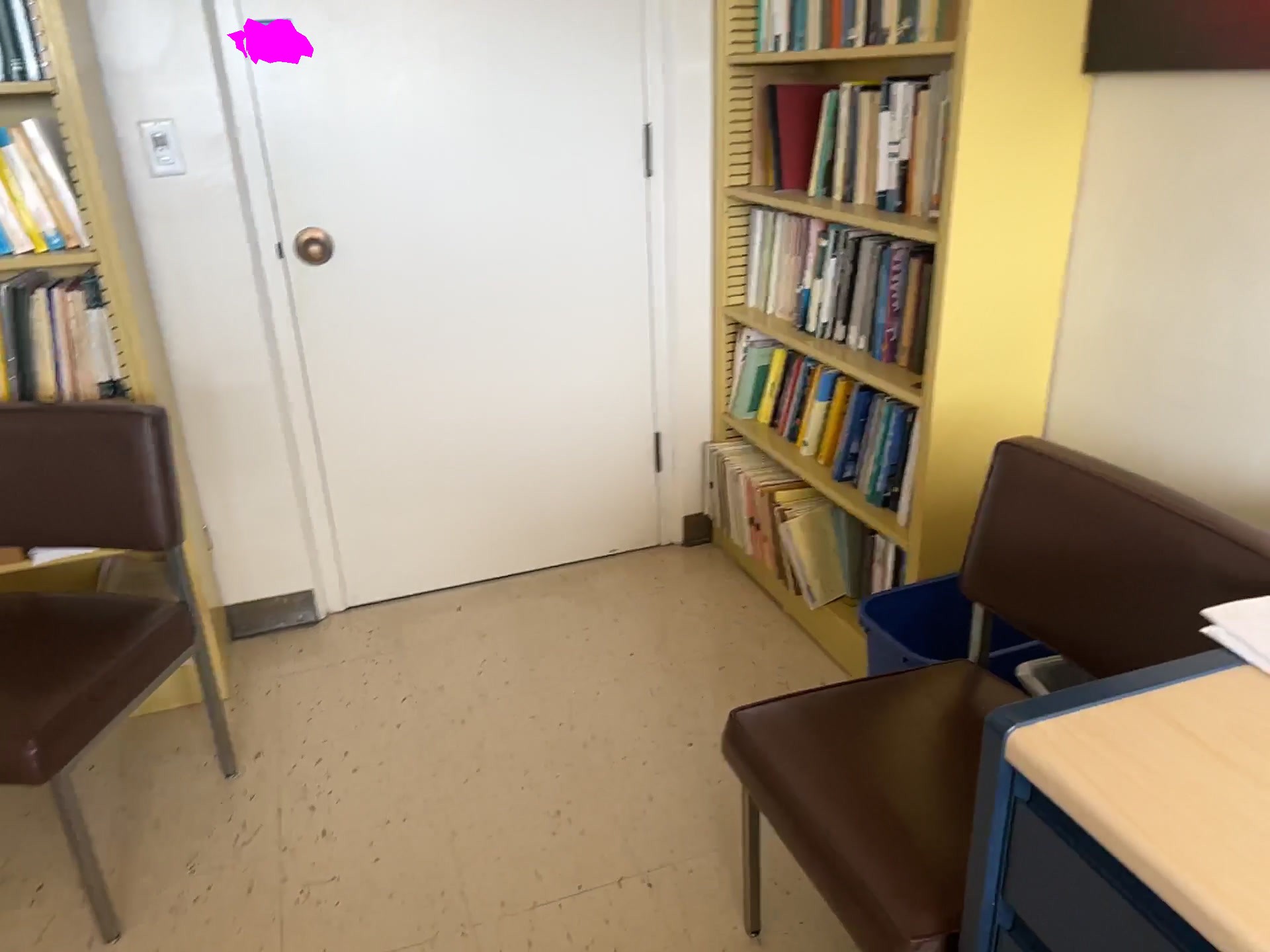}
        \end{minipage} \\ \hline
    \textbf{QA Content}  & 
        \begin{minipage}{0.72\textwidth}
            \textbf{Q:} From Figure 1 to Figure 2, in which vertical direction did the camera rotate, and what is the rotation angle? \par
            \textbf{A:} The camera tilted down by approximately 23.99 degrees.
        \end{minipage} \\ \hline
    \end{tabular}
\end{table}
\begin{table}[!htbp]
    \centering
    \captionof{table}{Definition and illustrative examples for Camera Translation (right) in Level II.}
    \label{tab:camera_translation_right}
    \small
    \renewcommand{\arraystretch}{1.5}
    \begin{tabular}{|l|p{0.75\textwidth}|}
    \hline
    \textbf{Category}   & Camera Translation (right) \\ \hline
    \textbf{Definition} & {Estimate the relative horizontal displacement distance along the right-left axis between two camera viewpoints.} \\ \hline
    \multicolumn{2}{|l|}{\cellcolor[HTML]{F5F5F5}\textbf{Example 1}} \\ \hline
    \textbf{Input Image} & 
        \begin{minipage}{0.72\textwidth}
            \centering
            \includegraphics[width=0.48\linewidth]{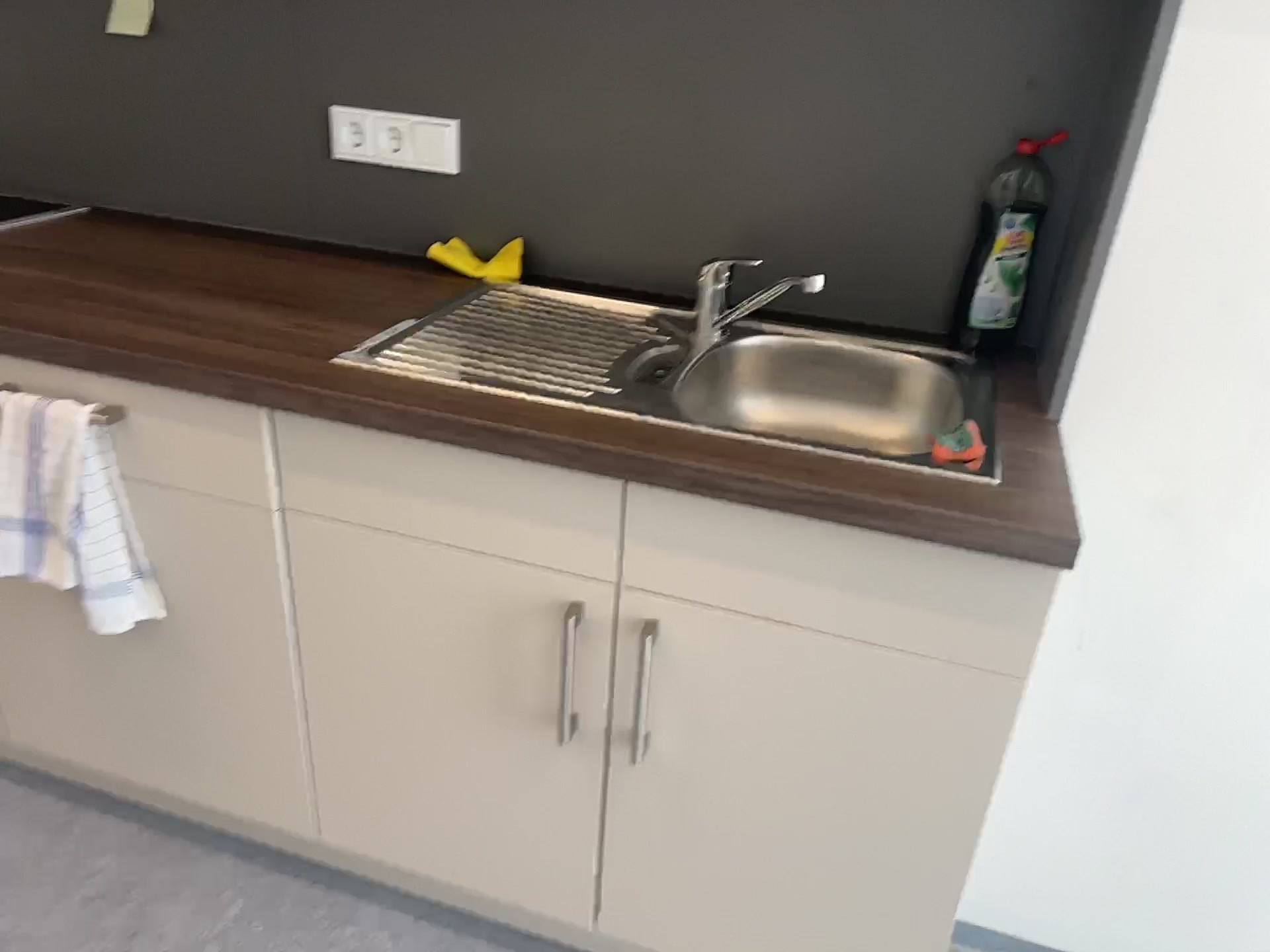} \hfill
            \includegraphics[width=0.48\linewidth]{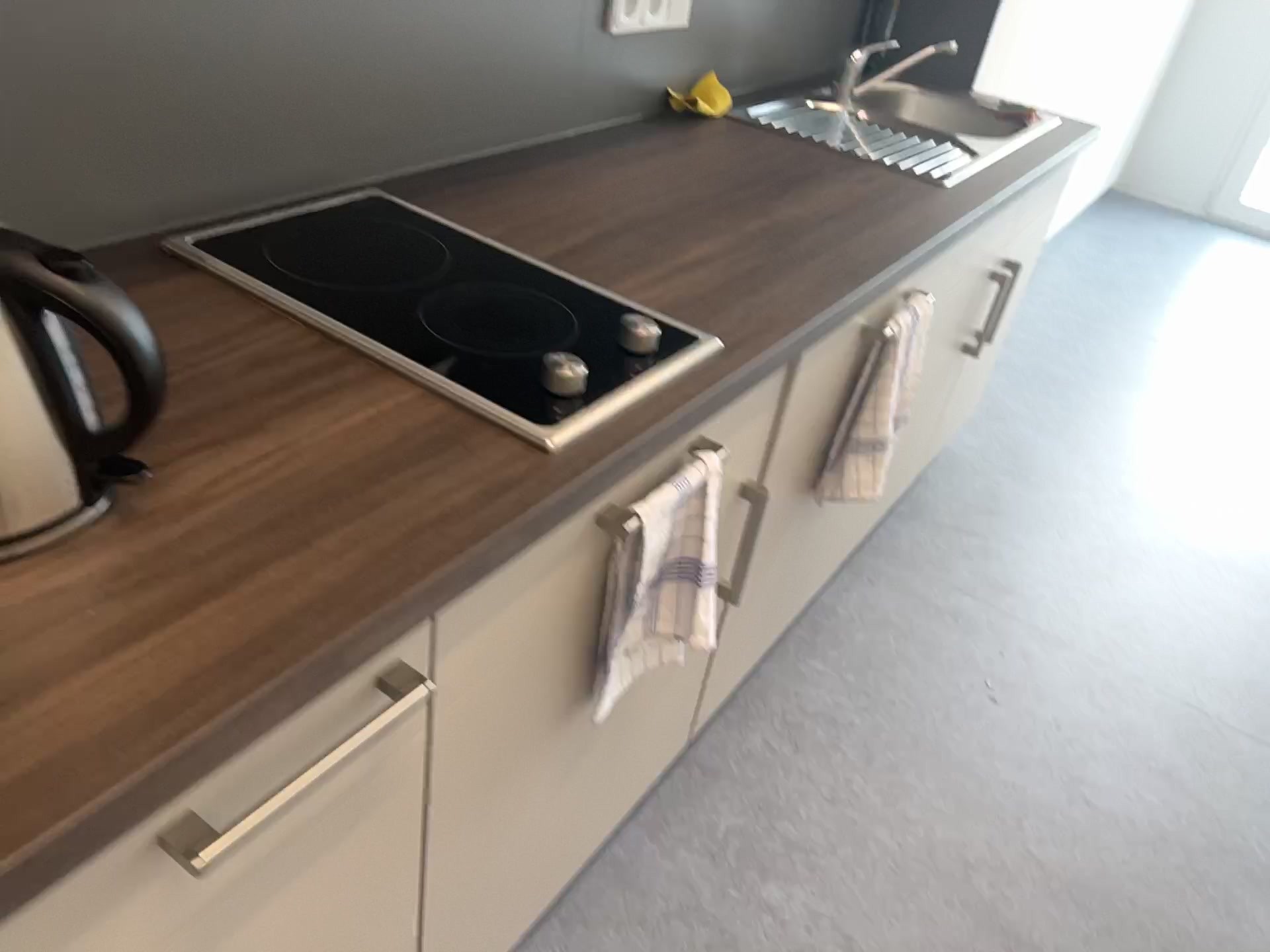}
        \end{minipage} \\ \hline
    \textbf{QA Content}  & 
        \begin{minipage}{0.72\textwidth}
            \textbf{Q:} In the local coordinate system of Figure 1, in which horizontal direction (left or right) did the camera shift to Figure 2, and what is the translation distance? \par
            \textbf{A:} The camera shifted to the left by approximately 2.0651 meters.
        \end{minipage} \\ \hline
    \end{tabular}
\end{table}
\begin{table}[!htbp]
    \centering
    \captionof{table}{Definition and illustrative examples for Camera Translation (forward) in Level II.}
    \label{tab:camera_translation_forward}
    \small
    \renewcommand{\arraystretch}{1.5}
    \begin{tabular}{|l|p{0.75\textwidth}|}
    \hline
    \textbf{Category}   & Camera Translation (forward) \\ \hline
    \textbf{Definition} & {Estimate the relative horizontal displacement distance along the forward-backward axis between two camera viewpoints.} \\ \hline
    \multicolumn{2}{|l|}{\cellcolor[HTML]{F5F5F5}\textbf{Example 1}} \\ \hline
    \textbf{Input Image} & 
        \begin{minipage}{0.72\textwidth}
            \centering
            \includegraphics[width=0.48\linewidth]{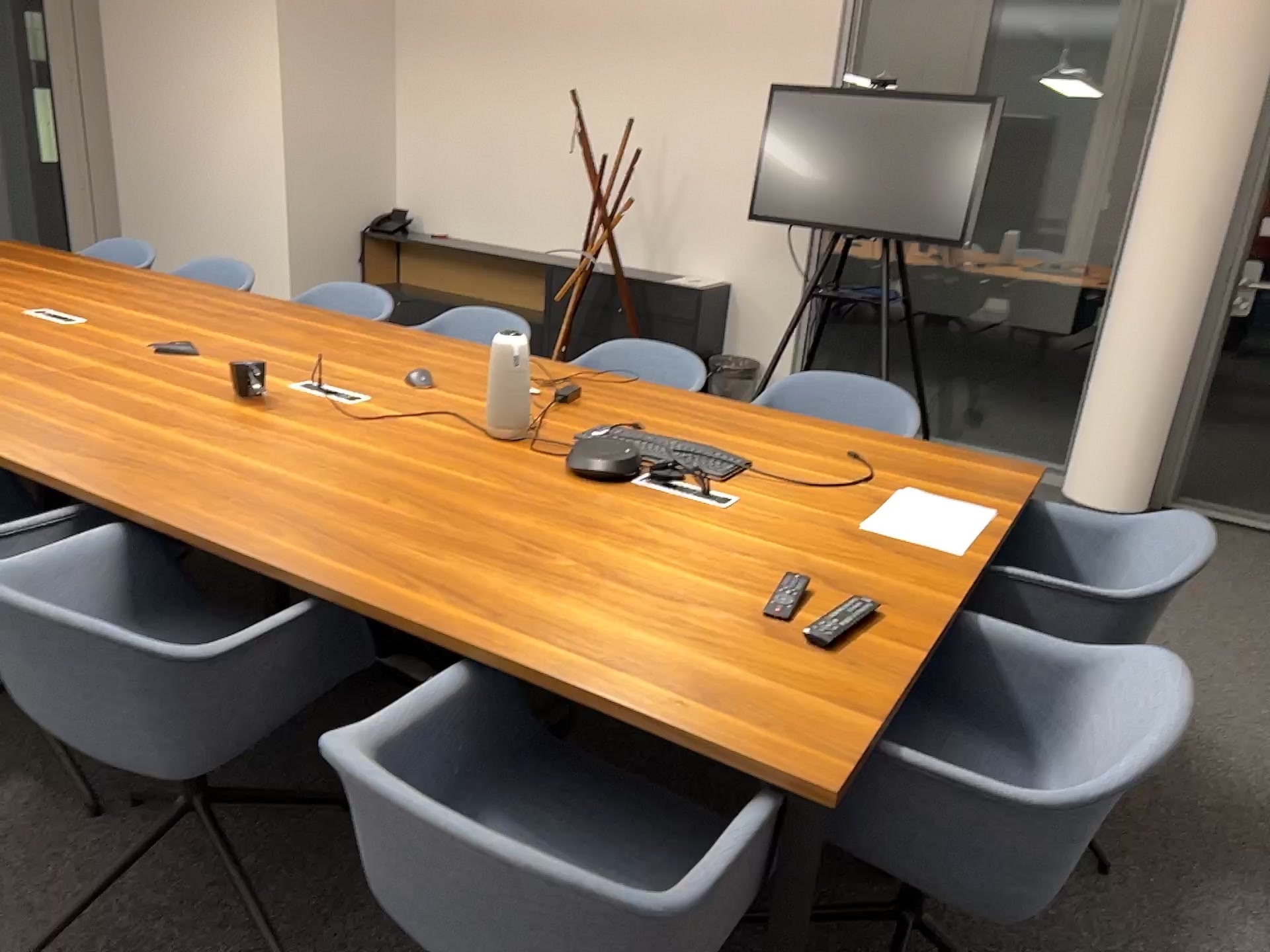} \hfill
            \includegraphics[width=0.48\linewidth]{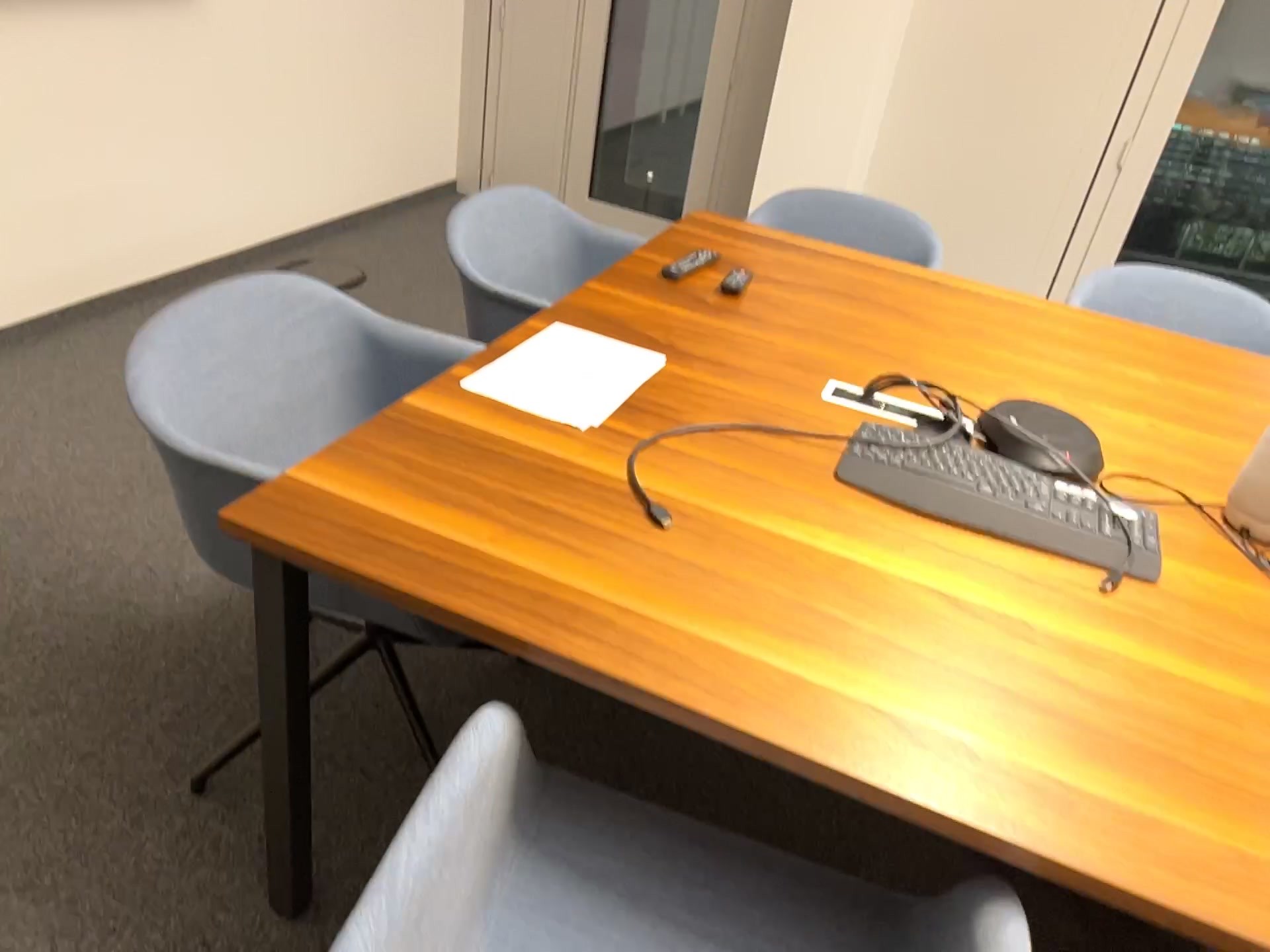}
        \end{minipage} \\ \hline
    \textbf{QA Content}  & 
        \begin{minipage}{0.72\textwidth}
            \textbf{Q:} Relative to the local coordinate system of the camera in Figure 1, did the camera move forward or backward when transitioning to Figure 2, and what is the distance? \par
            \textbf{A:} The camera moved forward by approximately 3.218 meters.
        \end{minipage} \\ \hline
    \end{tabular}
\end{table}
\begin{table}[!htbp]
    \centering
    \captionof{table}{Definition and illustrative examples for Camera-Object 3D Position in Level II.}
    \label{tab:camera_object_3d_position}
    \small
    \renewcommand{\arraystretch}{1.5}
    \begin{tabular}{|l|p{0.75\textwidth}|}
    \hline
    \textbf{Category}   & Camera-Object 3D Position \\ \hline
    \textbf{Definition} & {Estimate the 3D coordinate $(x, y, z)$ of a specified object relative to the camera's local coordinate system.} \\ \hline
    \multicolumn{2}{|l|}{\cellcolor[HTML]{F5F5F5}\textbf{Example 1}} \\ \hline
    \textbf{Input Image} & 
        \begin{minipage}{0.72\textwidth}
            \centering
            \includegraphics[width=0.48\linewidth]{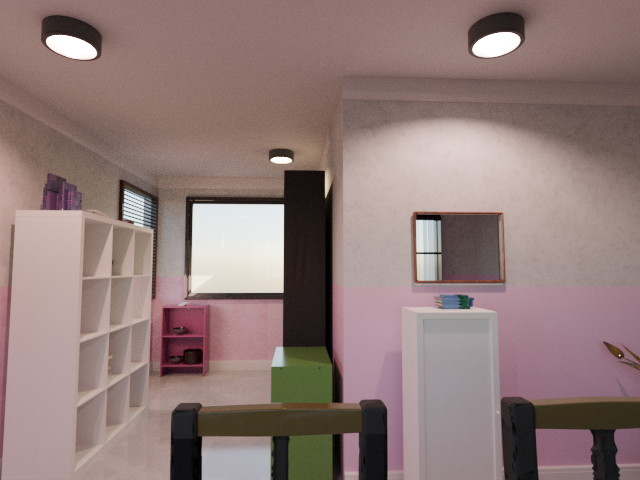}
        \end{minipage} \\ \hline
    \textbf{QA Content}  & 
        \begin{minipage}{0.72\textwidth}
            \textbf{Q:} Considering the camera's viewpoint in Figure, what are the 3D coordinates of the mirror center within the camera's own local coordinate system (where the camera is at [0, 0, 0])? \par
            \textbf{A:} In the local coordinate system of the camera of Figure, the mirror is positioned at approximately x=0.92, y=7.63, and z=-5.03 (measured in meters), where z represents the relative depth.
        \end{minipage} \\ \hline
    \end{tabular}
\end{table}

\subsubsection{Level III: Multi-View High-Level Contextual Reasoning.}
Level III focuses on high-order, context-dependent spatial reasoning, representing our most comprehensive task suite for multi-view scene interpretation. In \cref{tab:counting,tab:object_orientation,tab:size_comparison_3,tab:camera_motion,tab:position_relationship_cam_obj,tab:position_relationship_obj_obj,tab:virtual_perspective}, we detail the core definitions and illustrative examples for these categories.
\begin{table}[!htbp]
    \centering
    \captionof{table}{Definition and illustrative examples for Counting in Level III.}
    \label{tab:counting}
    \small
    \renewcommand{\arraystretch}{1.5}
    \begin{tabular}{|l|p{0.75\textwidth}|}
    \hline
    \textbf{Category}   & Counting \\ \hline
    \textbf{Definition} & {Estimate the total number of instances of a specific object category across multiple images, where objects are not all visible in any single viewpoint.} \\ \hline
    \multicolumn{2}{|l|}{\cellcolor[HTML]{F5F5F5}\textbf{Example 1}} \\ \hline
    \textbf{Input Image} & 
        \begin{minipage}{0.72\textwidth}
            \centering
            \includegraphics[width=0.48\linewidth]{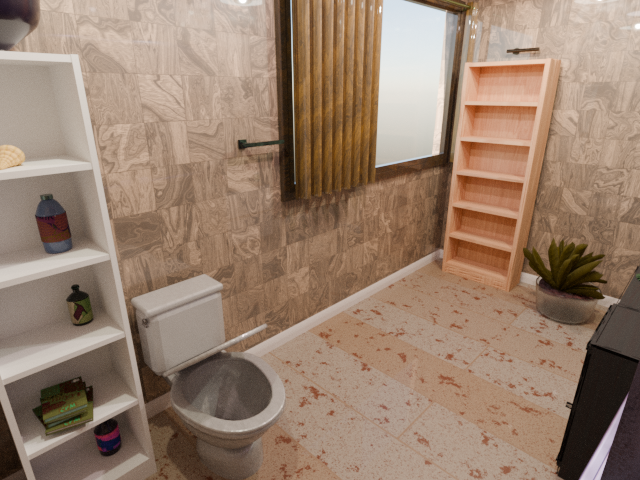} \hfill
            \includegraphics[width=0.48\linewidth]{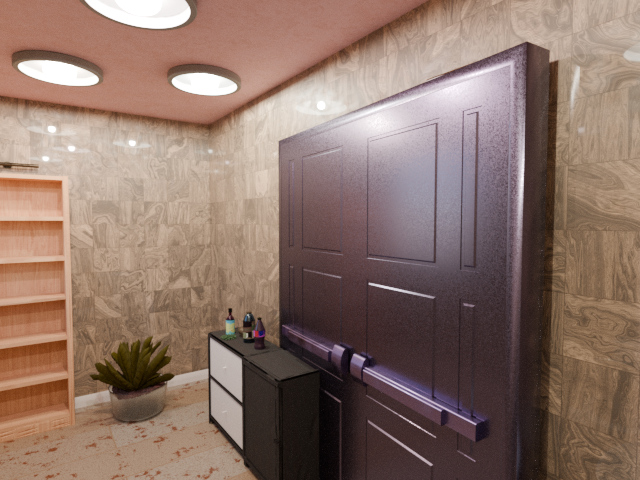}
        \end{minipage} \\ \hline
    \textbf{QA Content}  & 
        \begin{minipage}{0.72\textwidth}
            \textbf{Q:} From these two photos, how many different large shelf can you see? Options: A: 4, B: 3, C: 2, D: 5 \par
            \textbf{A:} C: 2
        \end{minipage} \\ \hline
    \multicolumn{2}{|l|}{\cellcolor[HTML]{F5F5F5}\textbf{Example 2}} \\ \hline
    \textbf{Input Image} & 
        \begin{minipage}{0.72\textwidth}
            \centering
            \includegraphics[width=0.48\linewidth]{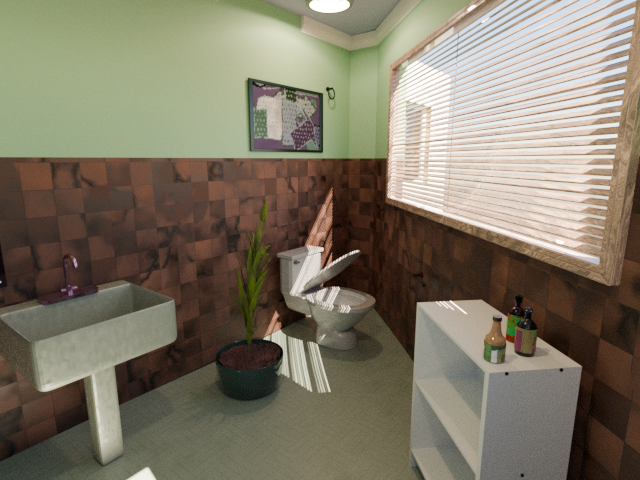} \hfill
            \includegraphics[width=0.48\linewidth]{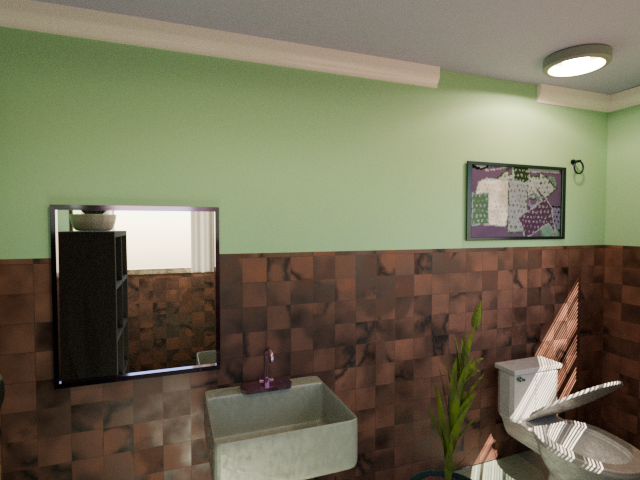} \hfill
            \includegraphics[width=0.48\linewidth]{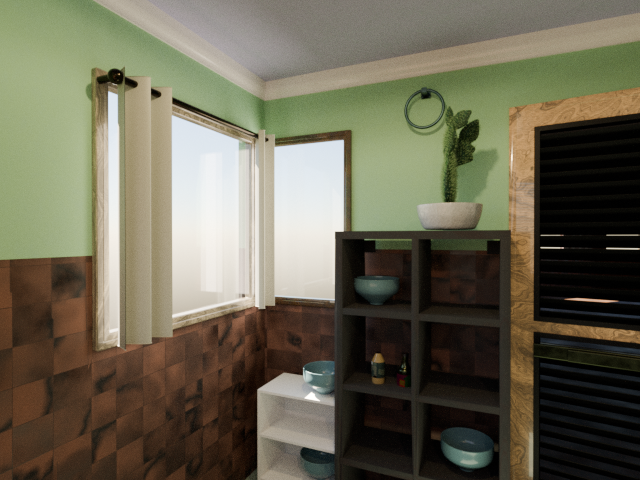}
        \end{minipage} \\ \hline
    \textbf{QA Content}  & 
        \begin{minipage}{0.72\textwidth}
            \textbf{Q:} How many bottle are there in total shown in these pictures? Options: A: 2, B: 4, C: 3, D: 5 \par
            \textbf{A:} D: 5
        \end{minipage} \\ \hline
    \end{tabular}
\end{table}
\begin{table}[!htbp]
    \centering
    \captionof{table}{Definition and illustrative examples for Object Orientation in Level III.}
    \label{tab:object_orientation}
    \small
    \renewcommand{\arraystretch}{1.5}
    \begin{tabular}{|l|p{0.75\textwidth}|}
    \hline
    \textbf{Category}   & Object Orientation \\ \hline
    \textbf{Definition} & {Determine the 3D orientation of a specific object in a target view, given its orientation in a reference view.} \\ \hline
    \multicolumn{2}{|l|}{\cellcolor[HTML]{F5F5F5}\textbf{Example 1}} \\ \hline
    \textbf{Input Image} & 
        \begin{minipage}{0.72\textwidth}
            \centering
            \includegraphics[width=0.48\linewidth]{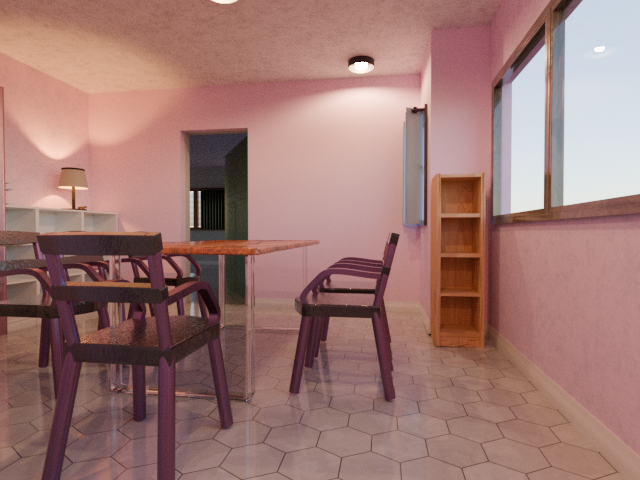} \hfill
            \includegraphics[width=0.48\linewidth]{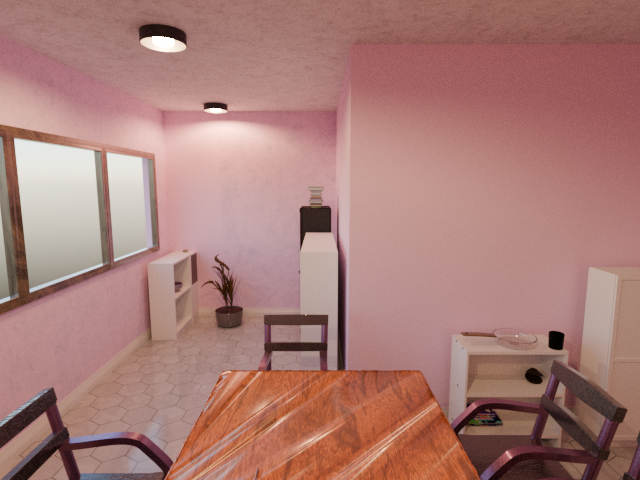}
        \end{minipage} \\ \hline
    \textbf{QA Content}  & 
        \begin{minipage}{0.72\textwidth}
            \textbf{Q:} If the part captured in Figure 1 is the left side of the table dining, then which side of the table dining is shown in Figure 2? Options: A: right, B: front, C: left, D: back \par
            \textbf{A:} A: right
        \end{minipage} \\ \hline
    \end{tabular}
\end{table}
\small
\renewcommand{\arraystretch}{1.5}
\begin{longtable}{|l|p{0.78\textwidth}|}
\caption{Definition and illustrative examples for Size Comparison in Level III.}
\label{tab:size_comparison_3} \\

\hline
\textbf{Category} & \textbf{Size Comparison} \\ \hline
\textbf{Definition} &
Compare the physical dimensions of two objects along a specific axis (e.g., height, width), given that they do not appear within the same viewpoint. \\ \hline
\endfirsthead

\hline
\textbf{Category} & \textbf{Size Comparison} \\ \hline
\endhead

\hline
\multicolumn{2}{|r|}{Continued on next page} \\ \hline
\endfoot

\hline
\endlastfoot

\multicolumn{2}{|l|}{\cellcolor[HTML]{F5F5F5}\textbf{Example 1}} \\ \hline

\textbf{Input Image} &
\begin{minipage}{\linewidth}
\centering
\includegraphics[width=0.48\linewidth]{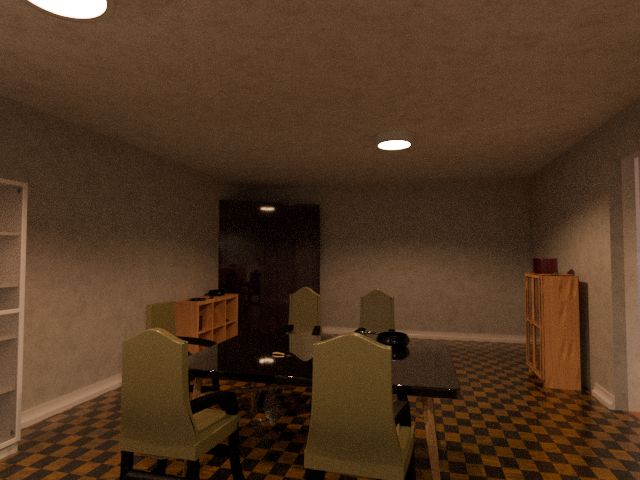}
\hfill
\includegraphics[width=0.48\linewidth]{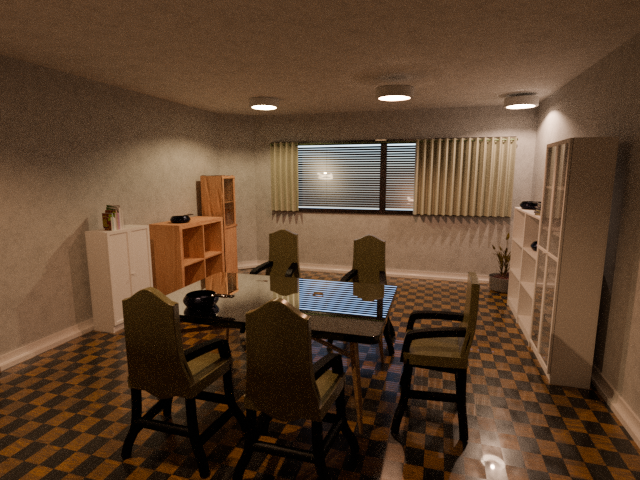}
\end{minipage} \\ \hline

\textbf{QA Content} &
\begin{minipage}{\linewidth}
\textbf{Q:} Which is taller, the height of the cell shelf in Figure 1 or the height of the cell shelf in Figure 2?  
Options:  
A: The height of the cell shelf in Figure 1  
B: Sometimes the former, sometimes the latter  
C: The same height  
D: The height of the cell shelf in Figure 2

\textbf{A:} D: The height of the cell shelf in Figure 2
\end{minipage} \\ \hline

\multicolumn{2}{|l|}{\cellcolor[HTML]{F5F5F5}\textbf{Example 2}} \\ \hline

\textbf{Input Image} &
\begin{minipage}{\linewidth}
\centering
\includegraphics[width=0.48\linewidth]{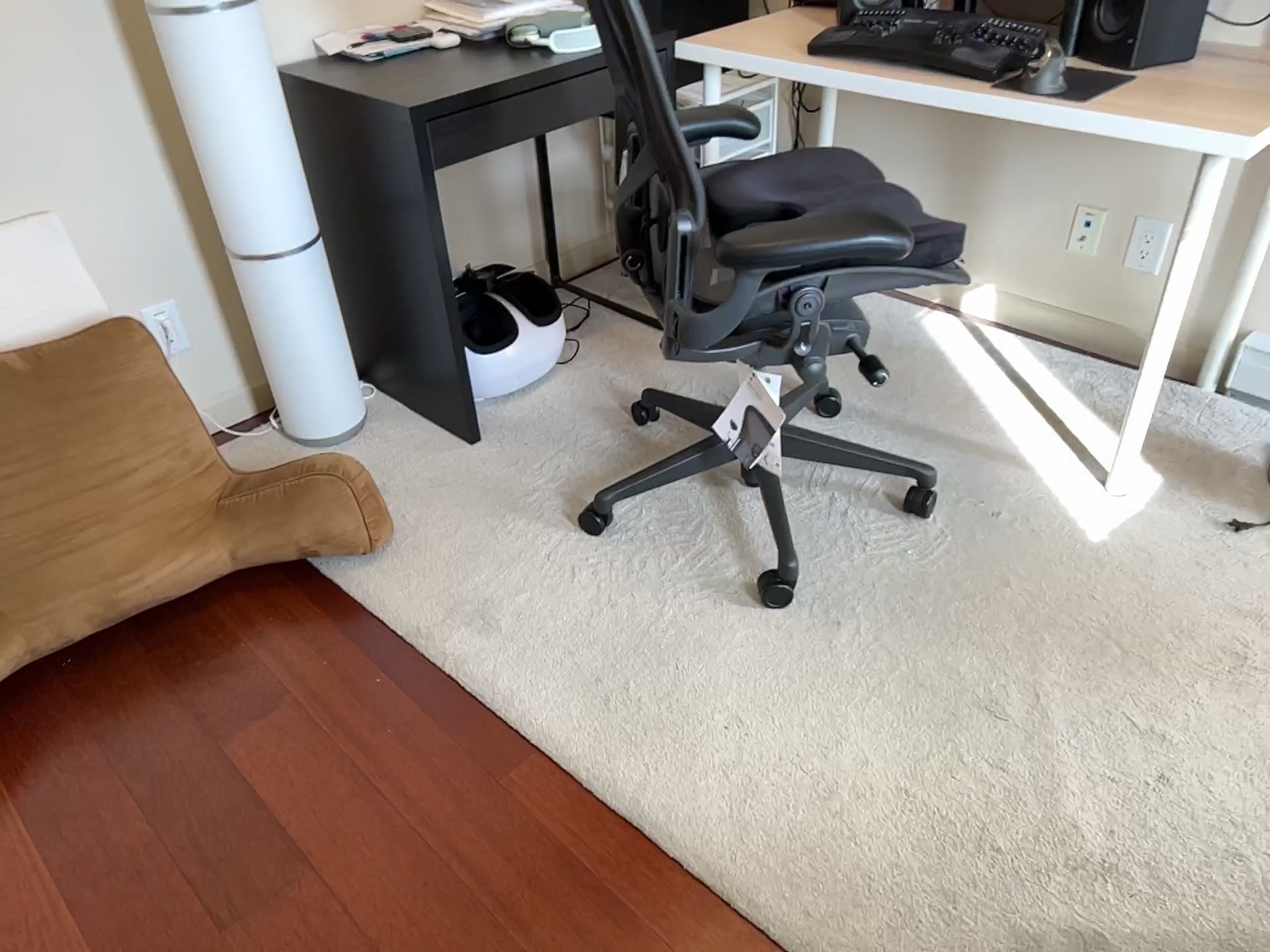}
\hfill
\includegraphics[width=0.48\linewidth]{images/assets/scannetpp/a24f64f7fb_iphone/frame_002630.jpg}
\end{minipage} \\ \hline

\textbf{QA Content} &
\begin{minipage}{\linewidth}
\textbf{Q:} Which has a greater height, the height of the sit-up pillow in Figure 1 or the height of the monitor in Figure 2?  
Options:  
A: The height of the sit-up pillow in Figure 1  
B: The height of the monitor in Figure 2  
C: Sometimes the former, sometimes the latter  
D: The same height

\textbf{A:} B: The height of the monitor in Figure 2
\end{minipage} \\ \hline

\multicolumn{2}{|l|}{\cellcolor[HTML]{F5F5F5}\textbf{Example 3}} \\ \hline

\textbf{Input Image} &
\begin{minipage}{\linewidth}
\centering
\includegraphics[width=0.48\linewidth]{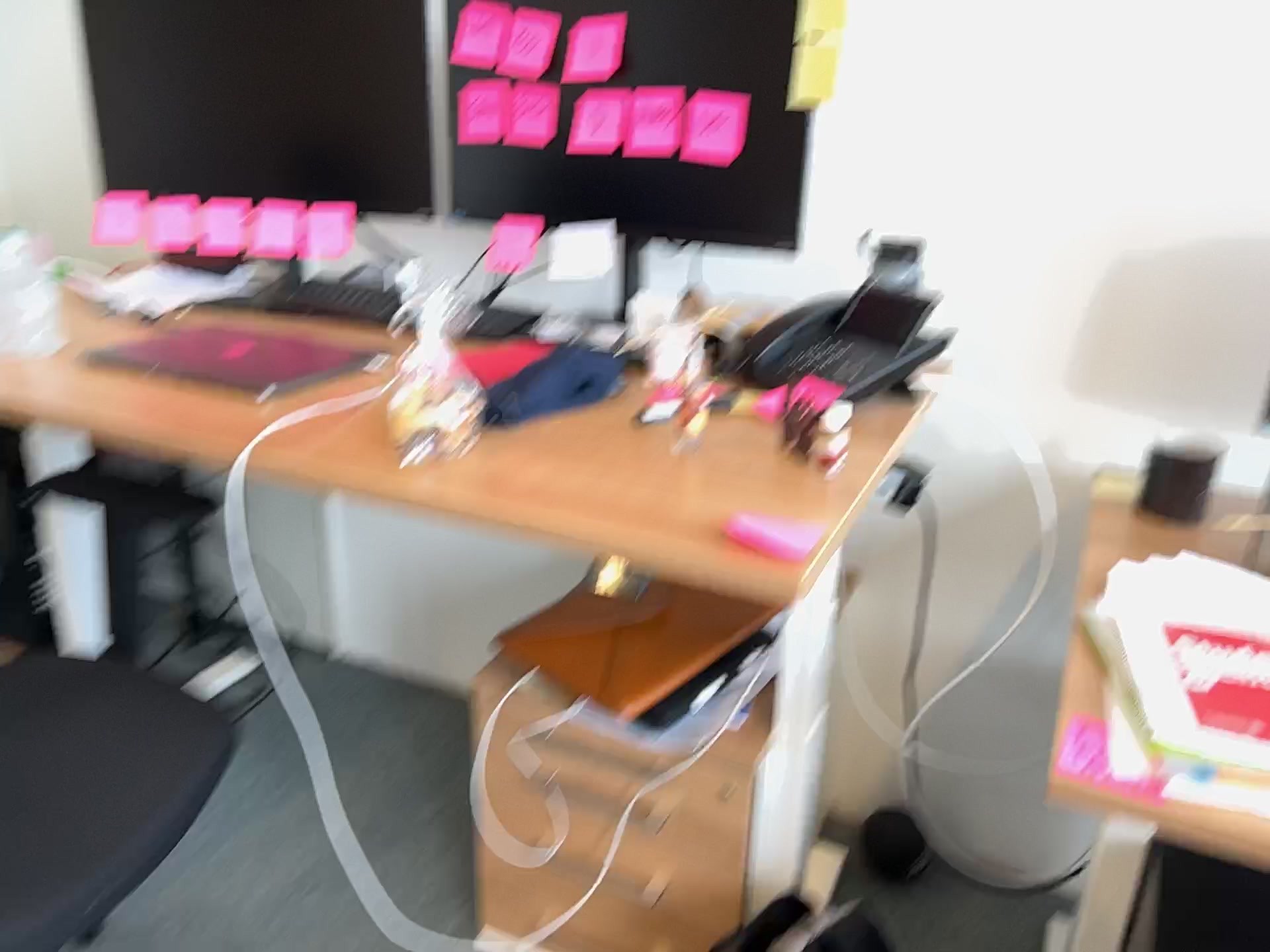}
\hfill
\includegraphics[width=0.48\linewidth]{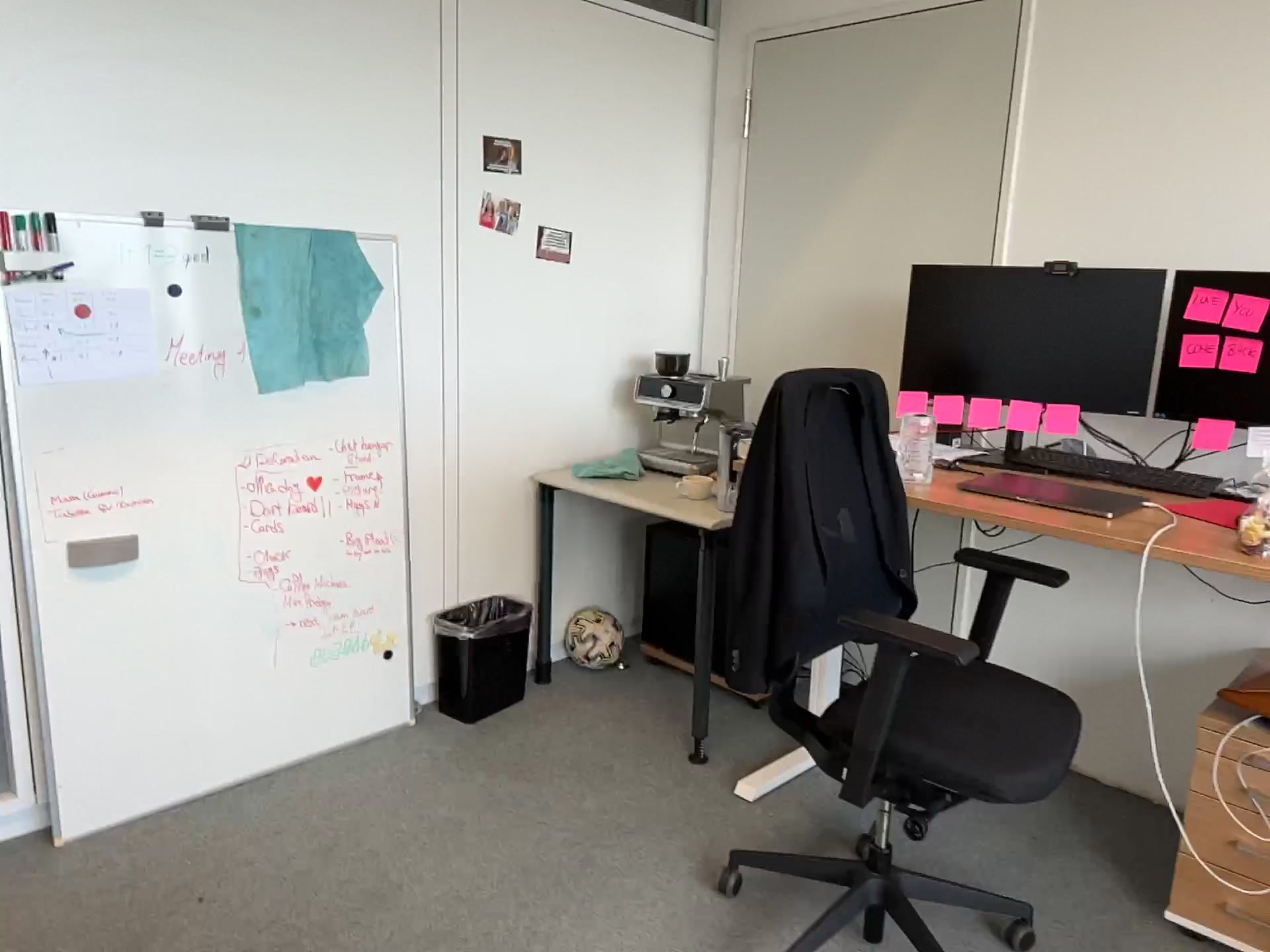}
\end{minipage} \\ \hline

\textbf{QA Content} &
\begin{minipage}{\linewidth}
\textbf{Q:} Which is narrower in width, the width of the rightmost book in Figure 1 or the width of the bin in Figure 2?  
Options:  
A: The same length  
B: The width of the bin in Figure 2  
C: The width of the rightmost book in Figure 1  
D: Sometimes the former, sometimes the latter

\textbf{A:} C: The width of the rightmost book in Figure 1
\end{minipage} \\ \hline

\end{longtable}

\small
\renewcommand{\arraystretch}{1.5}
\begin{longtable}{|l|p{0.78\textwidth}|}
\caption{Definition and illustrative examples for Camera Motion in Level III.}
\label{tab:camera_motion} \\

\hline
\textbf{Category} & \textbf{Camera Motion} \\ \hline
\textbf{Definition} &
Estimate the relative 6-DOF transformation (rotation and translation) between two camera viewpoints to recover the camera's movement path. \\ \hline
\endfirsthead

\hline
\textbf{Category} & \textbf{Camera Motion} \\ \hline
\endhead

\hline
\multicolumn{2}{|r|}{Continued on next page} \\ \hline
\endfoot

\hline
\endlastfoot
\multicolumn{2}{|l|}{\cellcolor[HTML]{F5F5F5}\textbf{Example 1}} \\ \hline
\textbf{Input Image} & 
    \begin{minipage}{0.72\textwidth}
        \centering
        \includegraphics[width=0.48\linewidth]{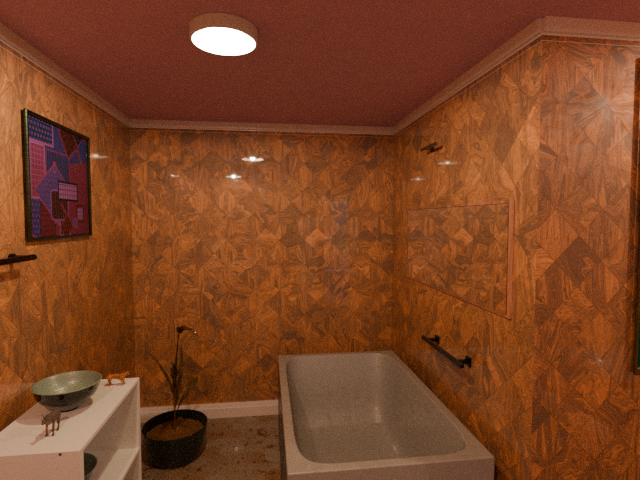} \hfill
        \includegraphics[width=0.48\linewidth]{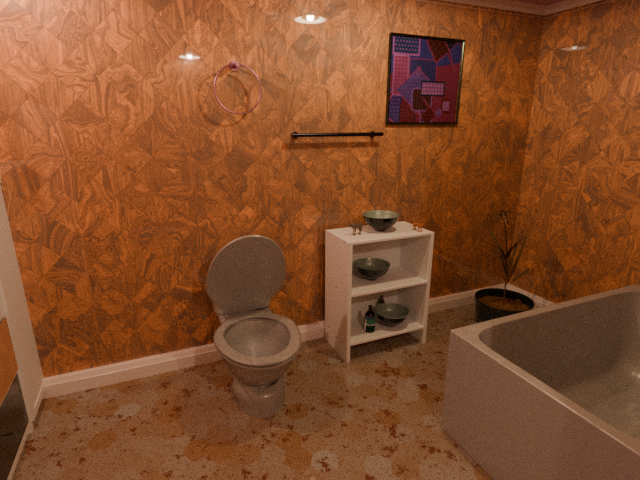} \hfill
        \includegraphics[width=0.48\linewidth]{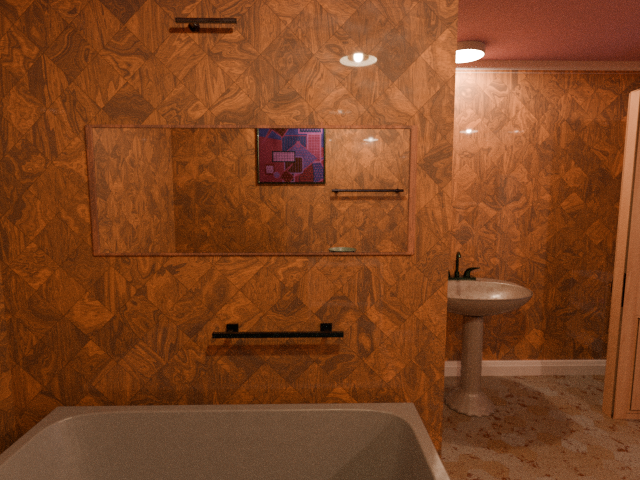}
    \end{minipage} \\ \hline
\textbf{QA Content}  & 
    \begin{minipage}{0.72\textwidth}
        \textbf{Q:} Assuming that the camera in Photo 3 is facing east, if I am currently taking Photo 1, then what is the direction of the camera relative to me when taking Photo 2? Options: A: north, B: west, C: east, D: southwest \par
        \textbf{A:} C: east
    \end{minipage} \\ \hline
\multicolumn{2}{|l|}{\cellcolor[HTML]{F5F5F5}\textbf{Example 2}} \\ \hline
\textbf{Input Image} & 
    \begin{minipage}{0.72\textwidth}
        \centering
        \includegraphics[width=0.48\linewidth]{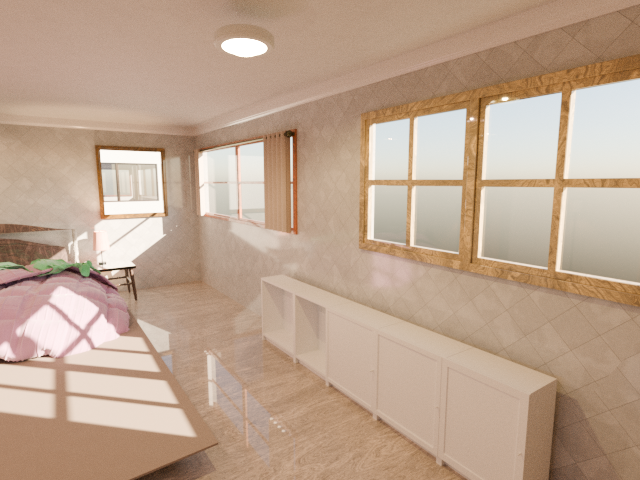} \hfill
        \includegraphics[width=0.48\linewidth]{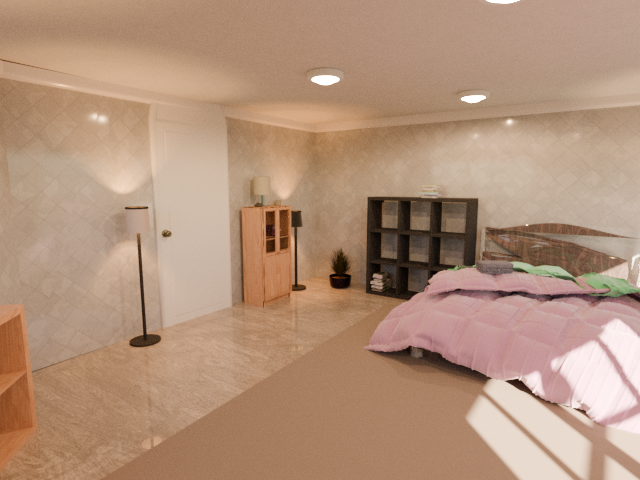}
    \end{minipage} \\ \hline
\textbf{QA Content}  & 
    \begin{minipage}{0.72\textwidth}
        \textbf{Q:} The camera coordinate system is defined as +Z up, +X forward, and is a right-handed system. How can the viewpoint in the second image be obtained from the viewpoint in the first image in its corresponding camera coordinate system? Options: A: rotate a negative angle around the Z-axis, B: rotate a negative angle around the Y-axis, C: rotate a positive angle around the Y-axis, D: rotate a positive angle around the Z-axis \par
        \textbf{A:} A: rotate a negative angle around the Z-axis
    \end{minipage} \\ \hline
\multicolumn{2}{|l|}{\cellcolor[HTML]{F5F5F5}\textbf{Example 3}} \\ \hline
\textbf{Input Image} & 
    \begin{minipage}{0.72\textwidth}
        \centering
        \includegraphics[width=0.48\linewidth]{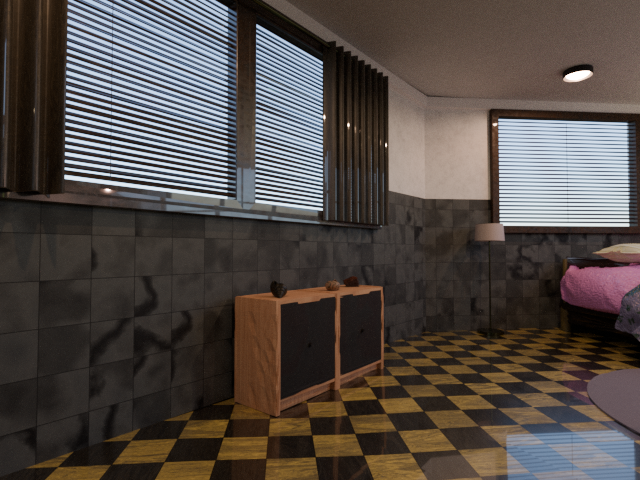} \hfill
        \includegraphics[width=0.48\linewidth]{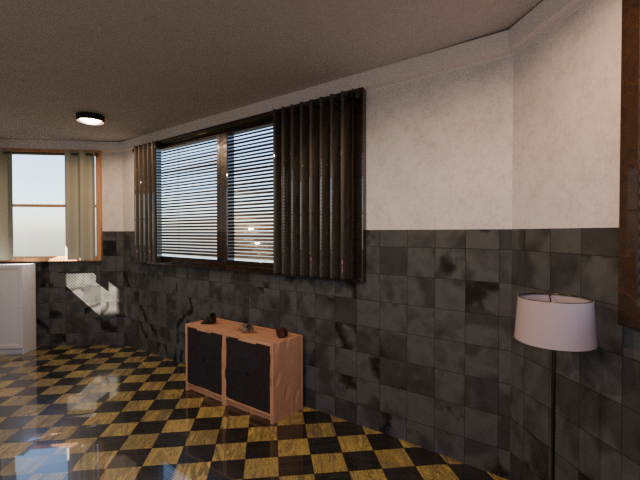}
    \end{minipage} \\ \hline
\textbf{QA Content}  & 
    \begin{minipage}{0.72\textwidth}
        \textbf{Q:} The image is a continuous first-person perspective shot. In which direction are you moving? Options: A: backward to the right, B: Not moving, C: forward to the right, D: backward to the left \par
        \textbf{A:} C: forward to the right
    \end{minipage} \\ \hline
\multicolumn{2}{|l|}{\cellcolor[HTML]{F5F5F5}\textbf{Example 4}} \\ \hline
\textbf{Input Image} & 
    \begin{minipage}{0.72\textwidth}
        \centering
        \includegraphics[width=0.48\linewidth]{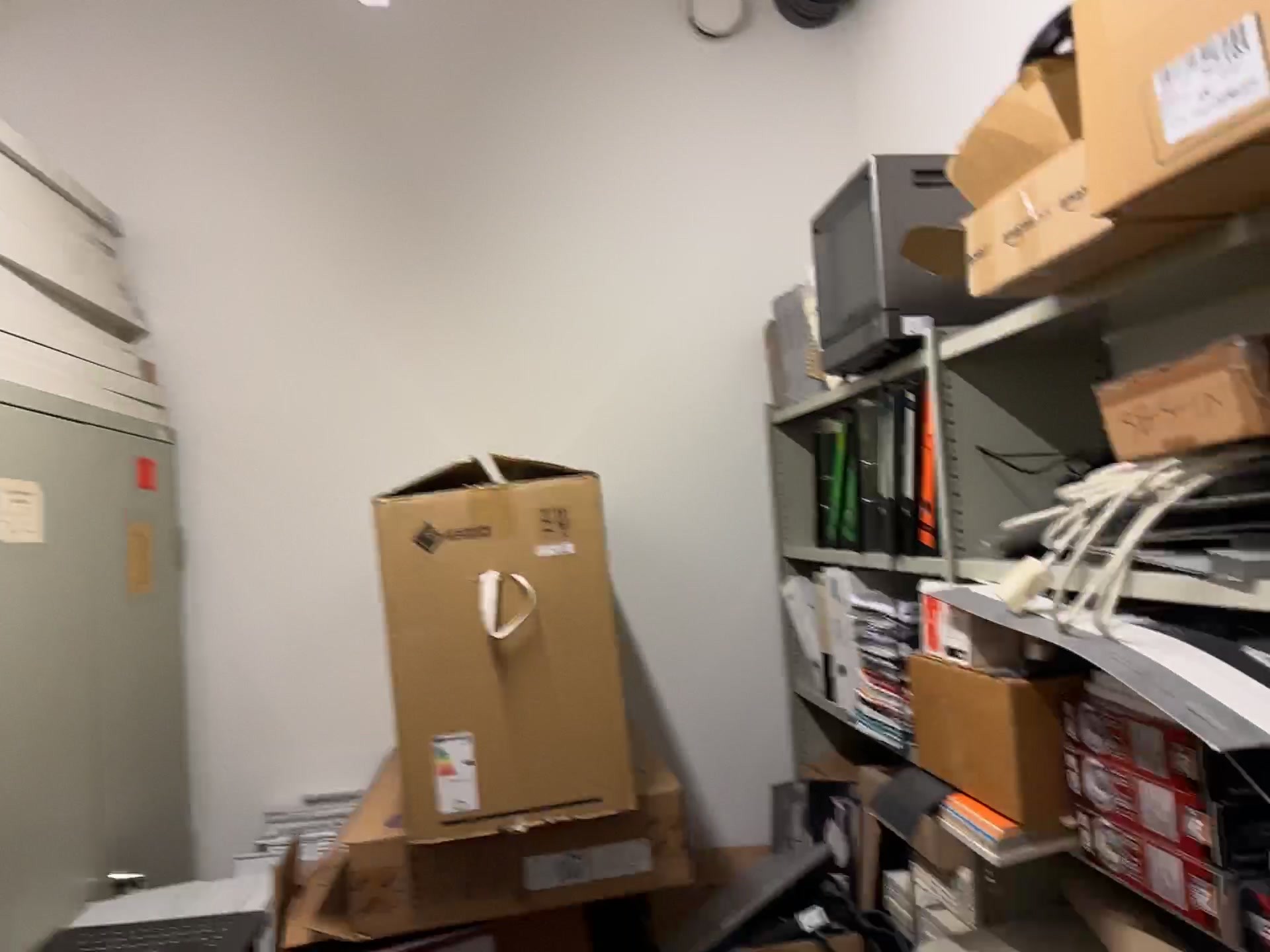} \hfill
        \includegraphics[width=0.48\linewidth]{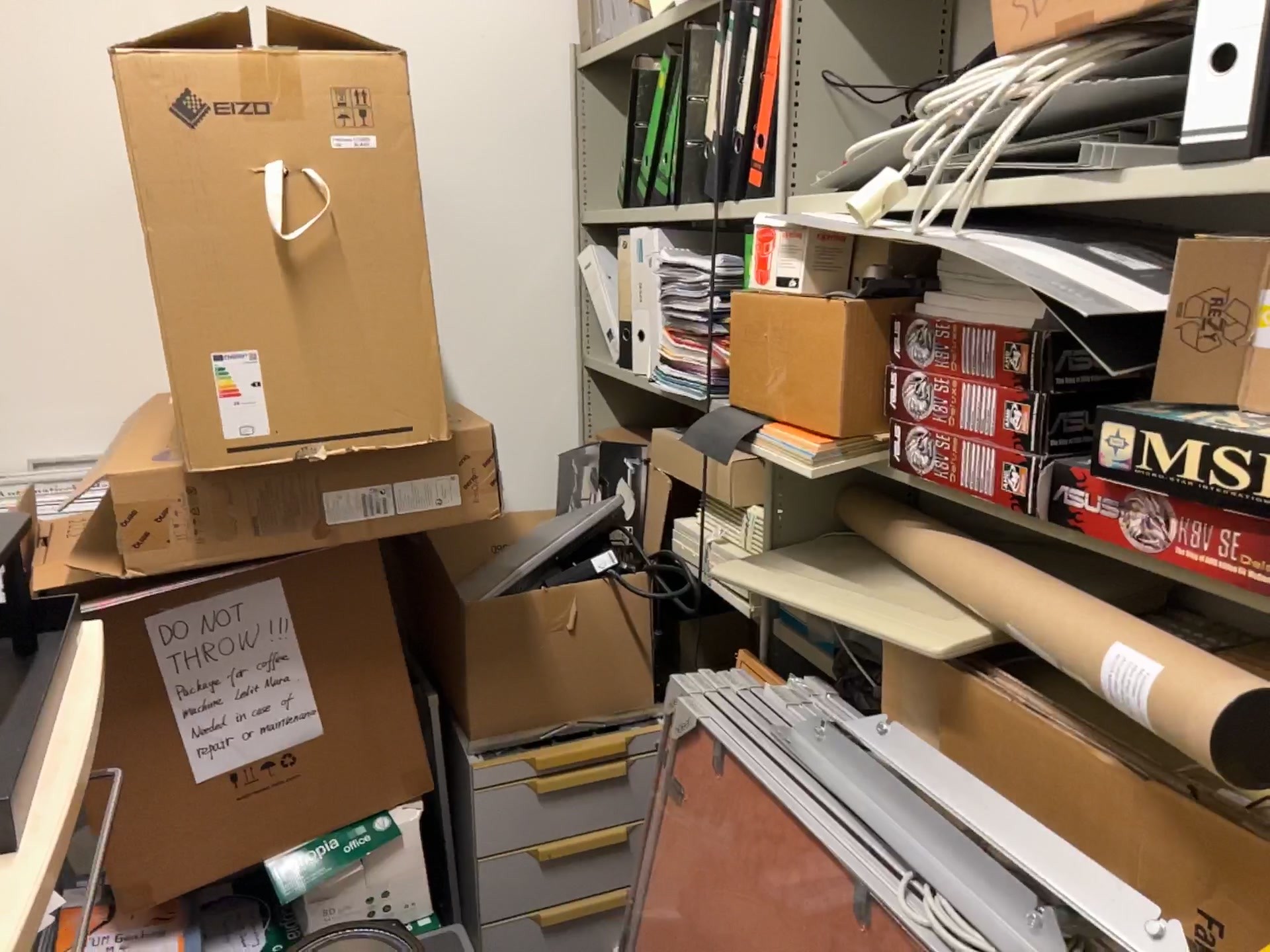}
    \end{minipage} \\ \hline
\textbf{QA Content}  & 
    \begin{minipage}{0.72\textwidth}
        \textbf{Q:} In a continuous first-person perspective shot, in which direction is the camera rotating? Options: A: lower left, B: up, C: upper right, D: lower right \par
        \textbf{A:} D: lower right
    \end{minipage} \\ \hline
\multicolumn{2}{|l|}{\cellcolor[HTML]{F5F5F5}\textbf{Example 5}} \\ \hline
\textbf{Input Image} & 
    \begin{minipage}{0.72\textwidth}
        \centering
        \includegraphics[width=0.48\linewidth]{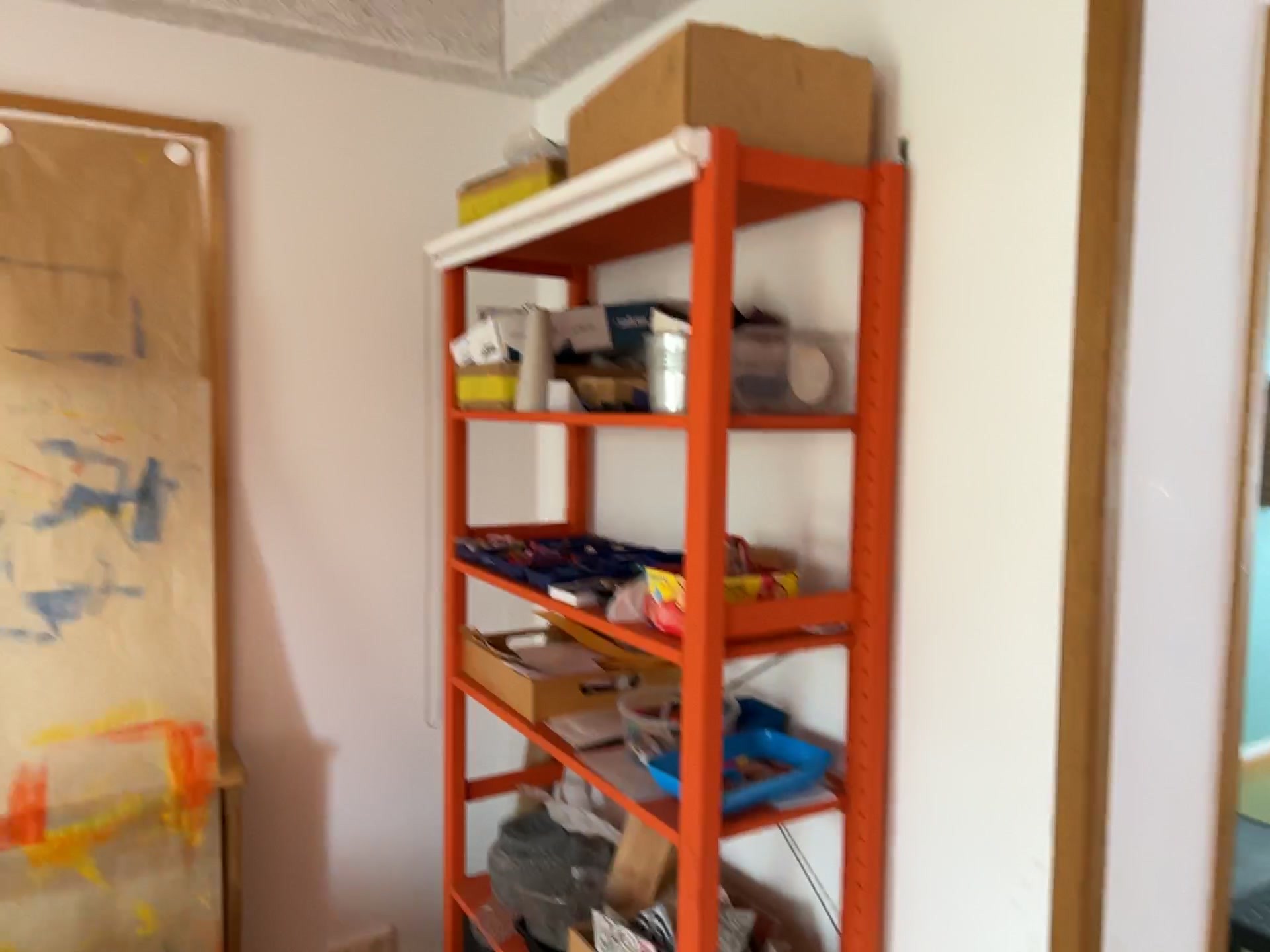} \hfill
        \includegraphics[width=0.48\linewidth]{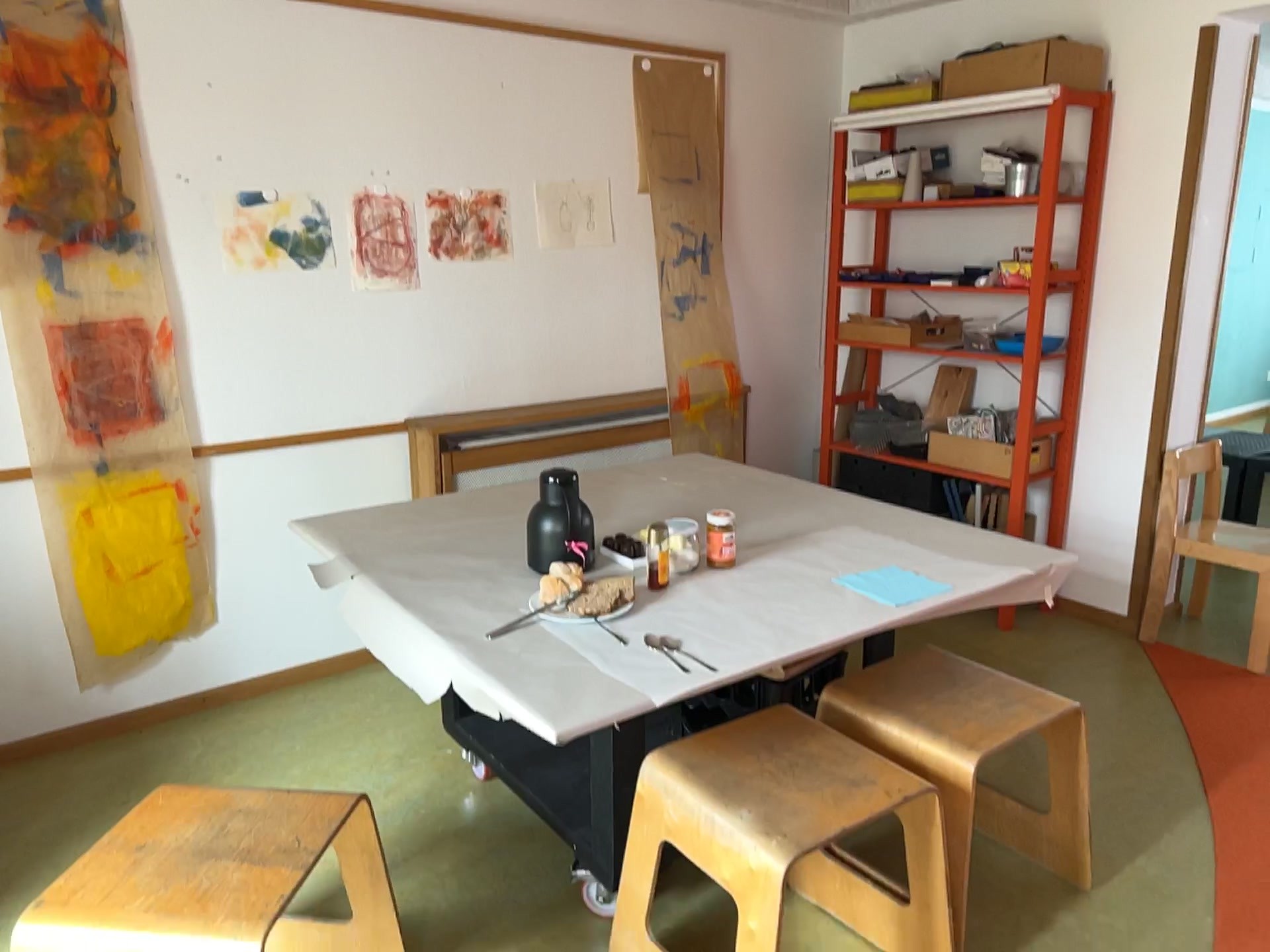}
    \end{minipage} \\ \hline
\textbf{QA Content}  & 
    \begin{minipage}{0.72\textwidth}
        \textbf{Q:} What is the position of the first photo relative to the position where the second photo was taken, using the direction faced when taking the second photo as the front? Options: A: backward left, B: forward left, C: forward right, D: backward right \par
        \textbf{A:} C: forward right
    \end{minipage} \\ \hline
\end{longtable}
\begin{table}[!htbp]
    \centering
    \captionof{table}{Definition and illustrative examples for Position Relationship (Cam.-Obj.) in Level III.}
    \label{tab:position_relationship_cam_obj}
    \small
    \renewcommand{\arraystretch}{1.5}
    \begin{tabular}{|l|p{0.75\textwidth}|}
    \hline
    \textbf{Category}   & Position Relationship (Cam.-Obj.) \\ \hline
    \textbf{Definition} & {Given an object that is visible in a secondary view but absent in the reference view, determine its spatial location relative to the reference camera's coordinate system.} \\ \hline
    \multicolumn{2}{|l|}{\cellcolor[HTML]{F5F5F5}\textbf{Example 1}} \\ \hline
    \textbf{Input Image} & 
        \begin{minipage}{0.72\textwidth}
            \centering
            \includegraphics[width=0.48\linewidth]{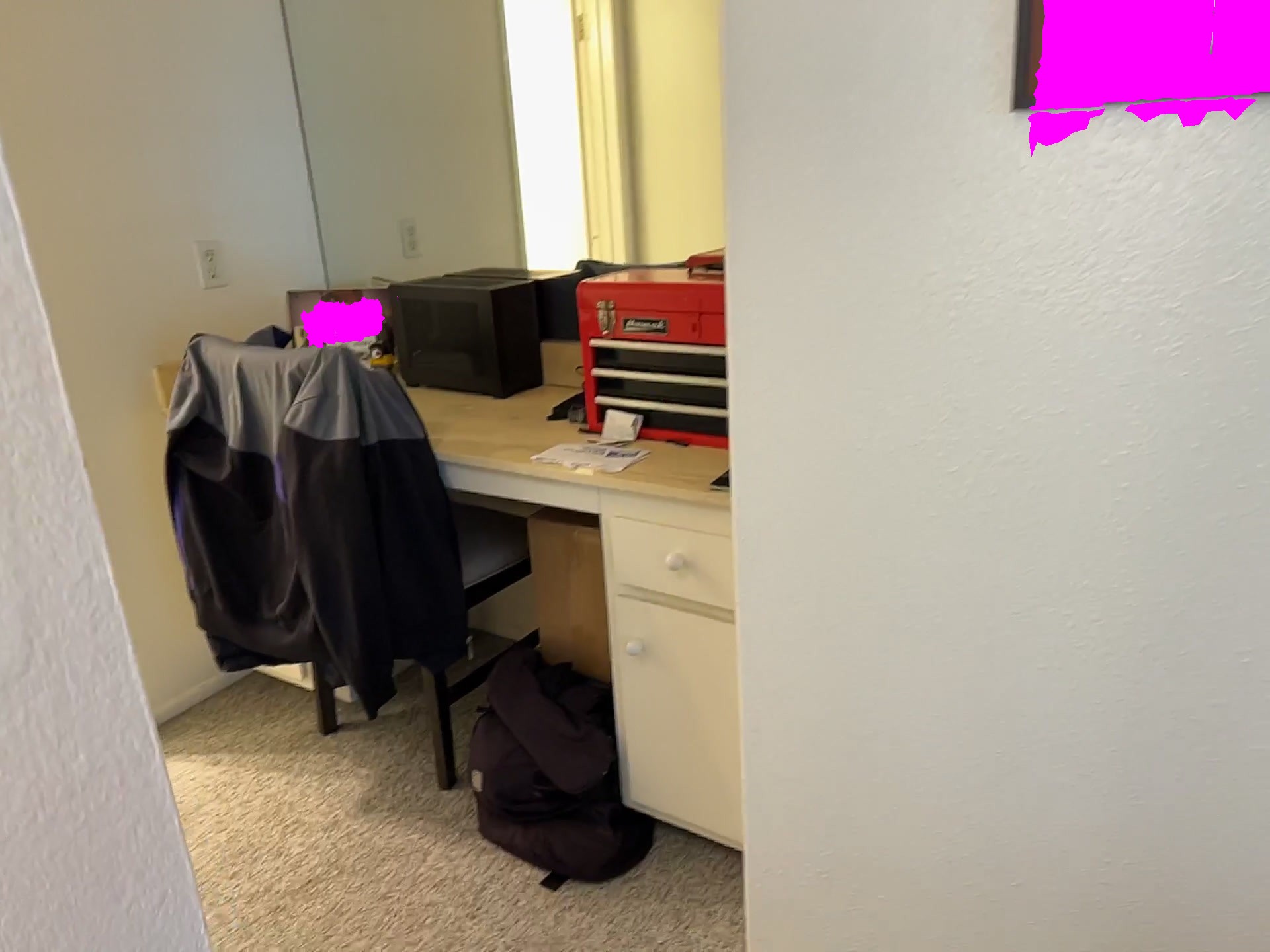} \hfill
            \includegraphics[width=0.48\linewidth]{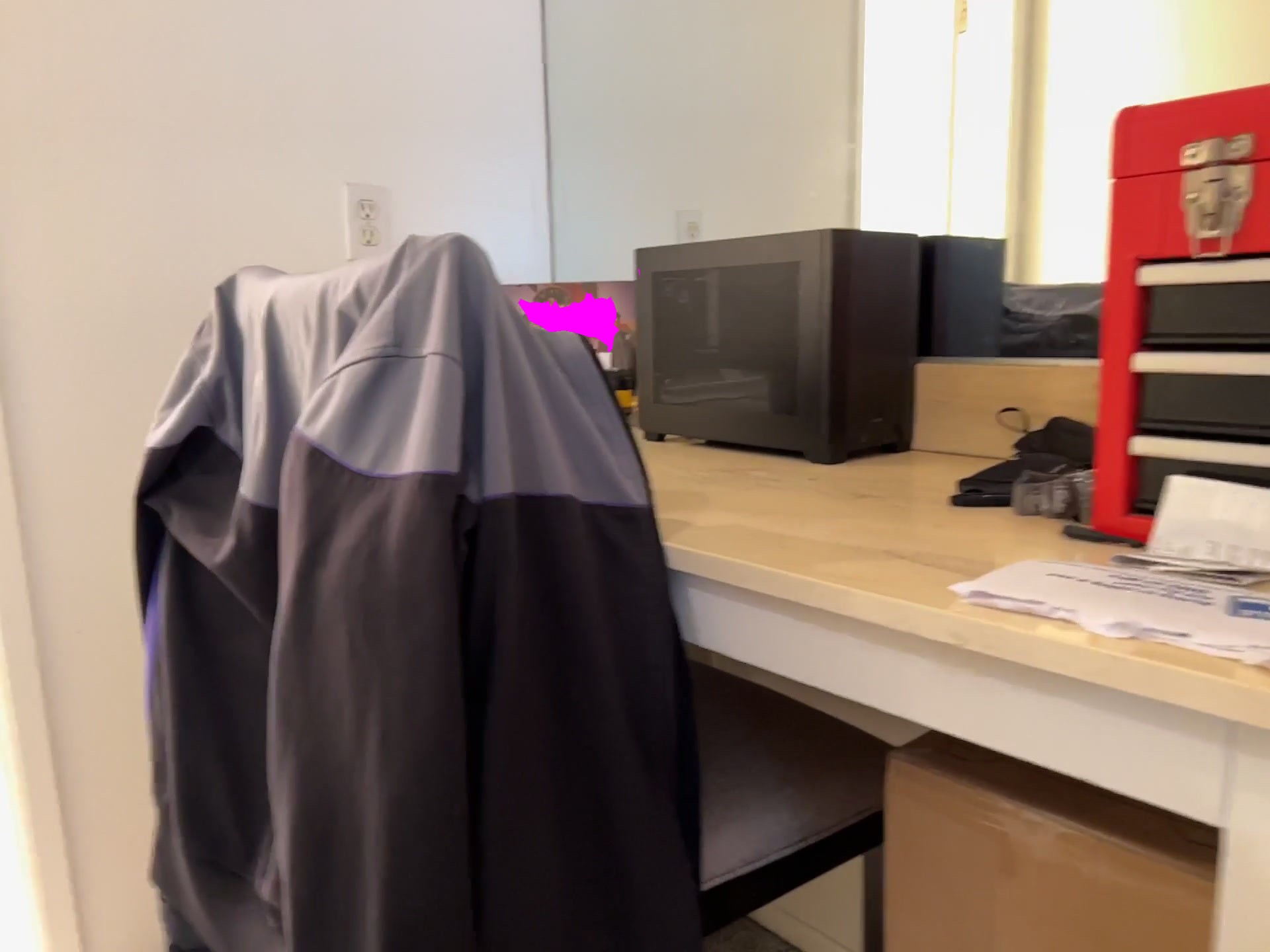}
        \end{minipage} \\ \hline
    \textbf{QA Content}  & 
        \begin{minipage}{0.72\textwidth}
            \textbf{Q:} The pictures are taken continuously from a first-person perspective, with two pictures overlapping at the frame. At the moment of the second picture, in which direction is the cabinet in the first picture relative to you? Options: A: back right, B: back left, C: forward right, D: forward left \par
            \textbf{A:} C: forward right
        \end{minipage} \\ \hline
    \multicolumn{2}{|l|}{\cellcolor[HTML]{F5F5F5}\textbf{Example 2}} \\ \hline
    \textbf{Input Image} & 
        \begin{minipage}{0.72\textwidth}
            \centering
            \includegraphics[width=0.48\linewidth]{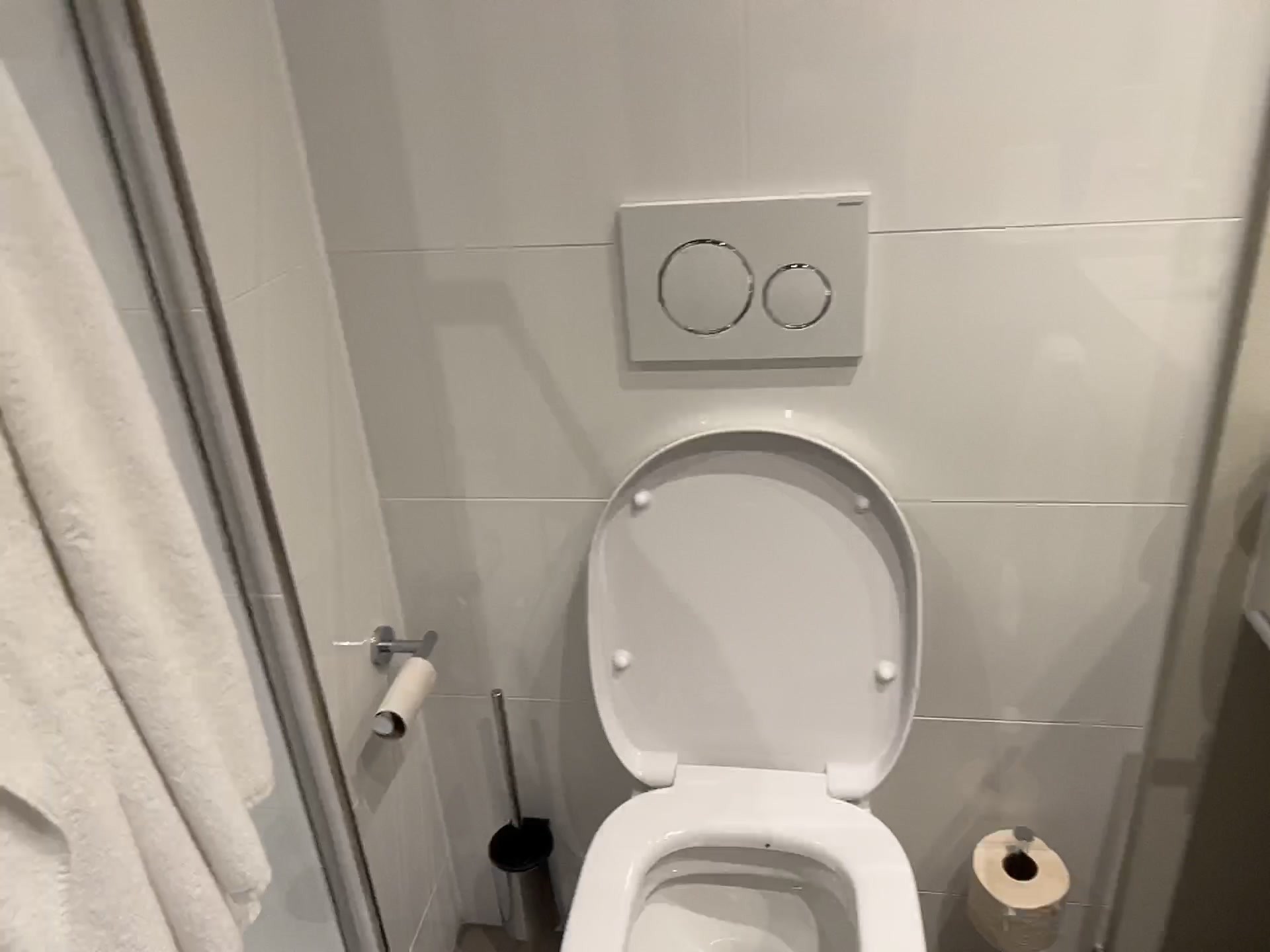} \hfill
            \includegraphics[width=0.48\linewidth]{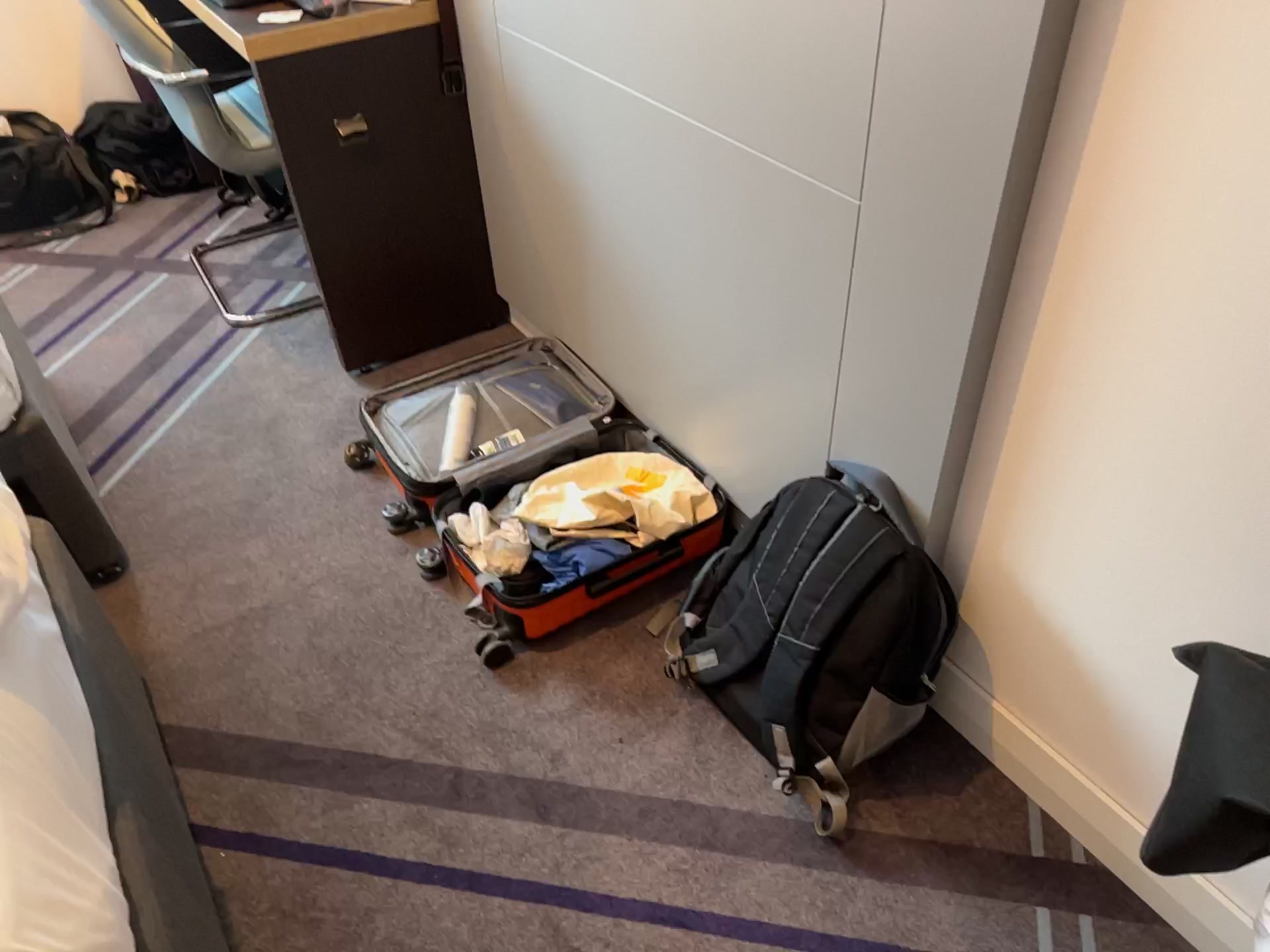} \hfill
            \includegraphics[width=0.48\linewidth]{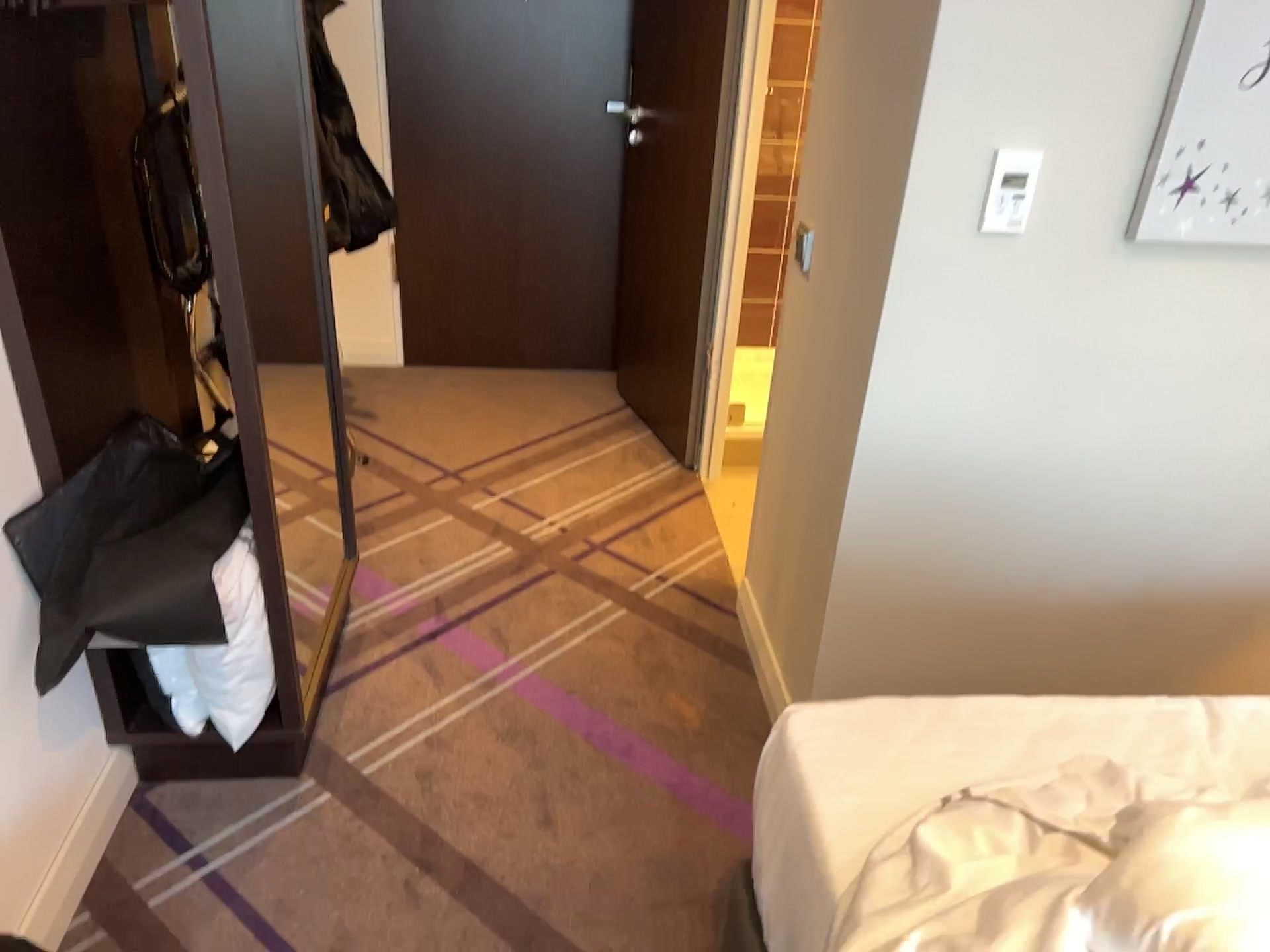}
        \end{minipage} \\ \hline
    \textbf{QA Content}  & 
        \begin{minipage}{0.72\textwidth}
            \textbf{Q:} When you were taking photo third, where was the table in relation to you? Options: A: rear left, B: front right, C: front left, D: rear right \par
            \textbf{A:} A: rear left
        \end{minipage} \\ \hline
    \end{tabular}
\end{table}
\small
\renewcommand{\arraystretch}{1.5}
\begin{longtable}{|l|p{0.78\textwidth}|}
    \caption{Definition and illustrative examples for Position Relationship (Obj.-Obj.) in Level III.}
    \label{tab:position_relationship_obj_obj} \\

    \hline
    \textbf{Category} & \textbf{Position Relationship (Obj.-Obj.)} \\ \hline
    \textbf{Definition} &
    Determine the relative spatial location of a target object with respect to a reference object based on predefined directional references. \\ \hline
    \endfirsthead
    
    \hline
    \textbf{Category} & \textbf{Position Relationship (Obj.-Obj.)} \\ \hline
    \endhead
    
    \hline
    \multicolumn{2}{|r|}{Continued on next page} \\ \hline
    \endfoot
    
    \hline
    \endlastfoot
    \multicolumn{2}{|l|}{\cellcolor[HTML]{F5F5F5}\textbf{Example 1}} \\ \hline
    \textbf{Input Image} & 
        \begin{minipage}{0.72\textwidth}
            \centering
            \includegraphics[width=0.48\linewidth]{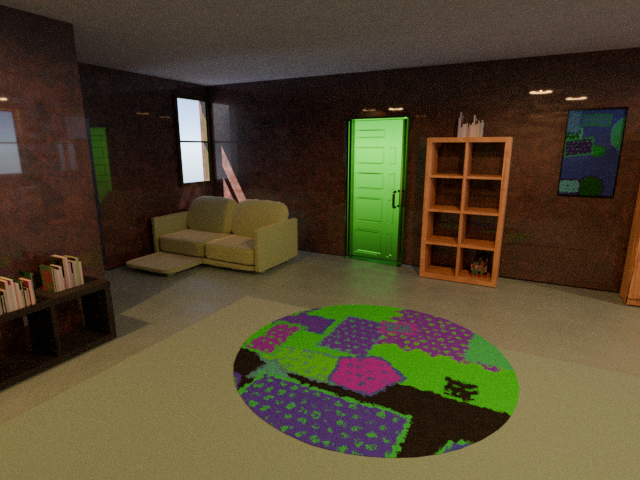} \hfill
            \includegraphics[width=0.48\linewidth]{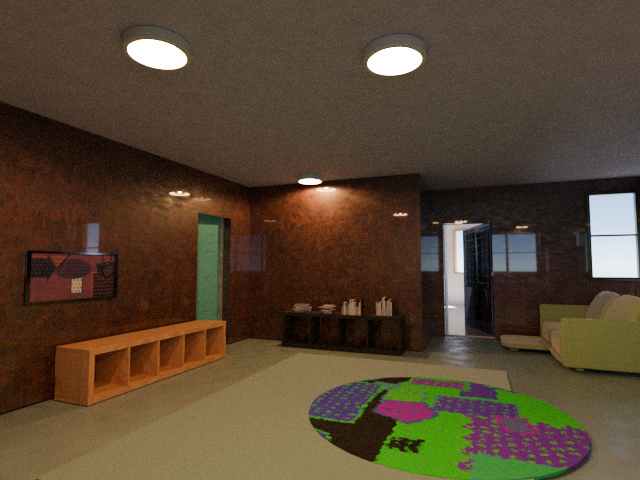}
        \end{minipage} \\ \hline
    \textbf{QA Content}  & 
        \begin{minipage}{0.72\textwidth}
            \textbf{Q:} If the direction in which rightmost cell shelf in figure 1 is from the leftmost rug in figure 2 is considered the east, what object is directly to the northeast of the leftmost rug in figure 2? Options: A: rightmost rug in figure 1, B: sofa in figure 1, C: leftmost cell shelf in figure 1, D: lite door in figure 2 \par
            \textbf{A:} B: sofa in figure 1
        \end{minipage} \\ \hline
    \multicolumn{2}{|l|}{\cellcolor[HTML]{F5F5F5}\textbf{Example 2}} \\ \hline
    \textbf{Input Image} & 
        \begin{minipage}{0.72\textwidth}
            \centering
            \includegraphics[width=0.48\linewidth]{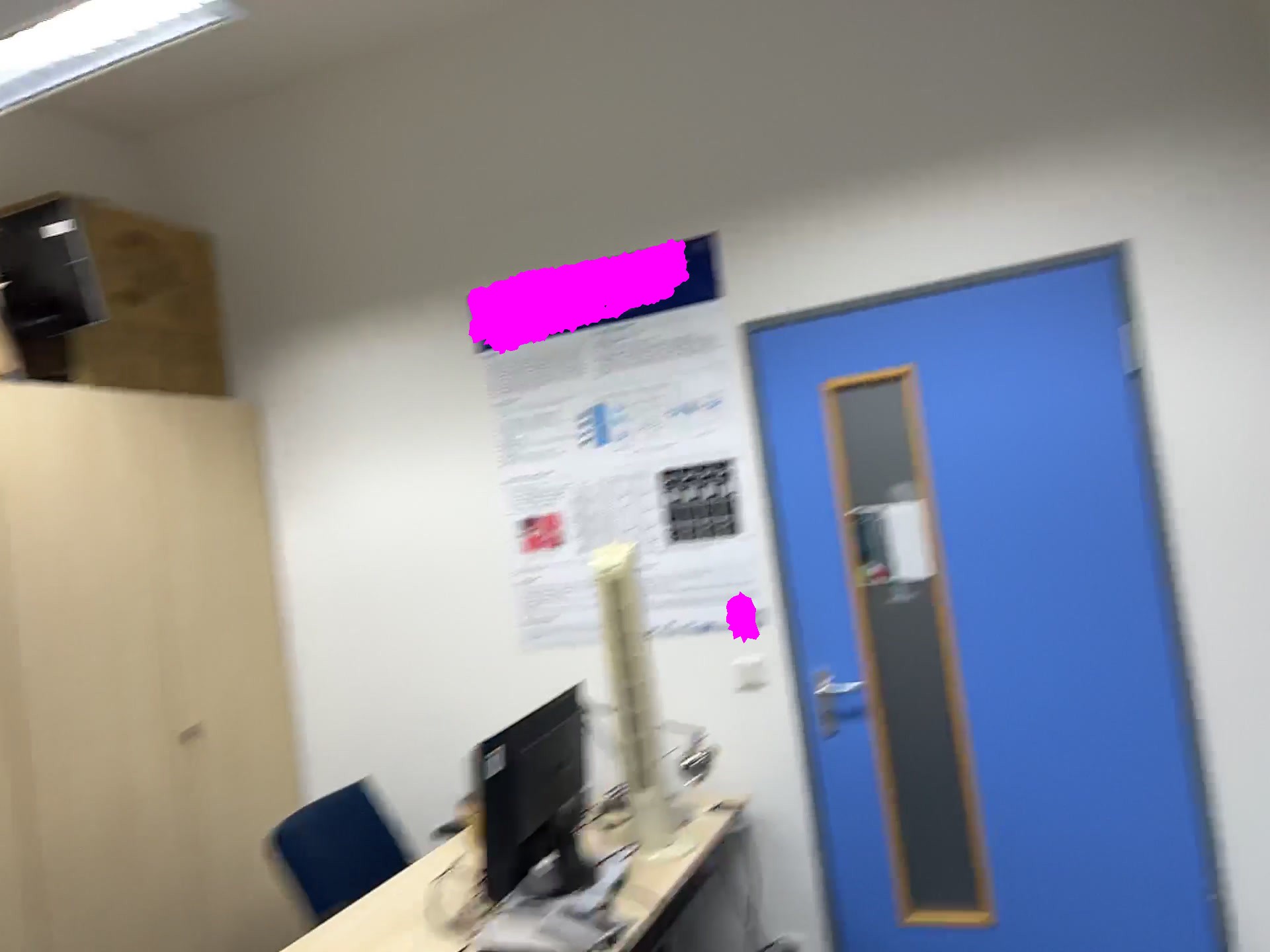} \hfill
            \includegraphics[width=0.48\linewidth]{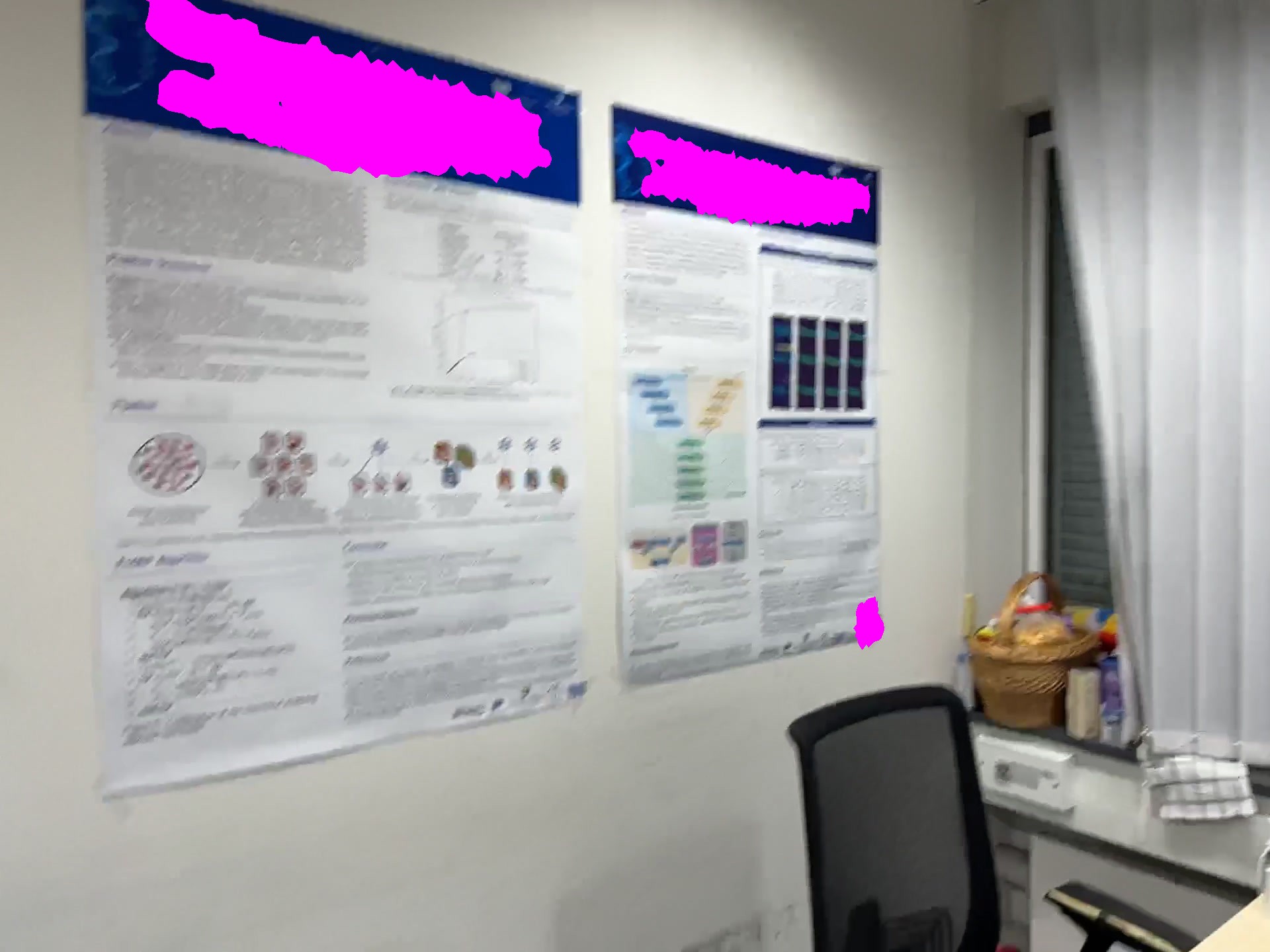} \hfill
            \includegraphics[width=0.48\linewidth]{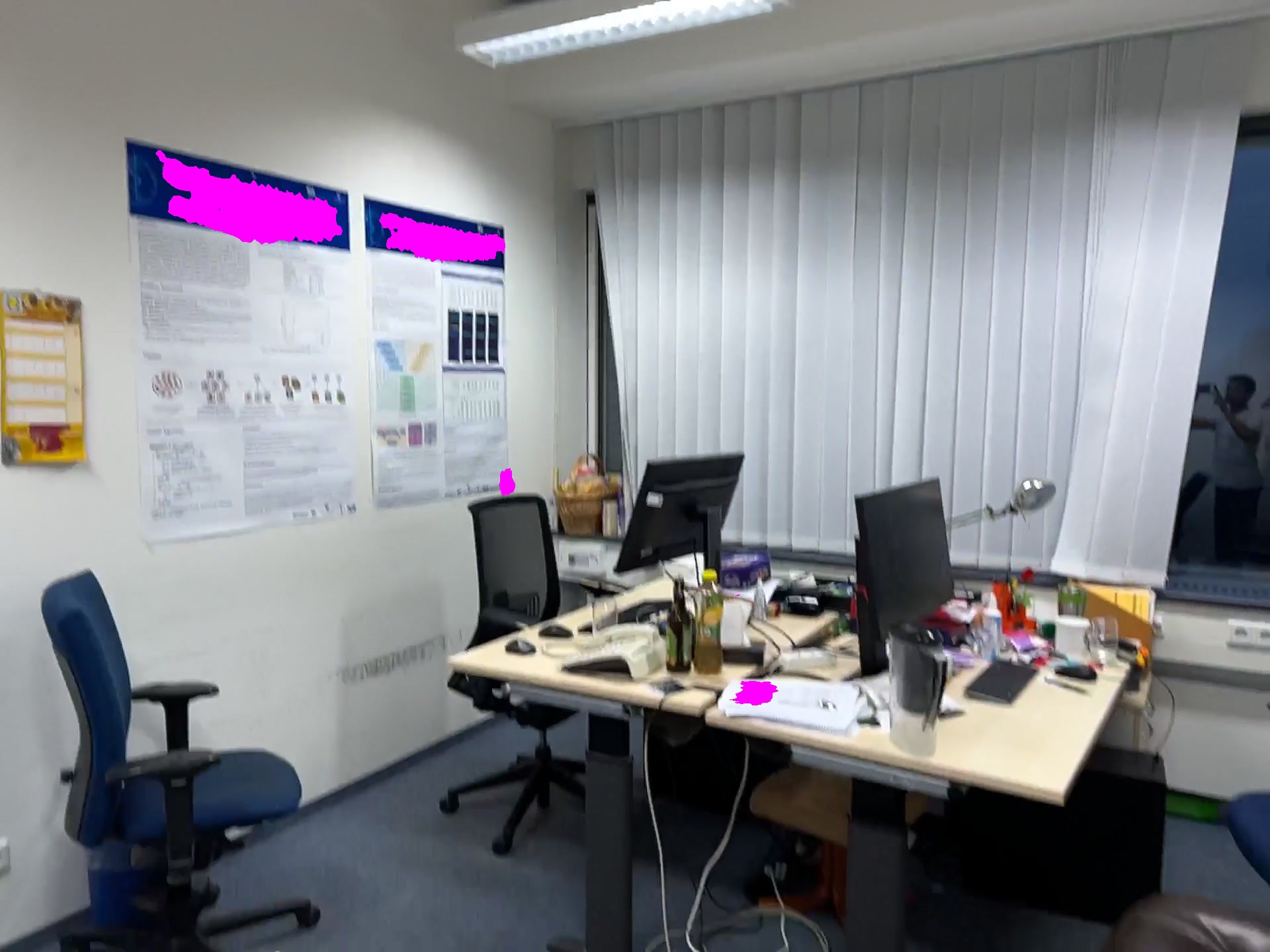}
        \end{minipage} \\ \hline
    \textbf{QA Content}  & 
        \begin{minipage}{0.72\textwidth}
            \textbf{Q:} The images are taken continuously from a first-person perspective, with the storage cabinet on the south side of the door frame. In which direction is the leftmost monitor in frame 3 located relative to the door frame? Options: A: south, B: northeast, C: southeast, D: east \par
            \textbf{A:} D: east
        \end{minipage} \\ \hline
    \multicolumn{2}{|l|}{\cellcolor[HTML]{F5F5F5}\textbf{Example 3}} \\ \hline
    \textbf{Input Image} & 
        \begin{minipage}{0.72\textwidth}
            \centering
            \includegraphics[width=0.48\linewidth]{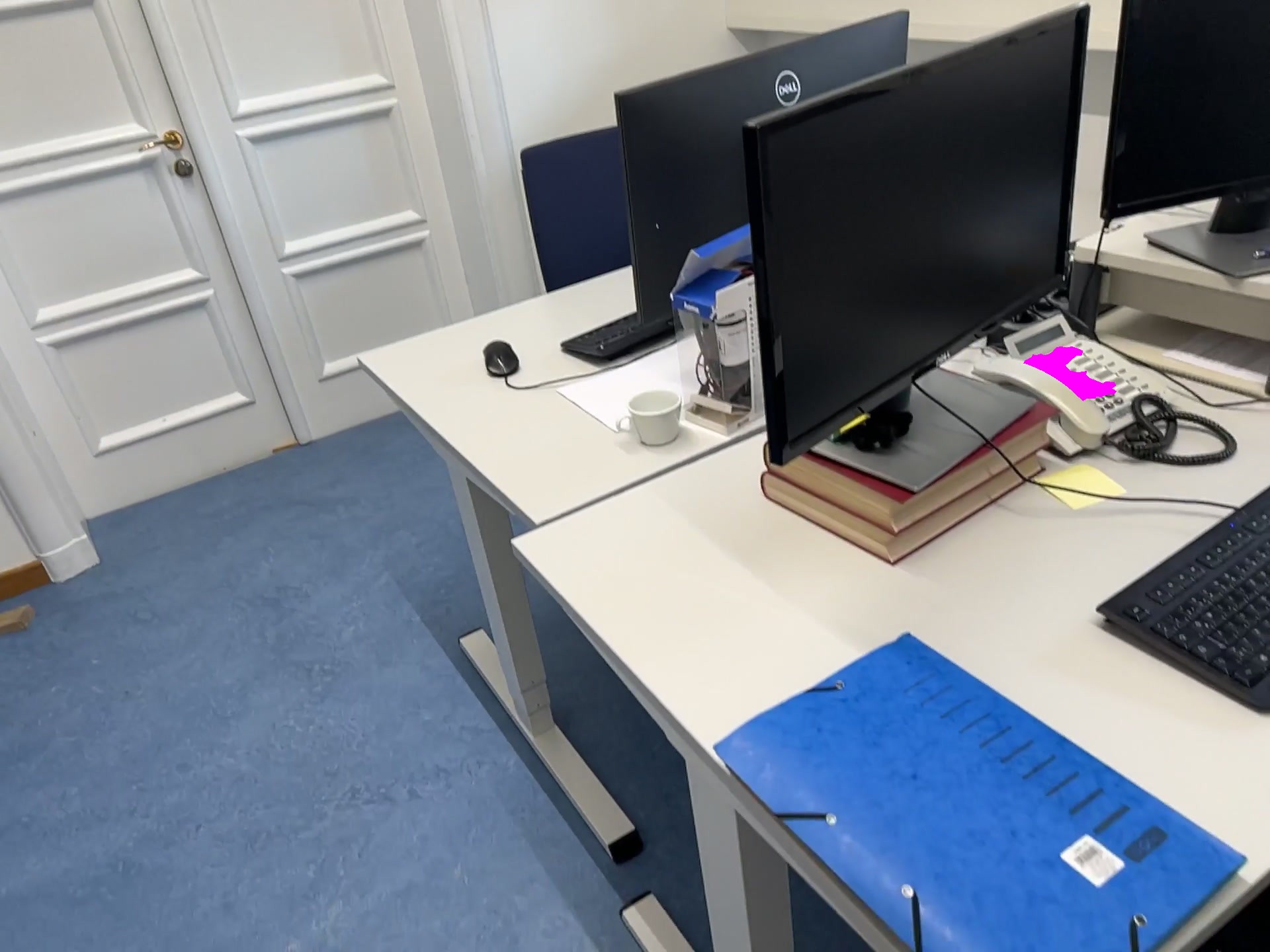} \hfill
            \includegraphics[width=0.48\linewidth]{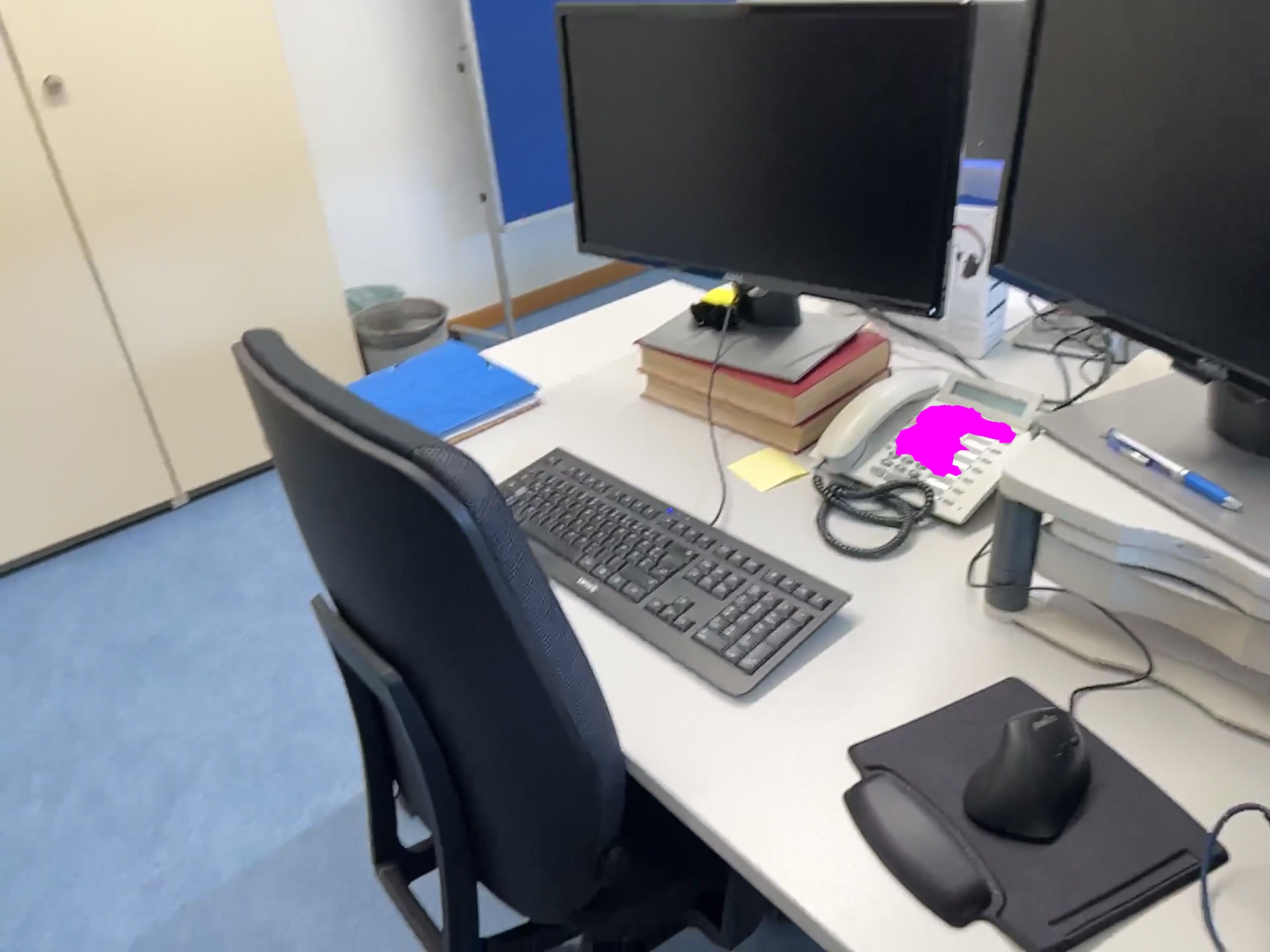}
        \end{minipage} \\ \hline
    \textbf{QA Content}  & 
        \begin{minipage}{0.72\textwidth}
            \textbf{Q:} In what direction is the office chair in figure 2 relative to the telephone in figure 2 (knowing that the office chair in figure 1 is in the left of the telephone in figure 2, and these two pictures overlap)? Options: A: right, B: back, C: front, D: left \par
            \textbf{A:} A: right
        \end{minipage} \\ \hline
\end{longtable}
\begin{table}[!htbp]
    \centering
    \captionof{table}{Definition and illustrative examples for Virtual Perspective in Level III.}
    \label{tab:virtual_perspective}
    \small
    \renewcommand{\arraystretch}{1.5}
    \begin{tabular}{|l|p{0.75\textwidth}|}
    \hline
    \textbf{Category}   & Virtual Perspective \\ \hline
    \textbf{Definition} & {Determine the relative spatial location of a target object from the egocentric viewpoint of a specified reference object.} \\ \hline
    \multicolumn{2}{|l|}{\cellcolor[HTML]{F5F5F5}\textbf{Example 1}} \\ \hline
    \textbf{Input Image} & 
        \begin{minipage}{0.72\textwidth}
            \centering
            \includegraphics[width=0.48\linewidth]{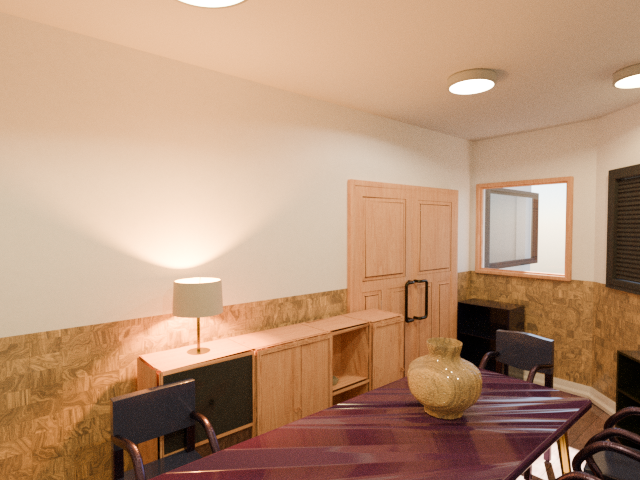} \hfill
            \includegraphics[width=0.48\linewidth]{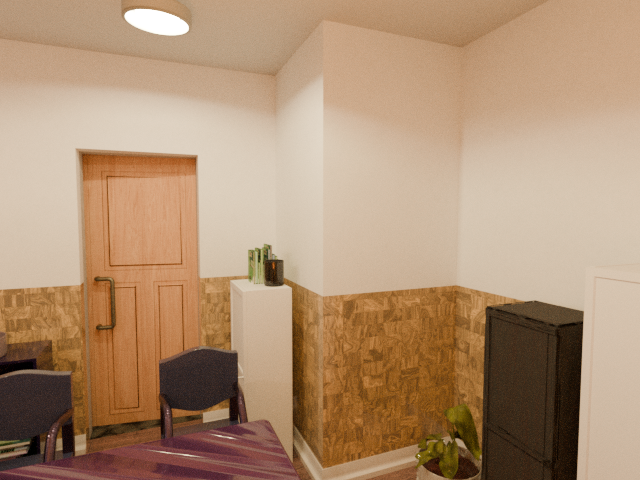}
        \end{minipage} \\ \hline
    \textbf{QA Content}  & 
        \begin{minipage}{0.72\textwidth}
            \textbf{Q:} If the direction faced by a person sitting on the rightmost chair in figure 2 is considered the northeast, where is the glass panel door in figure 1 located in relation to the rightmost chair in figure 2? Options: A: east, B: south, C: northwest, D: southeast \par
            \textbf{A:} A: east
        \end{minipage} \\ \hline
    \end{tabular}
\end{table}
\subsection{Spatial Cognition Process Question Group.}
\begin{figure}[!htbp]
  \centering
  \includegraphics[width=\linewidth]{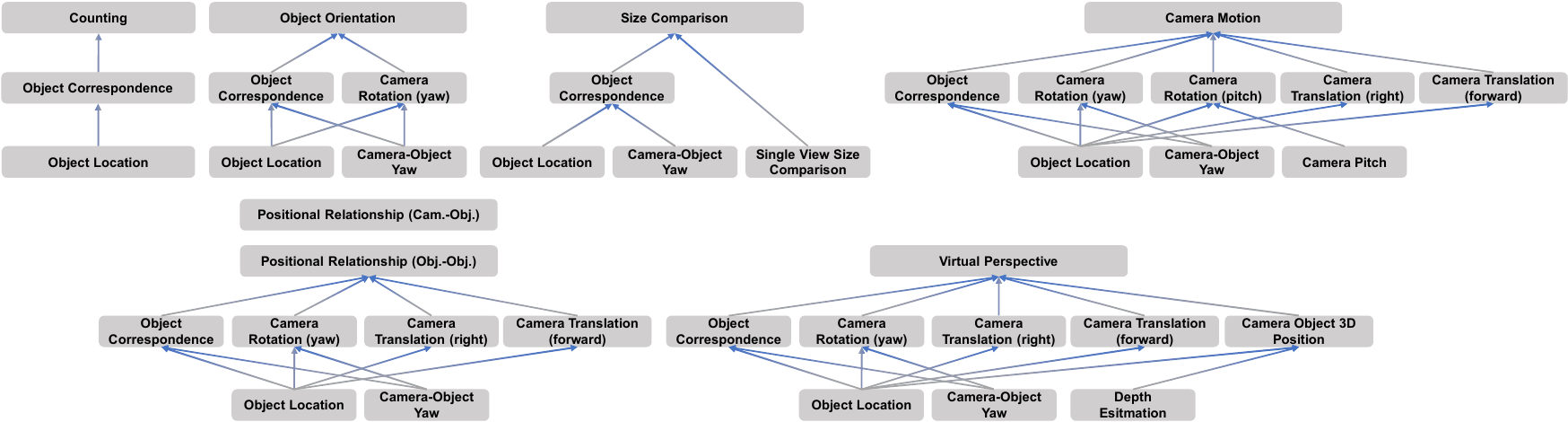}
  \caption{Comprehensive visualization of the Spatial Cognition Process Question Group across all categories in MV-STRIDE.}
  \label{fig:group}
\end{figure}

As described in the main text, we explicitly define the dependencies between capability items across different levels of our dataset, which are formulated as compositional question structures rooted in each task at Level III. \cref{fig:group} comprehensively illustrates all question compositions defined in this work.
\subsection{CoT annotation of Level III}
Following the methodology described in the main text, we leverage the proposed question groups to construct CoT annotations for all task categories in Level III. It should be noted that we exclude the Virtual Perspective task from this process, as the inherent limitations of the ScanNet++ dataset make it difficult to definitively determine the forward orientation of specific reference objects. Representative examples of the constructed CoT training data across diverse categories are summarized in \cref{tab:cot_counting,tab:cot_object_orientation,tab:cot_size_comparison,tab:cot_camera_motion,tab:cot_position_relationship_cam_obj,tab:cot_position_relationship_obj_obj}.
\begin{table}[!htbp]
    \centering
    \caption{Example of the constructed CoT training data for Counting in Level III.}
    \label{tab:cot_counting}
    \small
    \renewcommand{\arraystretch}{1.6} 
    \begin{tabular}{|p{0.92\textwidth}|}
    \hline
    \cellcolor[HTML]{F5F5F5}\textbf{Task Category:} Counting \\ \hline
    \textbf{Input Images:} \\
    \begin{minipage}{0.88\textwidth}
        \begin{center}
            \includegraphics[width=0.48\linewidth]{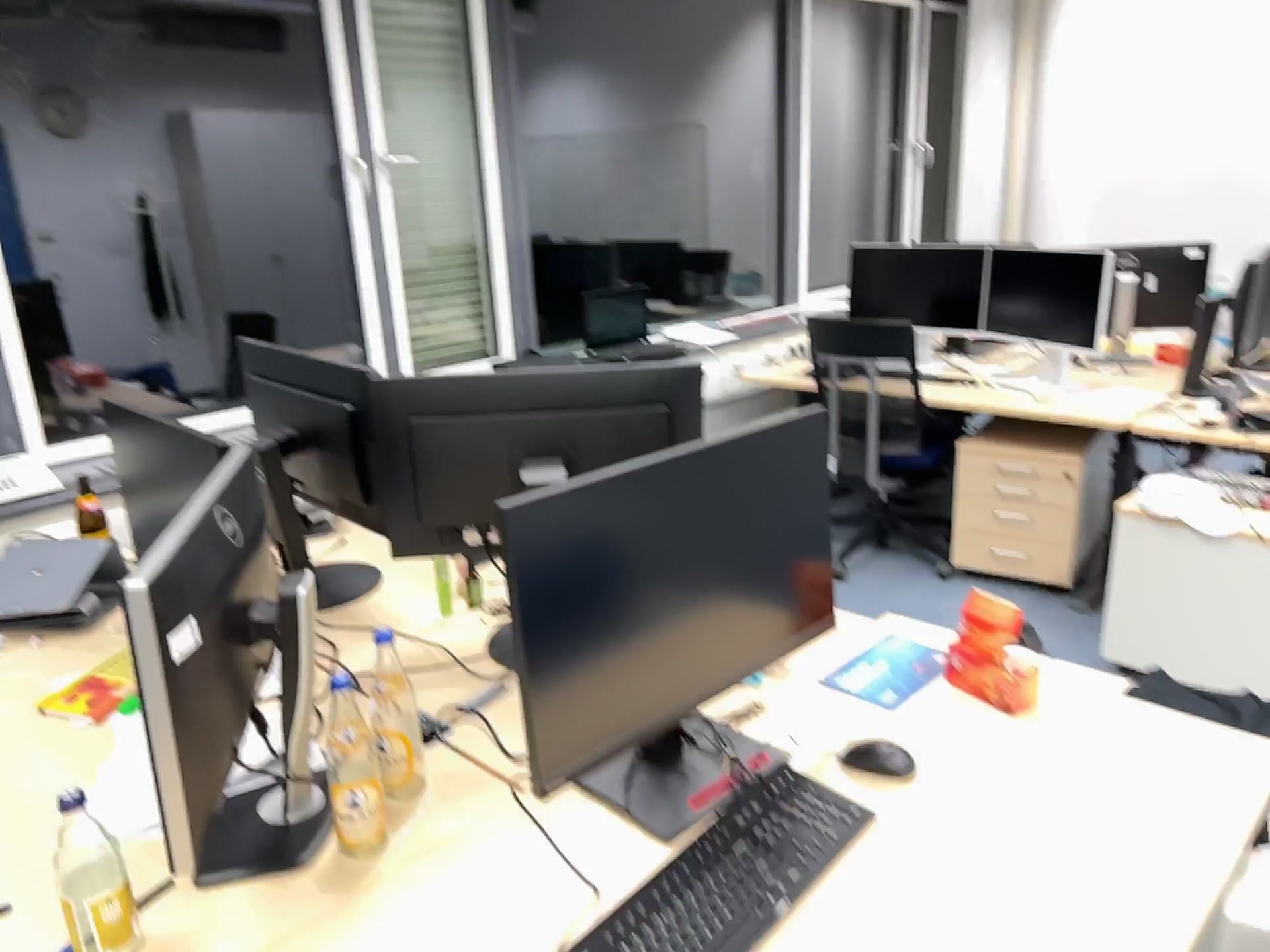} \hfill
            \includegraphics[width=0.48\linewidth]{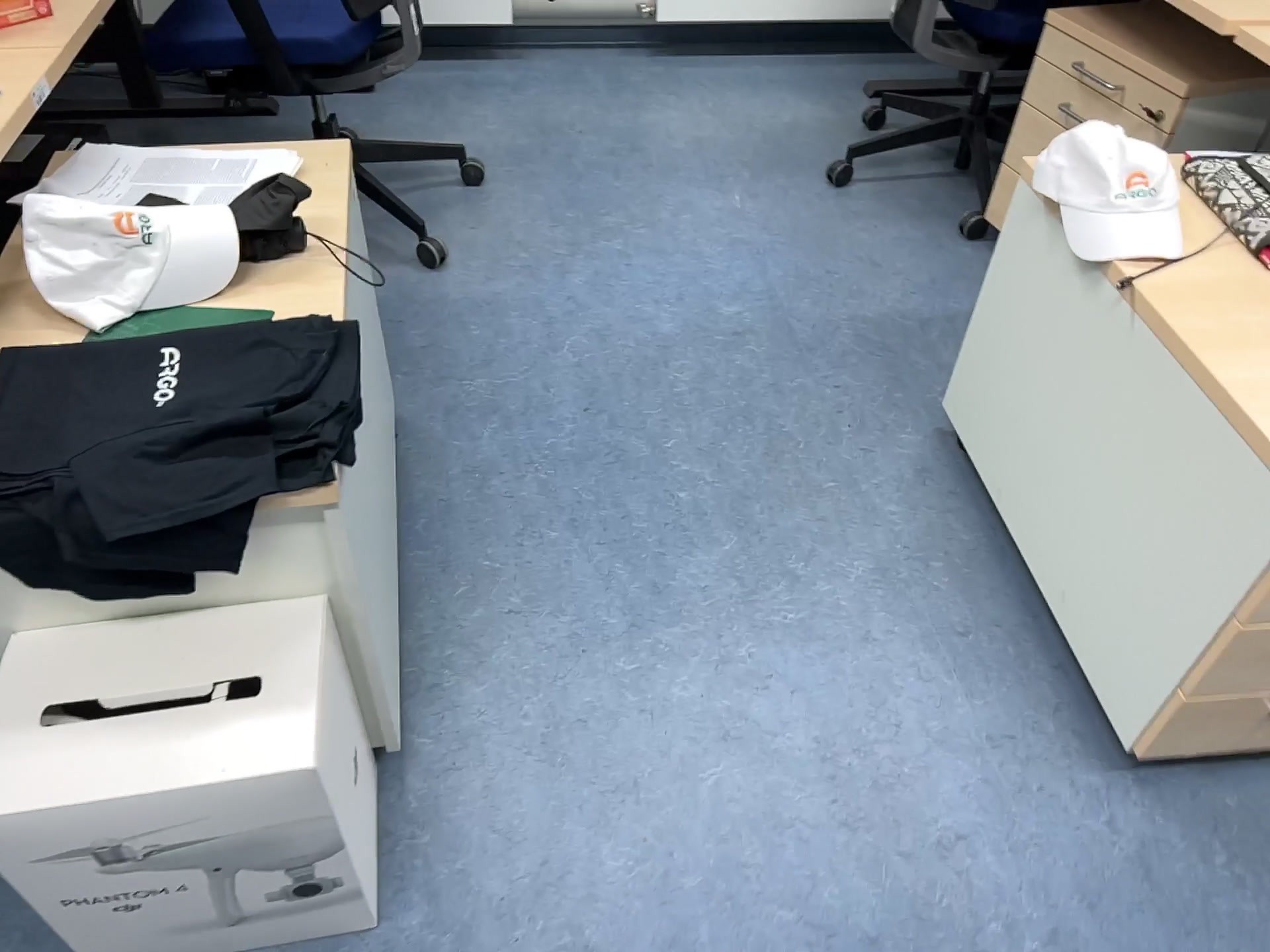}
        \end{center}
    \end{minipage} 
    \\ \hline
    \begin{minipage}{0.88\textwidth}
    {\scriptsize
        \textbf{Question:} How many office chair are in these images? Options: A: 2, B: 3, C: 5, D: 1 \par
        \textbf{Chain-of-Thought:} <think> 
        
Stage 1: Single-View Perception

- In Figure 1, I observe a cluttered office environment. Looking toward the background on the right side, I can identify one office chair tucked under a desk near a set of drawers. Its bounding box is approximately [625, 317, 750, 596].

- In Figure 2, the camera is positioned lower, looking down an aisle between desks. I can clearly see two office chairs. One is on the left side of the aisle ([116, 0, 386, 281]) and another is on the right side ([667, 0, 879, 258]).

Stage 2: Cross-View 3D Modeling

- By comparing the two images, I can determine the spatial relationship between the viewpoints. Figure 1 is a wide shot of the office, while Figure 2 is a close-up view of the floor area between two desk rows.

- I need to establish object correspondence. The office chair on the right in Figure 2 ([667, 0, 879, 258]) matches the chair seen in the background of Figure 1 ([625, 317, 750, 596]) based on its position relative to the light-colored wooden drawers and the desk structure.

- The office chair on the left in Figure 2 ([116, 0, 386, 281]) is located in an area that is off-camera or obscured by the foreground monitors in Figure 1. Therefore, it does not appear in Figure 1.

Stage 3: High-Level Contextual Inference

- To find the total number of unique office chairs, I combine the observations.

- Chair A: Seen in both Figure 1 and Figure 2 (right side).

- Chair B: Seen only in Figure 2 (left side).

- There are no other chairs visible in either image.

- Total unique office chairs = 1 (shared) + 1 (unique to Figure 2) = 2.

</think>

<answer>A: 2</answer>}
    \end{minipage} 
    \\ \hline
    \end{tabular}
\end{table}
\begin{table}[!htbp]
    \centering
    \caption{Example of the constructed CoT training data for Object Orientation in Level III.}
    \label{tab:cot_object_orientation}
    \small
    \renewcommand{\arraystretch}{1.6} 
    \begin{tabular}{|p{0.92\textwidth}|}
    \hline
    \cellcolor[HTML]{F5F5F5}\textbf{Task Category:} Object Orientation \\ \hline
    \textbf{Input Images:} \\
    \begin{minipage}{0.88\textwidth}
        \begin{center}
            \includegraphics[width=0.48\linewidth]{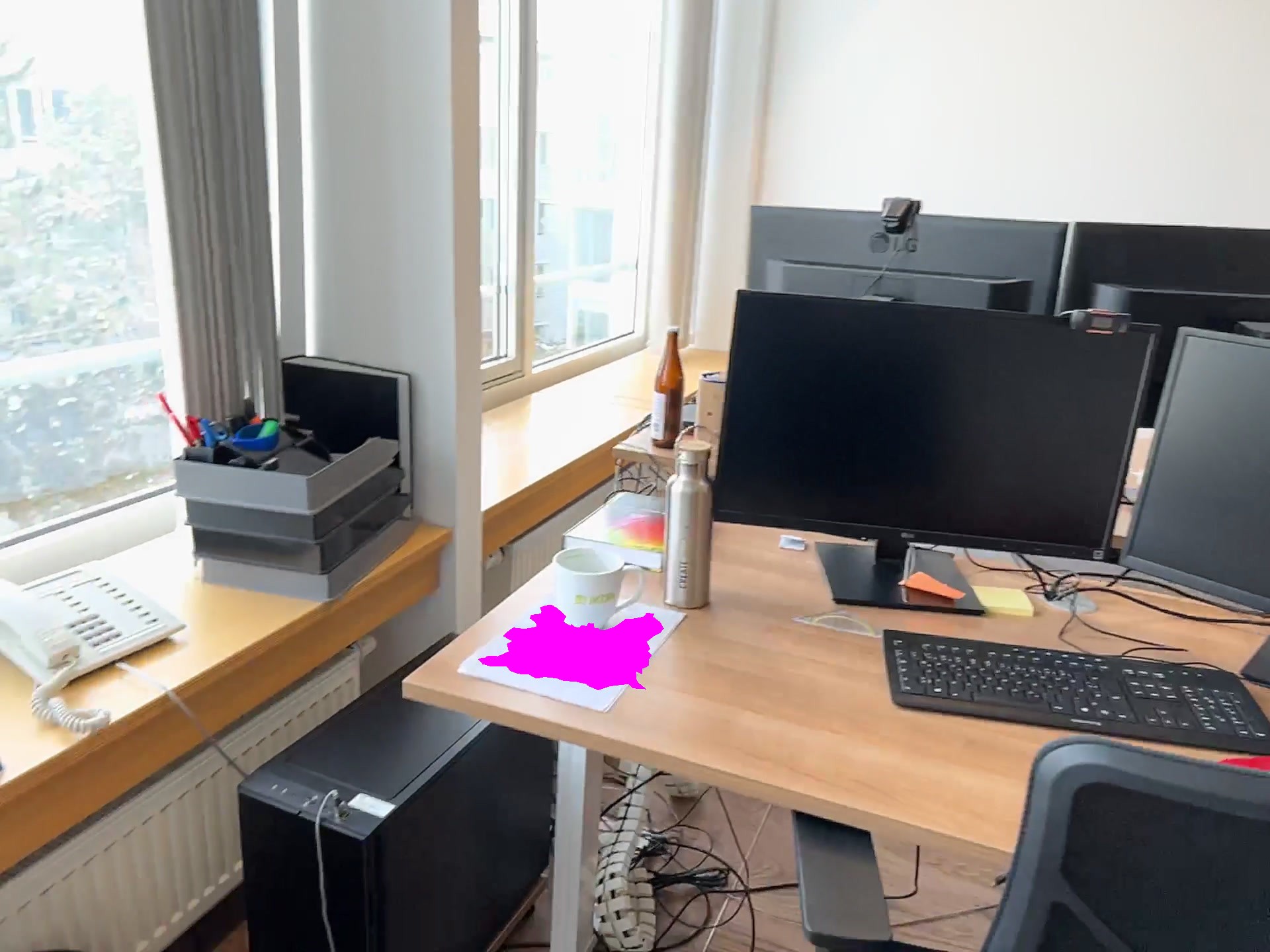} \hfill
            \includegraphics[width=0.48\linewidth]{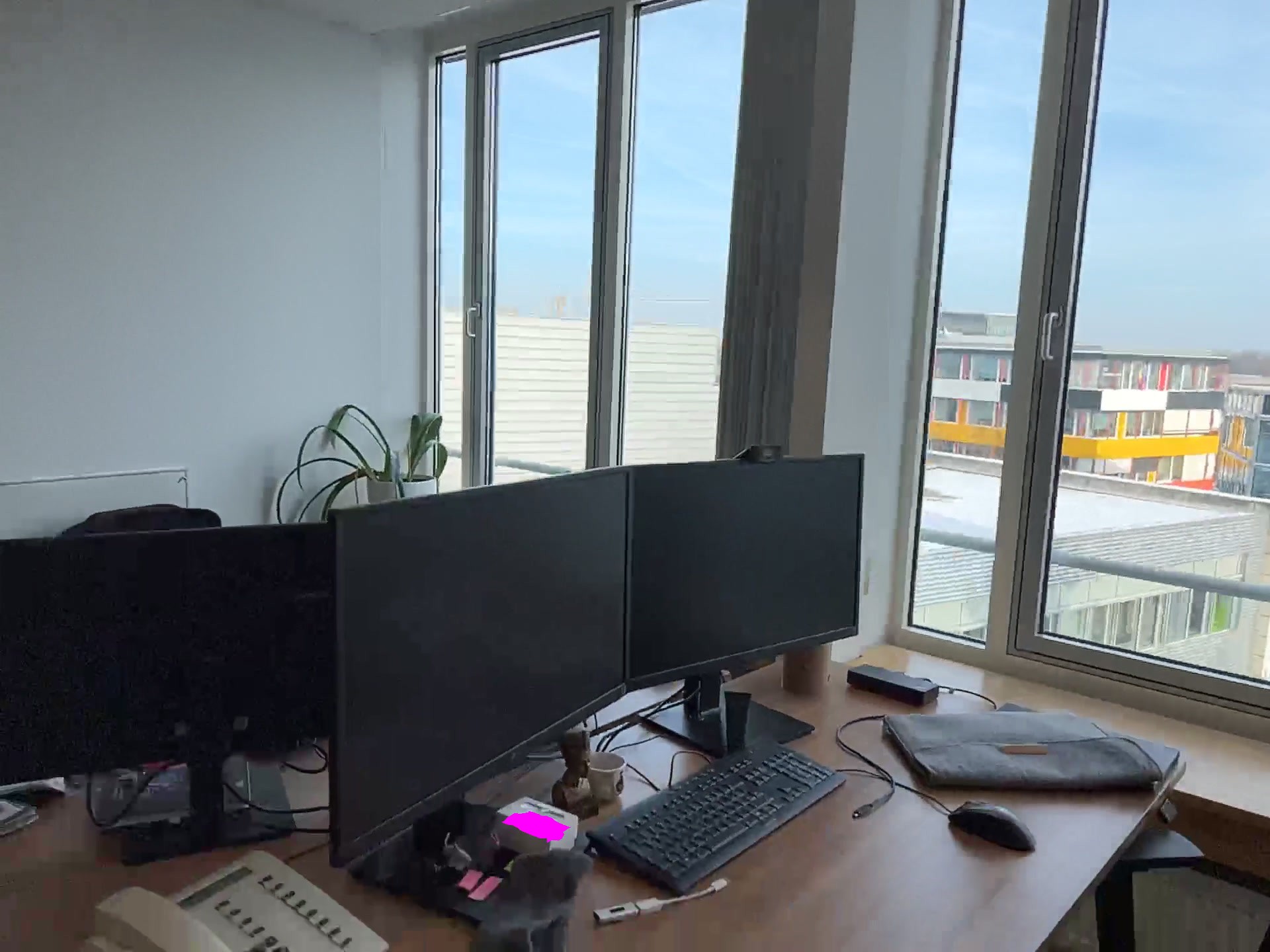}
        \end{center}
    \end{minipage} \\ \hline
    \begin{minipage}{0.88\textwidth}
    {\scriptsize
        \textbf{Question:} Assuming I took photo 1 from the southwest side of the table, from which direction did I take photo 2? Options: A: southeast, B: east, C: north, D: west \par
        \textbf{Chain-of-Thought:} <think>
        
Stage 1: Single-View Perception

- In Figure 1, I identify a wooden office table equipped with multiple monitors, a keyboard, a silver water bottle, and a white mug. The table is positioned in the lower right quadrant of the frame, with bounding box coordinates [335, 512, 999, 999]. Based on the camera's orientation, the center of this table is located at a horizontal angular offset of 24.23 degrees to the right of the camera's center line.

- In Figure 2, I observe the same wooden table. However, its position in the frame has shifted significantly to the left, now located at [0, 704, 549, 999]. In this view, the table is positioned 15.82 degrees to the left of the camera's central line of sight.

Stage 2: Cross-View 3D Modeling

- By comparing the objects (the specific arrangement of monitors, the silver bottle, and the white mug), I confirm that the table in Figure 1 and Figure 2 is the same physical object.
- Analyzing the transition between the two images, the camera underwent a substantial horizontal rotation. Specifically, the camera rotated to the left by approximately 105.7 degrees.

- I also notice a change in the background environment: in Figure 1, the large windows are on the left side of the frame, whereas in Figure 2, the windows are on the right. This indicates the camera has moved to a position roughly opposite its original viewpoint relative to the desk and the window wall.

Stage 3: High-Level Contextual Inference

- I am given that Figure 1 was taken from the southwest side of the table. To capture the table from the southwest, the camera must be facing roughly northeast (approximately 45 degrees on a compass).

- From this starting orientation (facing ~45°), the camera rotates 105.7 degrees to the left (counter-clockwise).

- Calculating the new orientation: 45° - 105.7° = -60.7°. On a standard 360-degree compass where North is 0°, -60.7° corresponds to 299.3° (which is roughly West-Northwest).

- If the camera is now facing West-Northwest to view the table, the camera itself must be positioned to the East-Southeast of the table.

- Among the provided options (A: southeast, B: east, C: north, D: west), "east" is the most accurate description of the camera's new position relative to the table after a ~106-degree leftward rotation from the southwest.

</think>

<answer>B: east</answer>}
    \end{minipage} \\ \hline
    \end{tabular}
\end{table}
\begin{table}[!htbp]
    \centering
    \caption{Example of the constructed CoT training data for Size Comparison in Level III.}
    \label{tab:cot_size_comparison}
    \small
    \renewcommand{\arraystretch}{1.6} 
    \begin{tabular}{|p{0.92\textwidth}|}
    \hline
    \cellcolor[HTML]{F5F5F5}\textbf{Task Category:} Size Comparison \\ \hline
    \textbf{Input Images:} \\
    \begin{minipage}{0.88\textwidth}
        \begin{center}
            \includegraphics[width=0.48\linewidth]{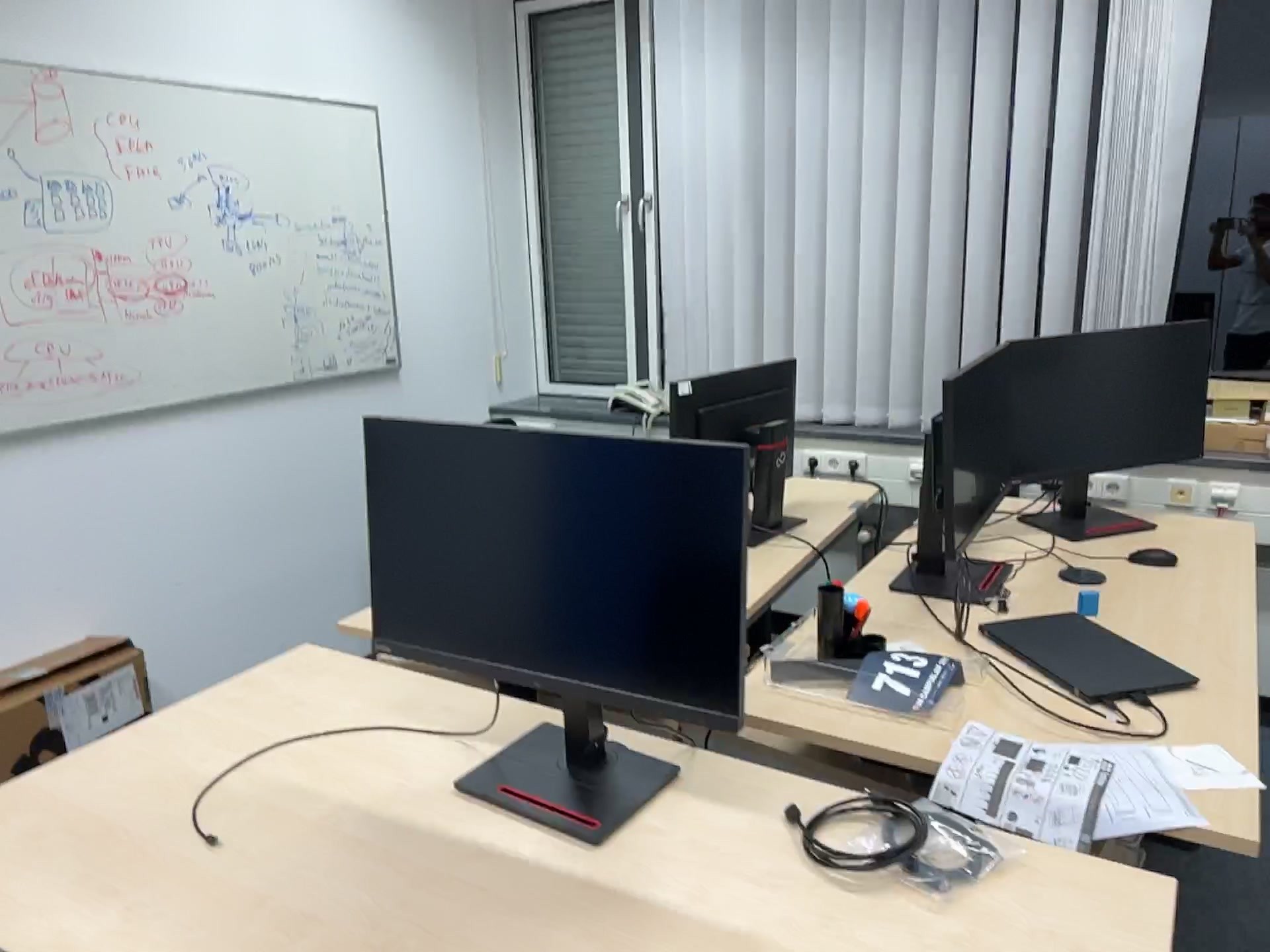} \hfill
            \includegraphics[width=0.48\linewidth]{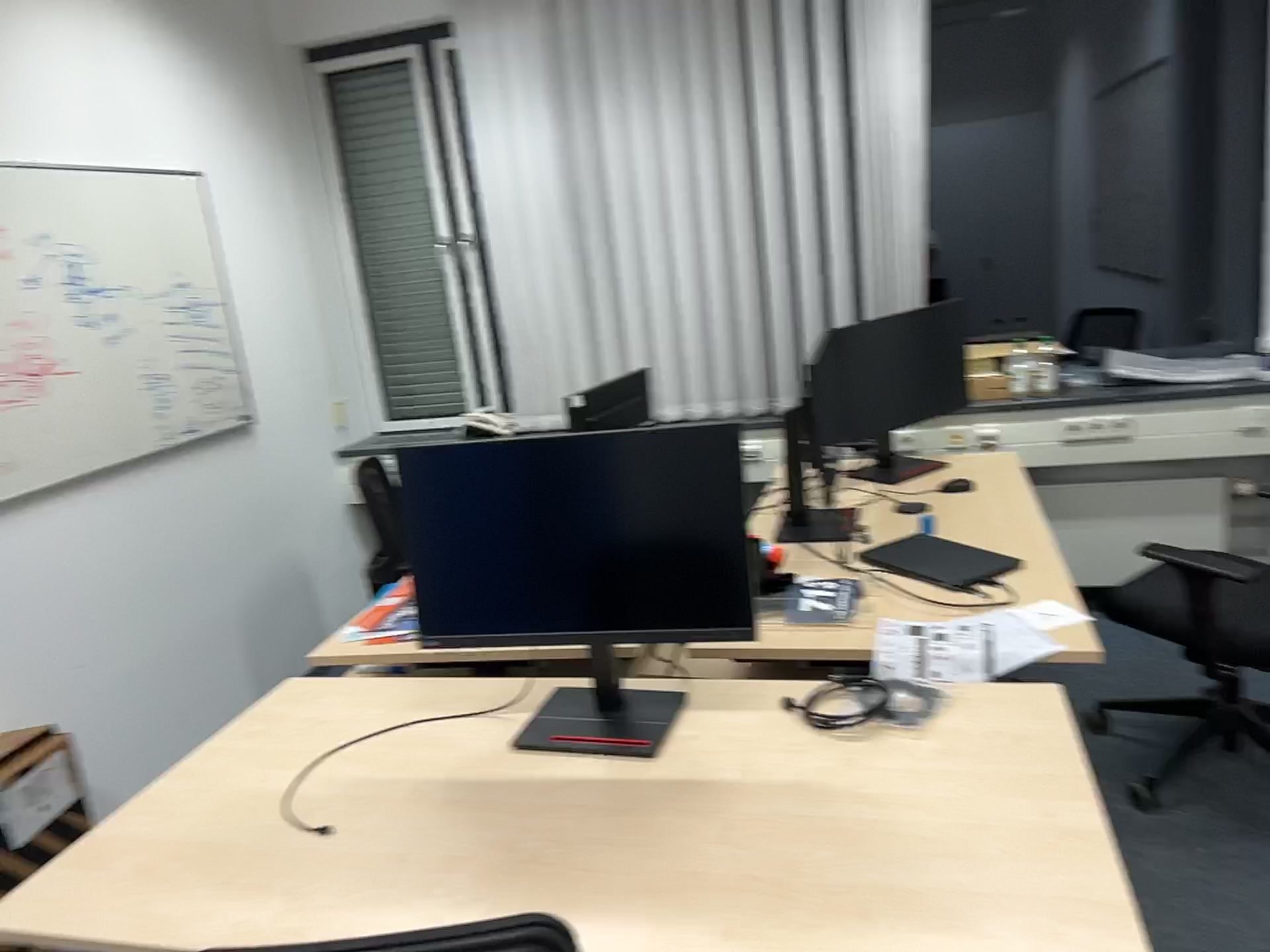}
        \end{center}
    \end{minipage} \\ \hline
    \begin{minipage}{0.88\textwidth}
    {\scriptsize
        \textbf{Question:} The two photos show the same scene. Comparing 4 times the width of the penholder in Photo 1 with width of the rightmost office chair in Photo 2, which one is wider? Options: A: The same length, B: Sometimes the former, sometimes the latter, C: The width of the rightmost office chair in Figure 2, D: The 4 times the width of the penholder in Figure 1 \par
        \textbf{Chain-of-Thought:} <think>
Stage 1 – Single-view perception

- In Figure 1, I identify a penholder located on the desk at [599, 612, 696, 700]. Based on spatial grounding and depth estimation, it is approximately 1.89 meters from the camera with a physical horizontal width of about 0.23 meters. I also note a box (the middle one) at [940, 425, 999, 481] at a depth of 3.37 meters, with a width of 0.26 meters.

- In Figure 2, I observe an office chair on the right side of the frame at [851, 508, 999, 848]. It is positioned at a depth of roughly 2.97 meters and has a significant physical width of approximately 1.03 meters. Additionally, I locate a box at [760, 378, 829, 422] at a depth of 3.84 meters, which also measures about 0.26 meters in width.

Stage 2 – Cross-view 3D modeling

- By comparing the objects across both views, I can confirm that the "middle box" in Figure 1 and the "rightmost box" in Figure 2 are the same physical object, serving as a spatial anchor. The camera in Figure 2 is positioned further back and angled more toward the right compared to Figure 1, which brings the office chair into full view.

- This 3D correspondence allows me to treat the measurements derived from both images as part of a single, consistent environment.

Stage 3 – High-Level Contextual Reasoning

- To answer the final question, I compare the calculated width of four penholders to the width of the office chair.

- Taking the penholder's width of 0.23 meters from Figure 1, four times this width equals $4 \times 0.23 = 0.92$ meters.

- Comparing this to the office chair's width of 1.03 meters from Figure 2, it is clear that 1.03 meters is greater than 0.92 meters.

- Therefore, the width of the rightmost office chair in Figure 2 is greater than four times the width of the penholder in Figure 1.

</think>

<answer>C: The width of the rightmost office chair in Figure 2</answer>}
    \end{minipage} \\ \hline
    \end{tabular}
\end{table}
\begin{table}[!htbp]
    \centering
    \caption{Example of the constructed CoT training data for Camera Motion in Level III.}
    \label{tab:cot_camera_motion}
    \small
    \renewcommand{\arraystretch}{1.6} 
    \begin{tabular}{|p{0.92\textwidth}|}
    \hline
    \cellcolor[HTML]{F5F5F5}\textbf{Task Category:} Camera Motion \\ \hline
    \textbf{Input Images:} \\
    \begin{minipage}{0.88\textwidth}
        \begin{center}
        \includegraphics[width=0.48\linewidth]{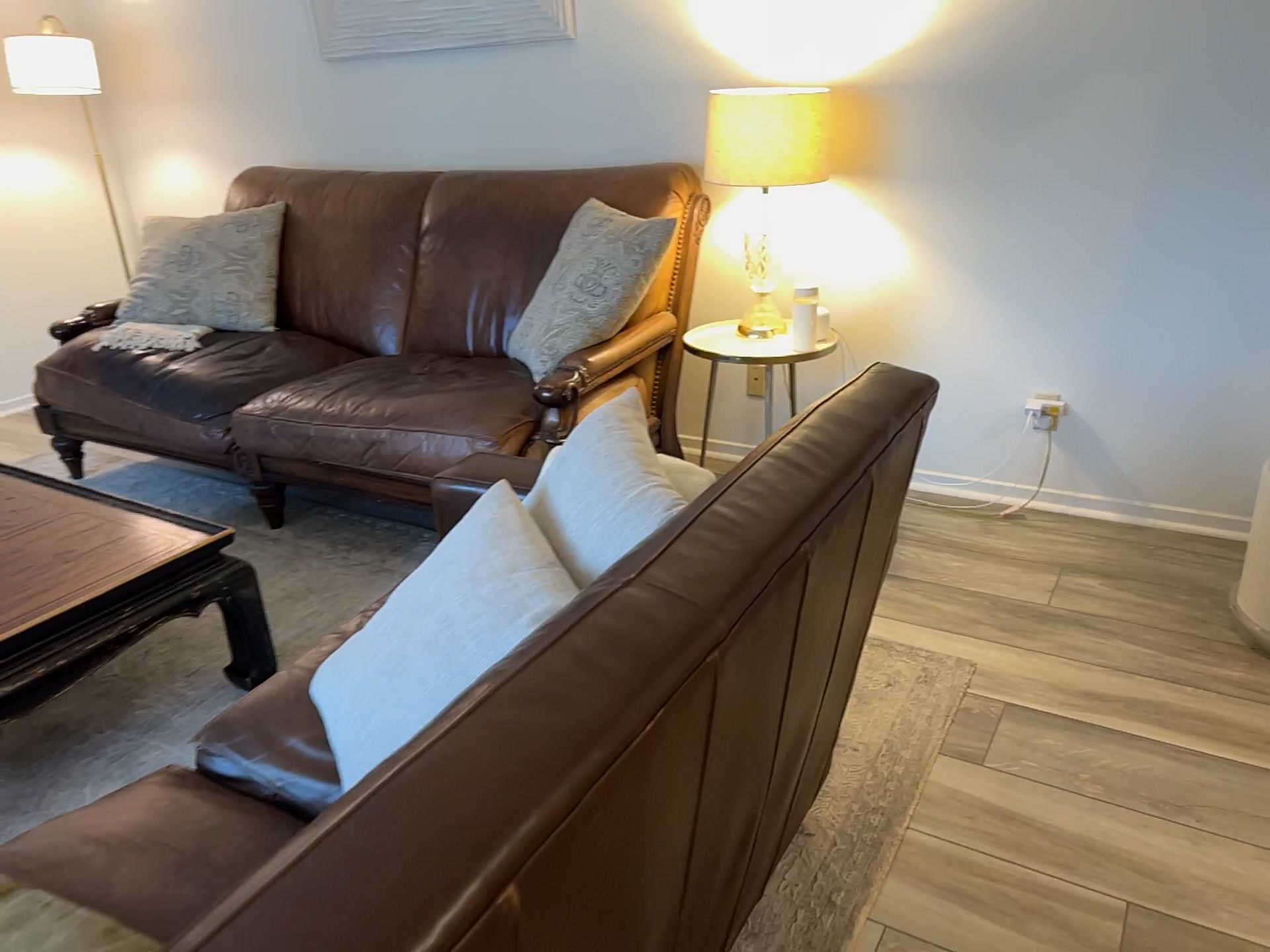} \hfill
        \includegraphics[width=0.48\linewidth]{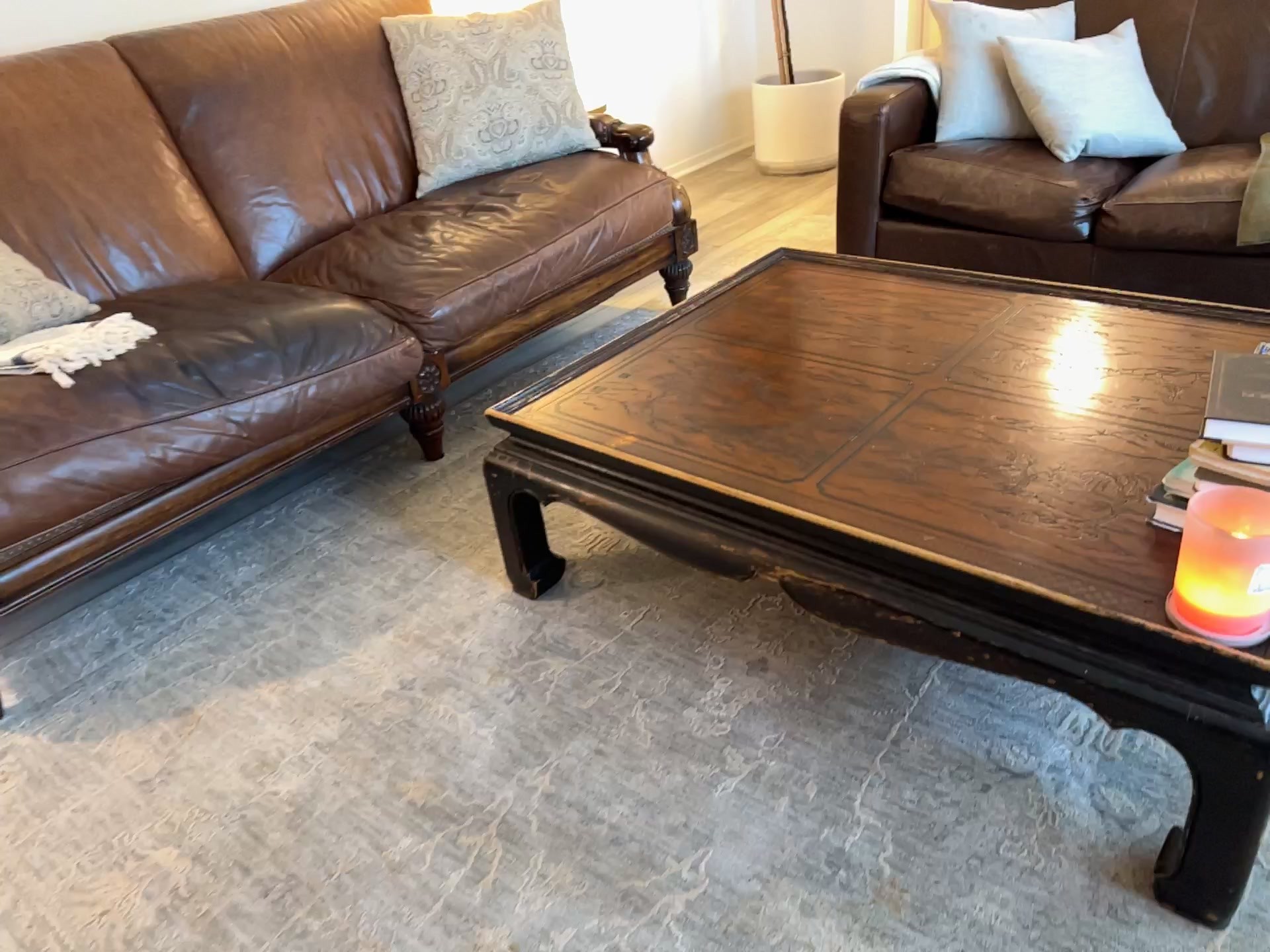}
        \end{center}
    \end{minipage} \\ \hline
    \begin{minipage}{0.88\textwidth}
    {\scriptsize
        \textbf{Question:} I went from taking the first picture to taking the second picture, how did I most likely move? Options: A: Not moving, B: backward to the right, C: forward to the left, D: backward to the left \par
        \textbf{Chain-of-Thought:} <think>
        
To determine the camera's motion between the two images, I will follow a multi-stage reasoning process: perception of individual views, cross-view modeling of the 3D scene, and high-level inference of the camera's trajectory.

\textbf{Stage 1: Single-View Perception}

In **Figure 1**, I observe a living room setting. The primary anchor is a large brown leather couch located in the background (topmost couch at [29, 171, 553, 556]). To its left, a dark wood coffee table is partially visible. The camera is positioned at a moderate height, tilted downwards at an angle of 22.45 degrees. The background couch is positioned about 16.11 degrees to the left of the camera's center line.

In **Figure 2**, the perspective has shifted significantly. The same brown leather couch is now the leftmost object in the frame ([0, 0, 549, 754]). The coffee table is now much more prominent and appears closer to the lens. The camera's downward tilt has increased to 30.40 degrees, and the couch is now 19.02 degrees to the left of the center line.

\textbf{Stage 2: Cross-View 3D Modeling}

By identifying the same physical couch in both images, I can establish a spatial correspondence. 

1.  **Rotation:** Between Figure 1 and Figure 2, the camera underwent a significant rightward rotation of approximately 67.78 degrees. This explains why the couch, which was in the background of Figure 1, now appears on the far left side of Figure 2. Additionally, the camera tilted further down by about 20.54 degrees (relative to the scene's horizon).

2.  **Translation:** 
    *   **Forward/Backward:** In Figure 2, the coffee table and the couch appear larger and the perspective on the floor rug is more "top-down," indicating the camera has moved closer to the center of the room. Quantitatively, the camera moved forward by approximately 1.7752 meters relative to its initial position in Figure 1.
    *   **Left/Right:** Comparing the alignment of the couch and the wall, the camera has shifted its lateral position. It moved to the left by approximately 3.1729 meters.

\textbf{Stage 3: High-Level Contextual Inference}

Combining the translation components derived from the 3D modeling:

- The camera moved **forward** (closer to the furniture).

- The camera shifted to the **left** (moving across the room).

This combined movement—forward and to the left—describes a diagonal path through the living space. This trajectory, combined with the rightward rotation, allows the camera to maintain a view of the seating area while changing its vantage point from a distant, slightly right-side view to a closer, side-on view.

Therefore, the most likely motion when transitioning from the first picture to the second is moving forward to the left.

</think>

<answer>C: forward to the left</answer>}
    \end{minipage} \\ \hline
    \end{tabular}
\end{table}
\begin{table}[!htbp]
    \centering
    \caption{Example of the constructed CoT training data for Position Relationship (Cam.-Obj.) in Level III.}
    \label{tab:cot_position_relationship_cam_obj}
    \small
    \renewcommand{\arraystretch}{1.6} 
    \begin{tabular}{|p{0.92\textwidth}|}
    \hline
    \cellcolor[HTML]{F5F5F5}\textbf{Task Category:} Position Relationship (Cam.-Obj.) \\ \hline
    \textbf{Input Images:} \\
    \begin{minipage}{0.88\textwidth}
        \begin{center}
        \includegraphics[width=0.48\linewidth]{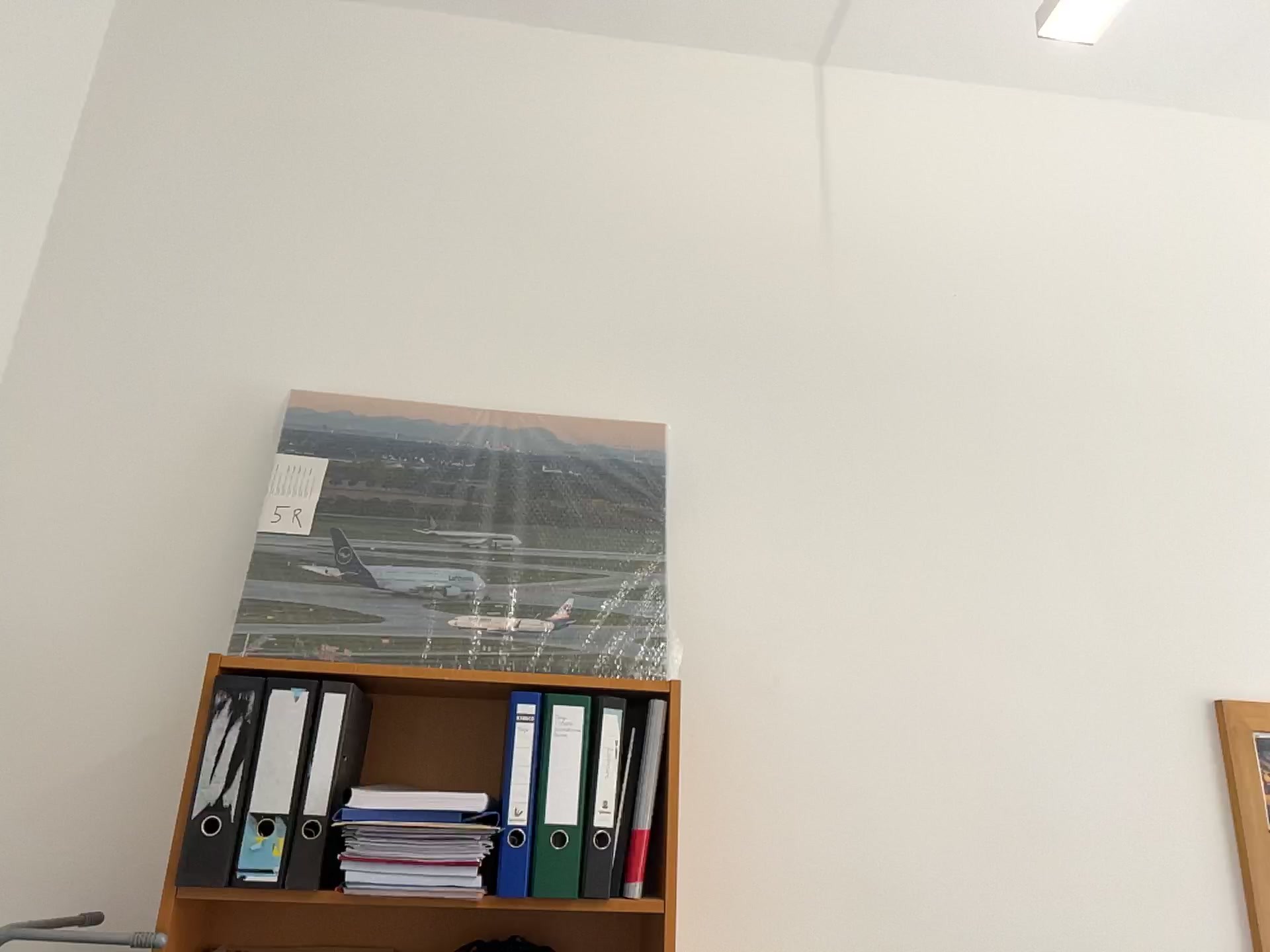} \hfill
        \includegraphics[width=0.48\linewidth]{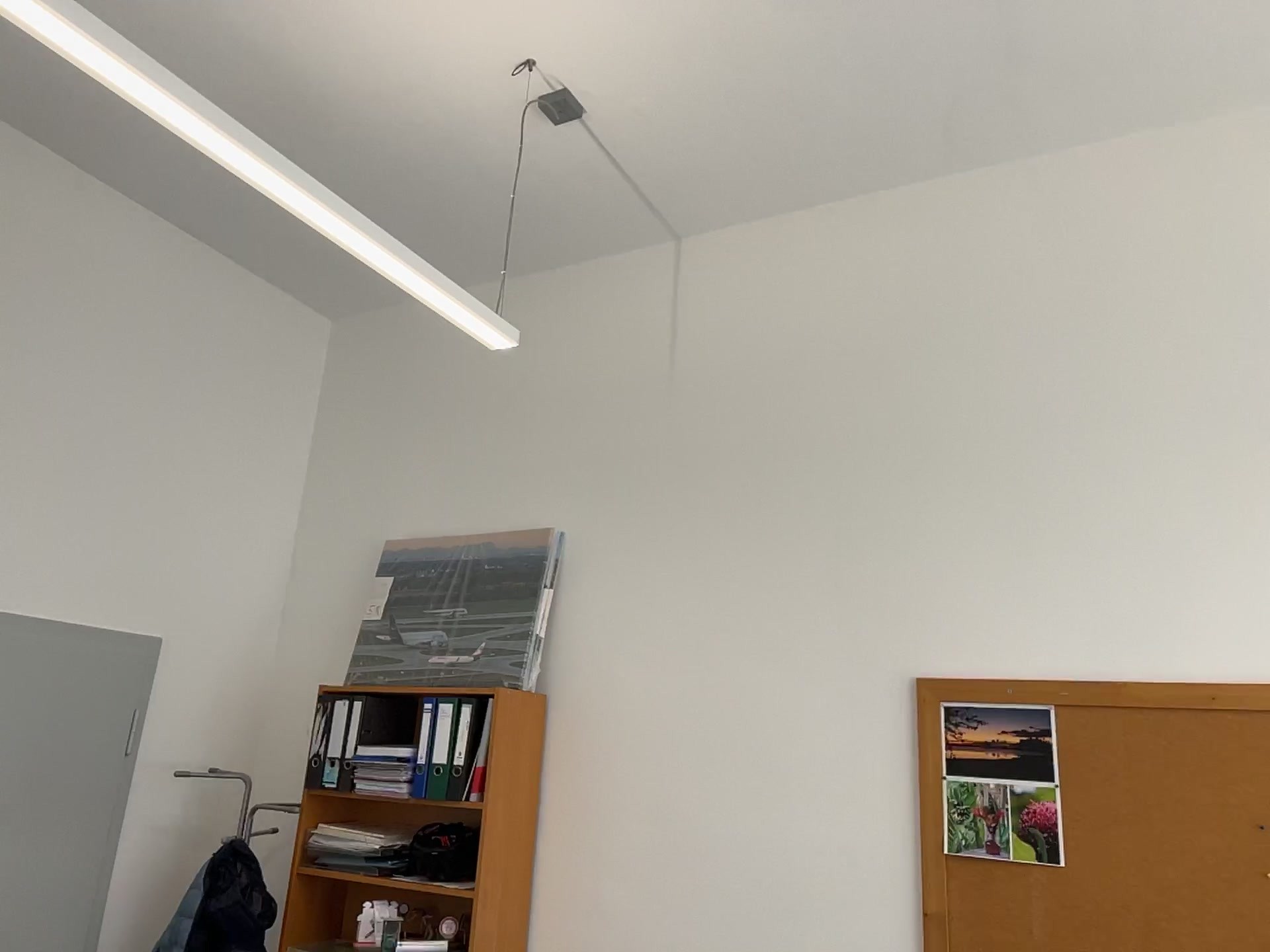}
        \end{center}
    \end{minipage} \\ \hline
    \begin{minipage}{0.88\textwidth}
    {\scriptsize
        \textbf{Question:} The pictures are taken continuously from a first-person perspective, with two pictures overlapping at the frame. At the moment of the first picture, in which direction is the door frame in the last picture relative to you? Options: A: backward right, B: backward left, C: forward right, D: forward left \par
        \textbf{Chain-of-Thought:} <think>
        
To determine the positional relationship of the door frame relative to the camera's initial position in Figure 1, I will follow a three-stage reasoning process: perception, modeling, and inference.

\textbf{Stage 1: Single-View Perception}

First, I analyze the individual images to identify key landmarks and their local spatial properties.

*   **Figure 1:** The camera is positioned close to a bookshelf containing several binders (black, blue, green, grey) and a large picture of an airport. The camera is facing a white wall directly.

*   **Figure 2:** The same bookshelf is visible, but from a wider perspective. I identify a **door frame** on the far left side of the image. Based on visual analysis, the door frame is located at bounding box coordinates **[0, 636, 125, 999]**.

*   **Local Angular Offset:** In Figure 2, the door frame is not centered; it is positioned at a horizontal angular offset of approximately **35.45 degrees to the left** of the camera's central line of sight.

\textbf{Stage 2: Cross-View 3D Modeling}

Next, I determine how the camera moved between the two frames to construct a global 3D spatial map.

*   **Camera Rotation:** Comparing the orientation of the bookshelf and the wall, the camera in Figure 2 has rotated **left by approximately 33.59 degrees** relative to its orientation in Figure 1.

*   **Camera Translation:** 
    *   By observing the change in perspective and the size of the bookshelf, I determine that the camera shifted **right by approximately 2.2514 meters** (relative to the Figure 1 coordinate system).
    *   Additionally, the bookshelf appears smaller and more of the room is visible, indicating the camera moved **backward by approximately 0.6009 meters**.

\textbf{Stage 3: High-Level Contextual Inference}

Now, I integrate the camera's movement with the object's detected position to find the door frame's location relative to the first viewpoint (Figure 1).

1.  **Angular Summation:** In Figure 2, the camera is already rotated 33.59° to the left. From that rotated heading, the door frame is another 35.45° to the left. This suggests that relative to the forward vector of Figure 1, the door frame is at a very wide angle to the left (roughly 69°).

2.  **Translational Impact:** The camera moved significantly to the right (2.25m) to reach the position in Figure 2. Despite moving right, the door frame still appears on the left side of the view. This implies the door frame is physically located well to the left of the initial camera position.

3.  **Depth Consideration:** The camera moved backward to Figure 2. Since the door frame is visible in Figure 2 along the same wall structure as the bookshelf (which is in front of the camera in Figure 1), the door frame is located in the forward space relative to the Figure 1 position.

Combining these observations: the door frame is located to the left of the initial camera's center line and in front of its initial position. Therefore, relative to the camera in Figure 1, the door frame is in the **forward left** direction.
</think>

<answer>D: forward left</answer>}
    \end{minipage} \\ \hline
    \end{tabular}
\end{table}
\begin{table}[!htbp]
    \centering
    \caption{Example of the constructed CoT training data for Position Relationship (Obj.-Obj.) in Level III.}
    \label{tab:cot_position_relationship_obj_obj}
    \small
    \renewcommand{\arraystretch}{1.6} 
    \begin{tabular}{|p{0.92\textwidth}|}
    \hline
    \cellcolor[HTML]{F5F5F5}\textbf{Task Category:} Position Relationship (Obj.-Obj.) \\ \hline
    \textbf{Input Images:} \\
    \begin{minipage}{0.88\textwidth}
        \begin{center}
        \includegraphics[width=0.48\linewidth]{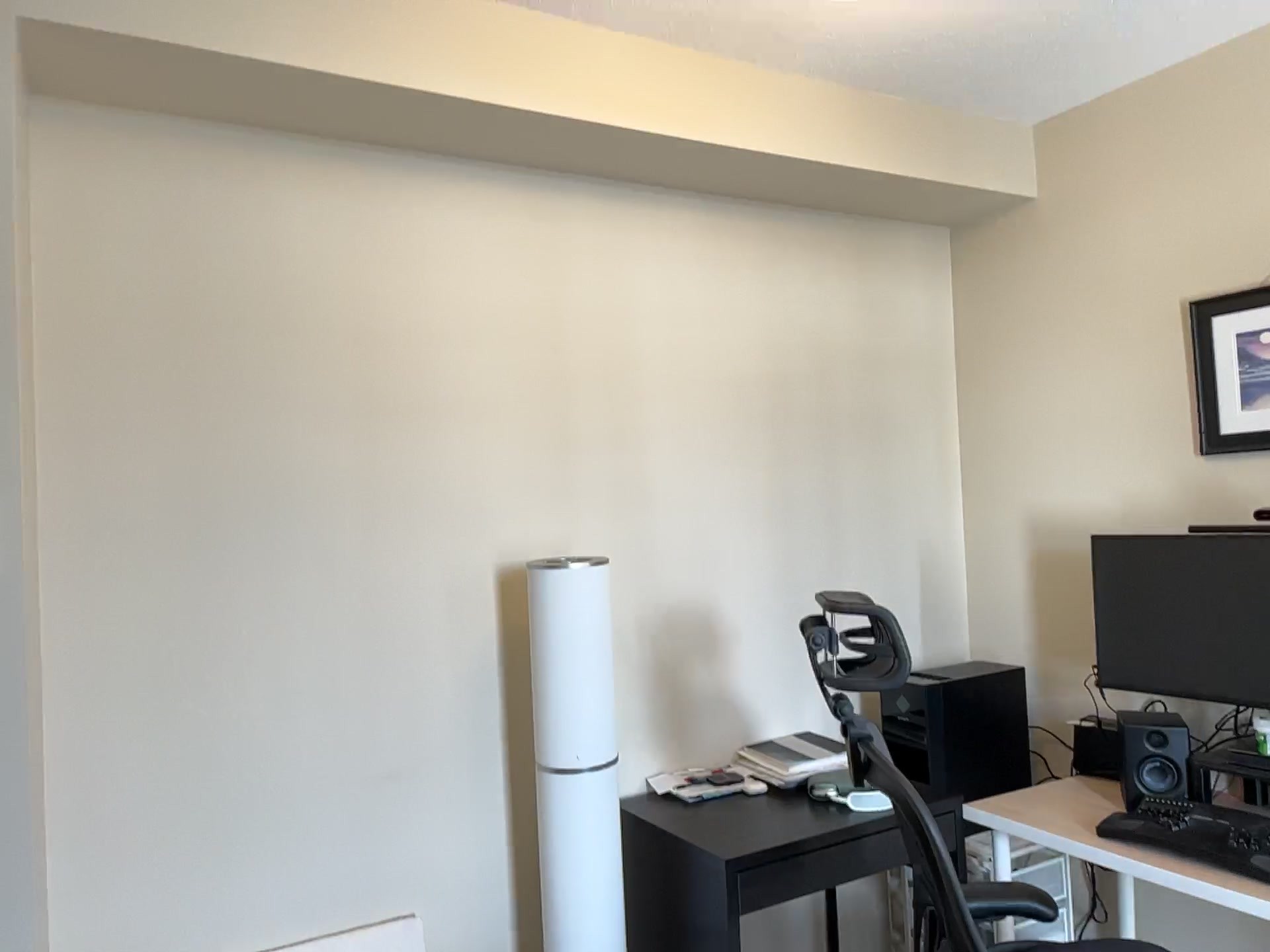} \hfill
        \includegraphics[width=0.48\linewidth]{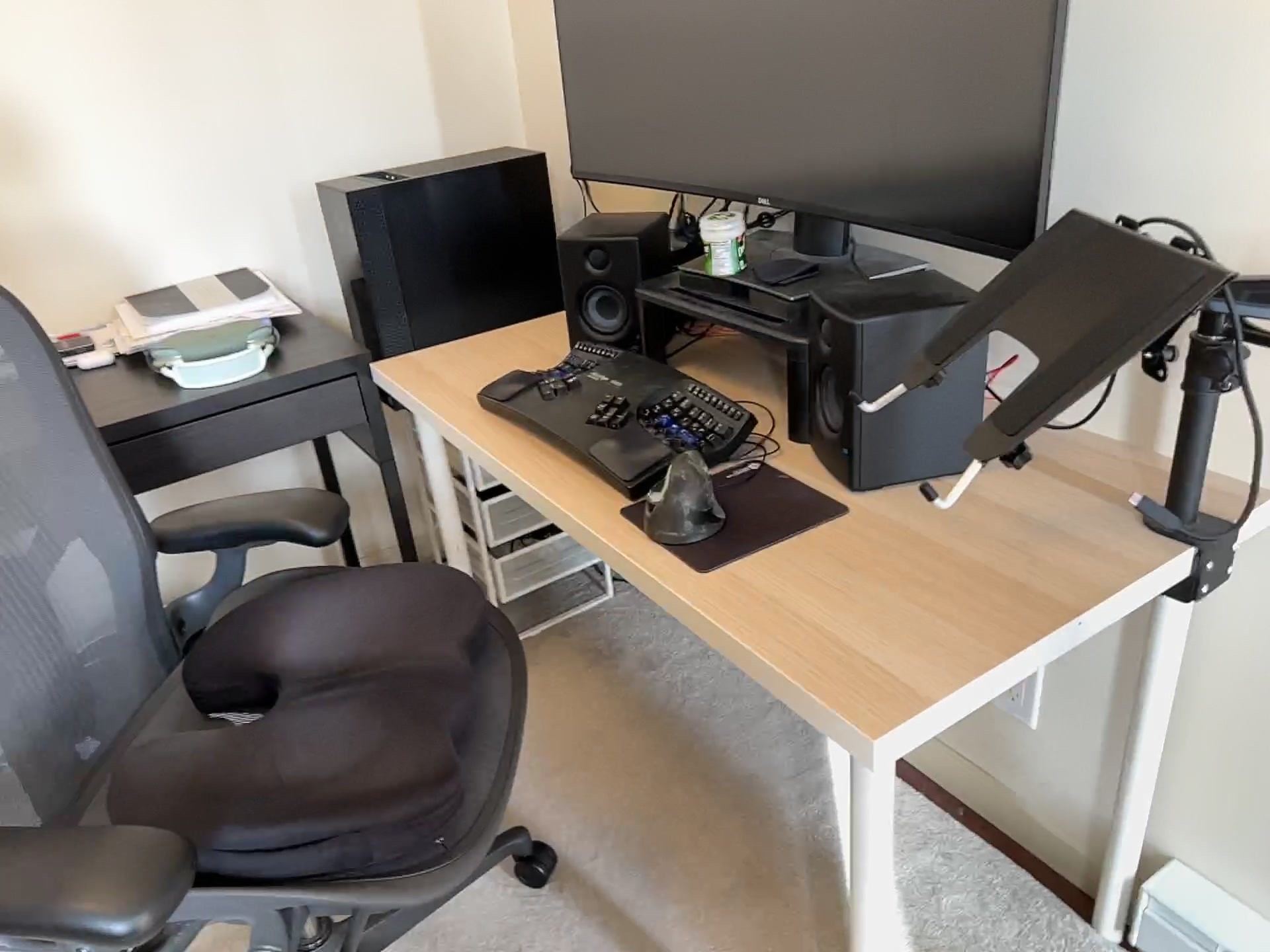}
        \end{center}
    \end{minipage} \\ \hline
    \begin{minipage}{0.88\textwidth}
    {\scriptsize
        \textbf{Question:} In which direction is the picture in figure 1 relative to the leftmost table in figure 2 (given that the rightmost speaker in figure 2 is to the back of the leftmost table in figure 2)?
Options: A: left, B: back, C: right, D: front \par
        \textbf{Chain-of-Thought:} <think>
        
Stage 1: Single-View Perception

In Figure 2, I identify the leftmost table, which is a dark-colored desk located at [0, 317, 329, 767]. This desk is positioned significantly to the left of the camera's center line, with a horizontal angular offset (yaw) of 24.58 degrees. In the same image, I observe a speaker on the right side of a light-colored desk, located at [624, 283, 781, 566], with a yaw of 16.25 degrees to the right.

In Figure 1, I locate a framed picture on the right-hand wall at [934, 298, 999, 476]. This picture is positioned at a horizontal angular offset of 35.04 degrees to the right of the camera's center line. I also see the same dark-colored desk (the leftmost table) now located at [510, 803, 759, 999], which is 8.85 degrees to the right of the camera's center line in this view.

Stage 2: Cross-View 3D Modeling

By comparing the two images, I can establish object correspondences and camera movement. The dark desk identified as the "leftmost table" in Figure 2 is the same physical object as the desk in the lower-right of Figure 1. 

Between Figure 1 and Figure 2, the camera underwent a complex transformation: it rotated right by approximately 2.87 degrees and shifted significantly. Specifically, the camera moved forward by about 0.7887 meters and to the right by 1.1843 meters. 

This movement explains why the dark desk, which was on the right side of the frame in Figure 1, appears on the far left in Figure 2. The camera has essentially moved past the dark desk and turned slightly toward the main workstation (the light-colored desk).

Stage 3: High-Level Contextual Inference

The question asks for the direction of the picture (from Figure 1) relative to the leftmost table (from Figure 2), using a specific orientation reference: the rightmost speaker in Figure 2 is considered to be to the "back" of the leftmost table.

1. From Figure 2, we see the dark desk (leftmost table) in the foreground and the light desk (containing the speaker) further away. If the speaker is to the "back" of the dark desk, then the "back" direction in this scene points deeper into the room, toward the wall where the workstation is set up.

2. In Figure 1, we can see the spatial relationship between the light desk and the picture. The picture is mounted on the wall directly behind and to the right of the monitor and speakers.

3. Since the speaker is "back" relative to the dark desk, and the picture is located on the wall that the speaker's desk is placed against, the picture is also situated in that same "back" direction relative to the dark desk.

4. Therefore, relative to the leftmost table, the picture is located in the back.

</think>

<answer>B: back</answer>}
    \end{minipage} \\ \hline
    \end{tabular}
\end{table}

\section{Hierarchical CoT and RL Data Construction Strategy}
\label{sec:cot_rl}
\subsection{CoT Generation Method} 
Specifically, we employ 20\% of the scenes from ScanNet++ for CoT annotation. We construct question groups where each group culminates in a final question corresponding to one Level III task. These question groups are then converted into a multi-turn dialogue format and fed into Gemini-3-Flash\cite{google_gemini_3} along with the corresponding images. We use the following system prompt to guide the model to output complete reasoning chains, which serve as our CoT annotations.
\newpage
\begin{PromptBox}{System Prompt for CoT Generation}
\tiny
\textbf{Role:}
You are an expert in 3D Scene Understanding and Multimodal Spatial Reasoning. Unlike standard 2D analysis, you possess a human-like 3D spatial cognitive system. You excel at reconstructing complex 3D environments from multi-view images by following a progressive pipeline: "Perception $\rightarrow$ Modeling $\rightarrow$ Reasoning." Your goal is to treat multi-view inputs as a unified 3D world, ensuring spatial consistency across different camera perspectives.

\textbf{Task:}
I will provide you with a set of Multi-view Images and a series of hierarchical Question-Answer (QA) pairs labeled Level I, II, and III. The answers are correct.

Level I(single-view perception): Foundational grounding, depth and attribute perception etc.

Level II(cross-view modeling): corresponding of objects between views, camera transformation etc.

Level III(multi-view context reasoning): Complex, multi-hop high-level spatial reasoning problems.

Your objective is to generate a long, coherent Chain of Thought (CoT) specifically for the final Level III question according to the information of input images and the logic and conclusions of level I and II QAs. 

\textbf{Reasoning Guidelines:}
When generating the reasoning process, you must follow the sequence below to construct a chain of thought that aligns with the human spatial cognition process. The reasoning is divided into three major stages, and each stage can be further broken down into specific logical steps as needed.

Stage 1: Single-View Perception: Identify the 2D and 3D spatial information of cameras and objects in images as the basis for reasoning.

Stage 2: Cross-View 3D Modeling: Integrate information from different perspectives to determine camera transformations and object correspondences. Construct a 3D cognitive map to memory the relative positions and orientations of all elements in the global scene.

Stage 3: High-Level Contextual Inference: Integrate global information to perform high-order spatial reasoning in a 3D consistent manner, eg virtual viewpoint imagination.

\textbf{Strict Requirements:}

1. Simulated Discovery \& anchor: Please assume that you have only received the image and the Level III question, and try to generate a multi-stage reasoning process and answer for the question. You can directly follow the thought progression and answers from Level I and II, and pretend that you have discovered or derived these conclusions on your own, rather than directly quoting them. You need to naturally transform the Q\&A content of level x into the reasoning process of stage x, and always ensure that the reasoning results and the question answers remain consistent, forming a logical and coherent cycle.

2. Image-Text Verification: For most of the time, you can trust the authenticity of all questions and answers, especially the level III answer is definitely correct. But images are the "source of truth." If you find there are inconsistencies, omissions, or errors between the QA content and the image content, you must prioritize correcting your reasoning path based on the visual evidence in the image. You can also add some facts you have observed from the images to make the reasoning logic more coherent.

3. Logical Connectivity: Explicitly bridge the gap between different levels and different question types in the same level. Explain how the perception in Level I supports the modeling in Level II, and how that modeling enables the final inference in Level III.

\textbf{Output Format:}
You must generate a comprehensive reasoning process, and place the reasoning process within <think></think> tags and place the final answer within <answer></answer> tags.

Even if the answer seems obvious from the provided facts, you are \textbf{STRICTLY PROHIBITED} from skipping the <think></think> stage. A response without a detailed <think></think> section is considered a failure.

Here is an example:

<think>

Stage 1 – Single-view perception

- In Figure 1, I observe one prominent tabletop across the lower-left foreground (bounding box [0, 793, 1143, 1439]).

- In Figure 2, I observe two separate tabletops: one stretching across the right side (bounding box [1122, 162, 1919, 1180]) and a shallow band-like tabletop along the upper center (bounding box [707, 0, 1226, 218]).
Stage 2 – Cross-view 3D modeling

- The viewpoints differ: Figure 1 looks from near the front-right toward the room, while Figure 2 is taken more from the left/side, looking along rows of desks.

- Comparing shapes, extents, and placements relative to monitors and room boundaries, none of the Figure 2 tabletops align in position/appearance with the single large foreground tabletop in Figure 1. Each appears to be a different physical desk segment, likely in other rows or out of Figure 1’s field of view.

Stage 3 – High-Level Contextual Reasoning

- Since the two tabletops in Figure 2 do not correspond to the one in Figure 1, all three are distinct physical tables.

- Total unique tables across both images = 2 (Figure 2) + 1 (Figure 1) = 3.

[The reasoning process here is generated according to Stage 1-3 of Reasoning Guidelines. You can split steps in every stage.]

</think>

<answer>B: 3[Final Answer]</answer>
\end{PromptBox}
\subsection{RL Data Filtering and Reward Design}
Specifically, we retain 20\% of the ScanNet++ scenes and 40\% of the Infinigen scenes for generating Level III QA data. This data is excluded from SFT training and instead used as candidate data for the RL training stage (Stage 3).
After completing Stage 1 training, we perform 10 inference runs on the above candidate data using the MV-STRIDE-SFT (Stage 1) model and compute the answer accuracy for each QA sample. We discard samples with 0\% and 100\% accuracy, effectively removing cases that are either trivial or excessively difficult. By retaining all remaining QA samples within the 0\%–100\% range, we construct a balanced and moderately difficult dataset for RL fine-tuning via GRPO.


\end{document}